\documentclass[twoside]{article}
\usepackage[accepted]{aistats2018}
\RequirePackage[l2tabu, orthodox]{nag} % checks for obsolete LaTeX packages and outdated commands
\usepackage{lipsum} % lorem ipsum blind text
\usepackage{calc}

\usepackage{natbib}

\usepackage[pdftex]{graphicx}
\graphicspath{{./figs/}}
\DeclareGraphicsExtensions{.pdf,.jpeg,.png}

\usepackage{adjustbox} %http://tex.stackexchange.com/questions/63229/shift-entire-tikz-picture-vertically-into-part-of-align-environment

\usepackage{tikz,pgfplots,pgfplotstable}
\usetikzlibrary{decorations.pathreplacing}
\usetikzlibrary{decorations.markings}
\usetikzlibrary{patterns}
\usetikzlibrary{positioning}
\usetikzlibrary{shapes,arrows}
\usetikzlibrary{calc}
\usetikzlibrary{intersections}
\usetikzlibrary{plotmarks}
\usetikzlibrary{spy,backgrounds}
\usetikzlibrary{pgfplots.groupplots}
\usepgflibrary{plotmarks}
\usepgfplotslibrary{units}
\usepgfplotslibrary{fillbetween}
\pgfplotsset{compat=1.12}
\usetikzlibrary{external}
\usepackage{mathtools} %extends amsmath and fixes bugs
\usepackage{scalerel,stackengine} % for \equalhat
\usepackage[eulergreek]{sansmath} % sans serif font available in math mode
\usepackage{bbm}
\usepackage{amssymb}

\usepackage[ruled,vlined]{algorithm2e}

\usepackage{array}
\usepackage{subcaption}

\usepackage{booktabs}
\usepackage{multirow}

\usepackage{wrapfig}
\usepackage{dblfloatfix}    % To enable figures at the bottom of page
\usepackage{placeins} % for \FloatBarrier
\usepackage{capt-of}
\usepackage{chngcntr}
\usepackage{siunitx} 
\usepackage[super]{nth} %http://tex.stackexchange.com/questions/4118/whats-the-quickest-way-to-write-2nd-3rd-etc-in-latex

\usepackage{microtype} % improves the spacing between words and letters

\usepackage[hidelinks]{hyperref}  % hyperlinks last, but before cleveref
\usepackage{cleveref}
\crefname{equation}{Eq.\!}{Eqs.}
\Crefname{equation}{Eq.\!}{Eqs.}
\crefname{table}{Tab.\!}{Tabs.}
\Crefname{table}{Tab.\!}{Tabs.}
\crefname{figure}{Fig.\!}{Figs.}
\Crefname{figure}{Fig.\!}{Figs.}
\crefname{algorithm}{Alg.\!}{Algs.}
\Crefname{algorithm}{Alg.\!}{Algs.}
\crefname{appendix}{App.\!}{Apps.}
\Crefname{appendix}{App.\!}{Apps.}
\crefname{pluralequation}{Eqs.\!}{Eqs.}
\Crefname{pluralequation}{Eqs.\!}{Eqs.}
\crefformat{pluralequation}{#2Eqs.\,(#1)#3}
\Crefformat{pluralequation}{#2Eqs.\,(#1)#3}
\crefrangeformat{pluralequation}{Eqs.~(#3#1#4) to~(#5#2#6)}
\Crefrangeformat{pluralequation}{Eqs.~(#3#1#4) to~(#5#2#6)}
\crefmultiformat{pluralequation}{Eqs.~(#2#1#3)}{ and~(#2#1#3)}{, (#2#1#3)}{ and~(#2#1#3)}
\Crefmultiformat{pluralequation}{Eqs.~(#2#1#3)}{ and~(#2#1#3)}{, (#2#1#3)}{ and~(#2#1#3)}

\definecolor{white}{rgb}{1.0,1.0,1.0}
\definecolor{brightred}{rgb}{1.0,0.1,0.1}
\definecolor{brightblue}{rgb}{0.0,0.0,0.8}
\definecolor{darkblue}{rgb}{0.0,0.0,0.5}
\definecolor{darkgreen}{rgb}{0.0,0.3,0.0}
\definecolor{brightgreen}{rgb}{0.0,0.8,0.0}
\definecolor{darkblack}{rgb}{0.0,0.0,0.0}
\definecolor{grey}{rgb}{0.3,0.3,0.3}

\newcommand{\changed}[1]{{\color{darkgreen}{}#1}}
\graphicspath{{./}}
\newcommand{\FF}{{\mathcal F}}
\newcommand{\GG}{{\mathcal G}}

\newcommand{\II}{{\mathcal I}}
\newcommand{\LL}{{\mathcal L}}

\newcommand{\OO}{{\mathcal O}}

\newcommand{\RRR}{\mathbbm{R}}

\newcommand{\KK}{{\mathcal K}}
\newcommand{\KKn}{\KK^{(n)}}

\newcommand{\GGn}{\GG^{(n)}}
\newcommand{\GGc}{\GG_{c}}

\newcommand{\muVec}{\vec{\mu}}

\newcommand{\scn}{s_c^{(n)}}
\newcommand{\qcm}{q_c^{(m)}}
\newcommand{\scm}{s_c^{(m)}}

\newcommand{\ct}{\tilde{c}}

\newcommand{\dcn}{d_{c}^{(n)}}
\newcommand{\dctn}{d_{\ct}^{(n)}}
\newcommand{\dnct}{d_{\ct}^{(n)}}
\newcommand{\dcpn}{d_{c'}^{(n)}}

\newcommand{\dcct}{d_{c\tilde{c}}}

\newcommand{\classk}{k}

\newcommand{\subc}{c}

\newcommand{\inpn}{n}

\NewDocumentCommand{\y}{O{}O{}}{y_{#2}^{\,#1}\!}
\NewDocumentCommand{\yVec}{O{}}{\vec{\y}^{\,#1}\!}
\newcommand{\yVecN}{\yVec[(\inpn)]}

\newcommand{\ThetaHat}{\hat{\Theta}}

\NewDocumentCommand{\Wgen}{O{}O{}}{\mathcal{W}_{#1#2}}

\NewDocumentCommand{\Rgen}{O{}O{}}{\mathcal{R}_{#1#2}}

\NewDocumentCommand{\W}{O{}O{}}{W_{#1#2}}

\NewDocumentCommand{\R}{O{}O{}}{R_{#1#2}}

\NewDocumentCommand{\Sc}{O{\subc}O{}}{s_{#1}^{#2}}

\NewDocumentCommand{\sVec}{O{}}{\vec{s}_{\vphantom{\subc}}^{\,#1}}

\NewDocumentCommand{\Igenc}{O{\subc}O{}}{\mathcal{I}_{#1}^{#2}}

\NewDocumentCommand{\Ic}{O{\subc}O{}}{I_{#1}^{#2}}

\NewDocumentCommand{\tk}{O{\classk}O{}}{t_{#1}^{#2}}

\NewDocumentCommand{\tVec}{O{}}{\vec{t}_{\vphantom{\classk}}^{\,#1}}

\NewDocumentCommand{\eps}{O{}}{\epsilon_{\textnormal{\tiny $#1$}}}

\NewDocumentCommand{\epst}{O{}}{\tilde{\epsilon}_{\textnormal{\tiny $#1$}}}

\newcommand{\con}{c_o^{(n)}}

\newcommand{\qn}{q^{(n)}}

\newcommand{\disT}{\textstyle}

\newcommand{\algBreak}{\\[1mm]}

\def\Vhrulefill{\leavevmode\leaders\hrule height 0.7ex depth \dimexpr0.4pt-0.7ex\hfill\kern0pt}

\newcommand{\defeq}{\vcentcolon=}

\newcommand{\myargmax}[1]{ \underset{#1}{\mathrm{argmax}} }
\newcommand{\myargmin}[1]{ \underset{#1}{\mathrm{argmin}} }

\newcommand{\undersets}[2]{\underset{\hbox to 0pt{\hss{\scriptsize #1}\hss}}{#2}}
\newcommand{\oversets}[2]{\overset{\hbox to 0pt{\hss{\scriptsize #1}\hss}}{#2}}

\begin{document}

% If your paper is accepted and the title of your paper is very long,
% the style will print as headings an error message. Use the following
% command to supply a shorter title of your paper so that it can be
% used as headings.
%
\runningtitle{A variational EM acceleration of GMMs and $k$-means}

% If your paper is accepted and the number of authors is large, the
% style will print as headings an error message. Use the following
% command to supply a shorter version of the authors names so that
% they can be used as headings (for example, use only the surnames)
%
%\runningauthor{Surname 1, Surname 2, Surname 3, ...., Surname n}

\twocolumn[

\aistatstitle{Can clustering scale sublinearly with its clusters?\\
A variational EM acceleration of GMMs and $k$-means}

\aistatsauthor{ Dennis Forster \And J\"org L\"ucke }

\aistatsaddress{
  Machine Learning, University of Oldenburg \&\\
  FIAS, Goethe-University Frankfurt, Germany\\
  \And
  Machine Learning\\
%  Machine Learning \& Excellence Cluster H4a\\
  University of Oldenburg, Germany
} ]

\begin{abstract}
%
%\vspace{-2mm}
One iteration of standard $k$-means (i.e., Lloyd's algorithm) or standard EM for Gaussian mixture models (GMMs) scales linearly with the number of clusters $C$, data points $N$, and data dimensionality $D$.
In this study, we explore whether one iteration of $k$-means or EM for GMMs can scale sublinearly with $C$ at run-time, while improving the clustering objective remains effective.
The tool we apply for complexity reduction is variational EM, which is typically used to make training of generative models with exponentially many hidden states tractable.
Here, we apply novel theoretical results on truncated variational EM to make tractable clustering algorithms more efficient.
The basic idea is to use a partial variational E-step which reduces the linear complexity of $\OO(NCD)$ required for a full E-step to a sublinear complexity.
Our main observation is that the linear dependency on $C$ can be reduced to a dependency on a much smaller parameter $G$ which relates to cluster neighborhood relations. 
%, related to the cluster neighborhood relationship.
We focus on two versions of partial variational EM for clustering: variational GMM, scaling with $\OO(NG^2D)$, and variational $k$-means, scaling with $\OO(NGD)$ per iteration.
Empirical results show that these algorithms still require comparable numbers of iterations to improve the clustering objective to same values as $k$-means.
For data with many clusters, we consequently observe reductions of net computational demands between two and three orders of magnitude.
More generally, our results provide substantial empirical evidence in favor of clustering to scale sublinearly with~$C$.
\end{abstract}

\section{Introduction}
Clustering is a core Machine Learning task and one of the most widely used types of algorithms in general. % \citep[][]{}.
$k$-means and Gaussian mixture models (GMMs) represent two of the most popular clustering algorithms \citep[see, e.g.,][]{Berkhin2006,McLachlanPeel2004}.
Both require $\OO(NCD)$ numerical operations per iteration (see abstract for definitions of $N$, $C$, and $D$).
Active ongoing research provides ever improving bounds on convergence times in terms of iteration steps \citep{ArthurEtAl2009,MoitraValiant2010,XuEtAl2016}, and empirical results show very fast convergence in practice \citep{Duda2007}.
\paragraph{Related work and own contribution.}
Convergence times can be strongly improved by careful seeding, and seeding methods have recently been made very efficient \citep{BachemEtAl2016b}.
After seeding, the limiting factor for efficiency remains the complexity of one $k$-means or GMM iteration \citep[e.g.,][for a discussion]{BachemEtAl2016a}, which can be very relevant in practice \cite[e.g.,][]{RosenbaumWeiss2015}.
The reduction of complexity per iteration (e.g., of $k$-means) is therefore also the goal of other popular approaches.
The efficiency of distance computations can, e.g., be improved by exploiting the triangle inequality \citep{Elkan2003}, or random projections for independence of the dimensionality $D$ \citep{ChanLeung2017}.
And by following the idea of \citet{Moore1999} or coresets \citep{HarPeledMazumdat2004,FeldmanEtAl2011,LucicEtAl2017,BachemEtAl2017} the dependency on the number of data points $N$ can drastically be reduced.
In this work our focus is on reducing the linear dependency on the number of clusters~$C$.
There are two views which may highlight the importance of this dependency:
First, with very large numbers of data points $N$, a complexity reduction, e.g., from $\OO(NCD)$ to $\OO(NGD)$ with $G\ll C$ provides a very large reduction in terms of required computations.
Second, if coresets are used to reduce $N$, and triangle inequalities or random projections are used to reduce the dependence on $D$, then the dependence on $C$ remains the main bottleneck.

Our tool for complexity reduction is the application of truncated distributions to approximate exact posteriors.
Truncated approximations have been applied to a number of probabilistic data models, and most frequently to multiple-causes models for which the number of latent states increases
exponentially with the number of latents \citep{PuertasEtAl2010,DaiEtAl2013,HennigesEtAl2014,SheikhEtAl2014}.
Truncated approximations do not assume a-posteriori independence like factored variational approaches \citep{SaulEtAl1996,JordanEtAl1999}, they typically result in tight free energy bounds, and they are very efficient \citep{SheikhEtAl2014,SheikhLucke2016}.
%For sparse coding, for instance, the largest scale results reported so far ($10\,000$ generative fields) were obtained using truncated approaches for training \citep{SheikhLucke2016}.
%
For mixture models, truncated distributions represent a very natural choice.
If considering the posterior of a given data point (i.e., its cluster responsibilities), then typically only few clusters contribute significantly to the over-all posterior mass.
%All remaining clusters have very low probability values (very low responsibilities).
Truncated distributions approximate the full posteriors by maintaining just the $C'$ highest posterior values while setting all other values to zero (see \cref{FigVarGMM}, top).
\citet{DaiLucke2014} used truncated distributions to make position invariant mixture models for images more efficient,
\citet{ForsterLucke2017} applied truncated distributions to standard Poisson mixtures, and \citet{SheltonEtAl2017}, \citet{HughesSudderth2016}, and \citet{LuckeForster2017} used truncated posteriors for standard GMMs.
In all these applications, truncated distributions reduced the computational cost compared to exact EM.
None of these contributions aimed to show or have shown that one complete EM iteration for clustering can scale sublinearly with $C$.
While, e.g., \citet{HughesSudderth2016} and \citet{ForsterLucke2017} discuss the reduction of M-step complexity, and while $k$-means \citep[see][for its variational EM formulation]{LuckeForster2017} has an M-step complexity of $\OO(ND)$, all these algorithms use a full variational E-step with $NC$ distance evaluations ($\OO(NCD)$ computations).
Other work with focus on $k$-means also aimed at reducing the dependency on $C$.
Often such work remains theoretical, including run-time analysis in terms of iterations \citep[see][for a discussion of the literature]{KanungoEtAl2002}, but also concrete suggestions for practically applicable versions of $k$-means have been made \citep[][and others]{Phillips2002,ShindlerEtAl2011,Curtin2017}.
%\citet{Phillips2002}, \citet{ShindlerEtAl2011}, and \citet{Curtin2017} are all tailored to $k$-means. 
\citet{ShindlerEtAl2011} focus on efficient memory usage. \citet{Phillips2002} shows a reduction to $\OO(ND\gamma + C^2D + C \log(C))$ run-time complexity
per iteration but $\gamma$ is large (on the order of $C$) initially. \citet{Curtin2017} uses tree-based algorithms and novel pruning strategies to achieve a complexity (after initialization, per iteration, $D$ not considered) of $\OO(N + C \log(C))$ under mild assumptions.
%
%For instance \citet{Phillips2002} shows a reduction of $k$-means to a complexity of $\OO(ND\gamma + C^2D + C \log(C))$ per iteration.
%However, $\gamma$ is large (on the order of $C$) in the beginning of learning. \citet[][]{} has very recently shown that 

To the knowledge of the authors, neither \citet{Phillips2002,Curtin2017} nor any other contribution shows or has considered possible a complexity reduction in $C$ to a level similar to the one reported here.
More specifically, we are not aware of work that effectively improves a $k$-means or GMM objective while the complexity of one iteration does not depend on $C$.\vspace{-1mm}

\section{Truncated EM for GMMs}
Given a set of $N$ data points, $(\yVec^{(1)},\ldots,\yVec^{(N)})$, our goal is to find parameters $\Theta$ that maximize the data log-likelihood,
where $p(\yVec\,|\,\Theta)$ is given by a GMM with isotropic, equally weighted Gaussians:\vspace{0mm}
\begin{align}
&\disT \LL(\Theta) = \sum_{n} \log\!\big( p(\yVecN\,|\,\Theta) \big)\text{, where} \label{EqnLikelihood}\\
&\disT p(\yVec\,|\,\Theta) = \frac{1}{C} (2\pi\sigma^2)^{-\frac{D}{2}} \sum_{c} \exp\!\big(\!-\!\frac{1}{2\sigma^2}\|{}\yVec-\muVec_c\|{}^2\big).\label{EqnGMMIso}
\end{align}
The presumably most popular method to optimize a GMM %is EM which here is given by the following equations:
is EM, which consists for \cref{EqnGMMIso} of the updates: %given by: %of the following equations:
\begin{align}
&\!\! \disT \dcn = \|{}\yVecN-\muVec_c\|,\,\, \scn = \frac{\exp\!\big(-\frac{1}{2}(\dcn / \sigma)^2\big)} {\sum_{c'}\exp\!\big(-\frac{1}{2}(\dcpn / \sigma)^2\big)}, \label[pluralequation]{EqnGMMEStep} \\
&\!\! \disT \muVec^{\mathrm{\,new}}_c \!\! = \! \frac{\sum_{n}\!\scn\yVecN}{\sum_{n}\scn}\!,
\sigma_{\mathrm{new}}^2 \!\! = \! \frac{1}{DN}\!\sum\limits_{n,c}\scn \|\yVecN-\muVec^{\mathrm{\,new}}_c\|^2\!.\!\!\!
\label[pluralequation]{EqnGMMMStep}
\end{align}
Standard EM for the model (\ref{EqnGMMIso}) iterates E-step (\ref{EqnGMMEStep}) and M-step (\ref{EqnGMMMStep}), and each iteration changes the parameters $\Theta=(\muVec_{1:C},\sigma^2)$ such that the likelihood is monotonously increased (until convergence to potentially local maxima).
We will refer to this algorithm as standard GMM.\vspace{-1mm}
%
%Iterating E-step (\ref{EqnGMMEStep}) and M-step (\ref{EqnGMMMStep}) represents (a special case of) the well-known GMM clustering algorithm which
%we will here refer to as standard GMM or simply GMM. Standard GMM changes the parameters $\Theta=(\muVec_{1:C},\sigma^2)$ such that the likelihood
%is monotonously increased to (potentially local) maxima.\vspace{-1mm}

\begin{figure}[b!]
\centering
  \vspace{4pt}
	\resizebox{0.41\textwidth}{!}{
  		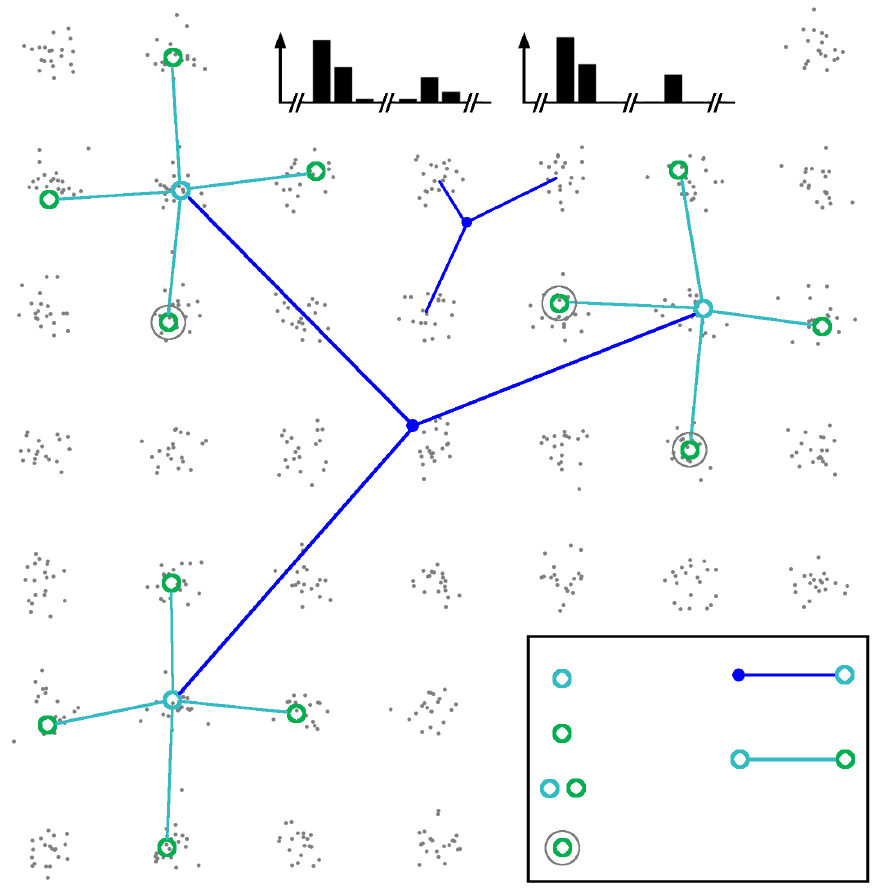
	}
  %\vspace{-3pt}
	\caption{
\textbf{Top:} Exact responsibilities $\scm$ and truncated responsibilities $\qcm$ for a data point $\yVec^{(m)}$ with $\KK^{(m)}$ containing the closest ${C'=3}$ clusters.
% ($\qcm$ abbreviates $q^{(m)}(c;\KK,\Theta)$ of \cref{EqnQMain}).
\textbf{Remainder:} Illustration of the search space $\GGn$ to find clusters increasingly close to $\yVecN$.
The search space consists of the clusters in $\KKn$ and the nearest neighbors of these clusters ($\GGc$ with $c\in\KKn$).
For the illustration we used well separated clusters, $|\KKn|=C'=3$ and $|\GGc|=G=5$.
Cluster centers $\muVec_c$ were assumed here to already represent the clusters well.
See \cref{app:IllustrationOFAlgorithms} for more information and a complete illustration across iterations.}
  \label{FigVarGMM}
\end{figure}

In order to reduce the complexity of standard GMM, we here apply variational EM.
Instead of seeking to maximize the likelihood directly, the basic idea of variational EM is to maximize a lower-bound of the likelihood, the free energy.
The free energy depends on variational distributions which are chosen to (A)~approximate exact posterior distributions $p(c\,|\,\yVec,\Theta)$ as closely as possible, and to (B)~result in a less complex optimization objective.
For our purposes, we use variational distributions $\qn(c;\KK,\ThetaHat)$ which depend 
on two types of variational parameters, sets of states $\KK=(\KK^{(1)},\ldots,\KK^{(N)})$ and $\ThetaHat$:
\begin{align}
\!\!\!\! \scn \approx \qn(c;\KK,\ThetaHat)
%
% next line can potentially be removed
%&= \frac{p(c|\yVecN,\ThetaHat)}{\!\!\!\!\sum\limits_{c'\in\KKn}\!\!\!\!p(c'|\yVecN,\ThetaHat)}\delta(c\in\KKn) \\
%
& = \frac{p(c,\yVecN|\ThetaHat)}{\!\!\!\!\sum\limits_{c'\in\KKn}\!\!\!\!p(c',\yVecN|\ThetaHat)}\,\delta(c\in\KKn),\!\!
\label{EqnQMain}
\end{align}
where $\delta(c\!\in\!\KKn)=1$ if $c\!\in\!\KKn$ and zero otherwise. 
The used variational distributions \cref{EqnQMain} are truncated posteriors, i.e., they are proportional to the exact posterior for all $c\in\KKn$ while they are exactly zero for all other $c\notin \KKn$.
\Cref{FigVarGMM} (top: $\scm$ and $\qcm$) illustrates the approximation.
If we choose $\KKn$ for the GMM to contain all $c$, i.e.\, $\KKn=\{1,\ldots,C\}$, we recover standard GMM. 
If we choose $\KKn$ to contain just one single element, we can recover standard $k$-means \citep{LuckeForster2017}, without having to take the limit to zero variances $\sigma^2$.
Given the variational distributions \cref{EqnQMain}, the corresponding free energy is: % given by:
\begin{align}
\FF(\KK,\ThetaHat,\Theta) \! \defeq &
    \sum\limits_{n} \! \Big( \!
      \sum_{c} \qn(c;\KK,\ThetaHat) 
       \log \!\big( p(c,\yVecN\,|\,\Theta)\big) \!
      \Big) \nonumber \\
      & + \sum\limits_{n}H\big(\qn(c;\KK,\ThetaHat)\big)\,,
\label{EqnFreeEnergy}
\end{align}
where $H(p(c))$ is the entropy of a distribution $p(c)$. 
Maximization of the free energy \cref{EqnFreeEnergy} involves a three-stages optimization w.r.t.\ the parameters $\KK$, $\ThetaHat$ and $\Theta$.
Because of the standard functional form of the free energy \cref{EqnFreeEnergy}, the update equations for $\Theta$ (the M-step) remain identical to \cref{EqnGMMMStep} but use the approximation \cref{EqnQMain} instead of the exact $\scn$ in \cref{EqnGMMEStep}.
%The computational 
The complexity of one M-step is given by the number of non-zero $\scn$ values and by $D$, i.e., for $|\KKn|=C'$ one update of $\Theta$ has a complexity of $\OO(NC'D)$.
%Finding the $\KKn$ which maximize the free energy (full E-step) in the first optimization (\ref{EqnFTrunc}) is of $\OO(NCD)$.

For the variational E-step it was shown \citep[][]{Lucke2016} that it is sufficient to set $\ThetaHat=\Theta$ and to maximize a simplified expression of $\FF(\KK,\Theta,\Theta)=:\FF(\KK,\Theta)$:
\begin{equation}
\KK^{\ast} = \myargmax{\KK}{\big\{\FF(\KK,\Theta)\big\}}
\end{equation}
Here, we now replace the maximization of the free energy by an increase of the free energy (partial E-step).
For the GMM of \cref{EqnGMMIso} it can be shown %\citep[see][]{LuckeForster2017}
that an increase of $\FF(\KK,\Theta)$ has a direct geometrical interpretation (\cref{app:ProofProp1} provides the proof following \citet{LuckeForster2017} and \citet{Lucke2016}):\vspace{-2mm}

\paragraph{Proposition 1.}
If we replace for an arbitrary $n$ a cluster $c\in\KKn$ by a cluster $\ct\not\in\KKn$ such that \mbox{$\|\yVecN-\muVec_{\ct}\|<\|\yVecN-\muVec_{c}\|$}, then $\FF(\KK,\Theta)$ increases.

%
%
%
%\section{Efficient Algorithms Based on Partial E-Steps}
\section{Efficient Partial EM}
Our algorithms iteratively increase the free energy using Prop.\,1 and M-step \cref{EqnGMMMStep} until convergence.
By relating partial EM to the geometric interpretation of cluster distances (Prop.\,1), our approach maintains the guarantee to monotonously increase the free energy whenever clusters are found that are closer to $\yVecN$ than those previously considered.
Finding closer clusters can be realized much more efficiently than finding the closest clusters (which is of order $\OO(NCD)$).
To reduce run-time complexity, our goal is to as efficiently as possible find clusters $c$ for each $\KKn$ such that the free energy is increased as effectively as possible.
The free energy will always increase if we use Prop.\,1 to update $\KKn$.
However, blindly (e.g. uniform randomly) searching for closer clusters can be expected to be very inefficient.
Instead, we here follow the strategy of defining a search space for each data point based on nearest cluster neighborhoods of each cluster in $\KKn$.

Although more efficient in terms of computational complexity, partial E-steps require us to memorize the variational parameters $\KKn$ across iterations.
With one set $\KKn$ for each data point, we have with $|\KKn|=C'$ an additional memory requirement of $\OO(NC')$ for all our variational algorithms.

\begin{algorithm}[t]
init $\muVec_{1:C}$, $\sigma$ and $\KKn$ for all $n$;\algBreak
\Repeat{$\muVec_{1:C}$ and $\sigma^2$ have converged\vspace{0.0mm}}
{
\For{$c=1:C$}{			       						
  \For{$\ct=1:C$}{								
    $\dcct = \|\muVec_{\ct}\,-\,\muVec_c\|$;                  
  }
  $\GGc= \{\ct\,|\,d_{c\ct}$ is among the $G$ \linebreak
  \phantom{$\GGc= \{\ct\,|$}smallest distances $d_{c:}\}$;
}
	\For{$n=1:N$}{   						                                                           
  $\GGn =\bigcup_{c\in\KKn}\GGc$;\algBreak
  \For{$c\in\GGn$}{
      $d^{(n)}_{c} = \|\yVecN\,-\,\muVec_{c}\|$;
  }
  $\KKn = \{c\,|\,d^{(n)}_{c}$ is among the $C'$ \linebreak
  \phantom{$\KKn=\{c\,|$}smallest distances$\}$;

}
\For{$n=1:N$}{   		
  \For{$c\in\KKn$}{								
            $s_c^{(n)} = \frac{\exp\!\big(-\frac{1}{2}(\dcn / \sigma)^2\big)}{\sum_{c'\in\KKn}\exp\!\big(-\frac{1}{2}{(\dcpn / \sigma)}^2\big)}$;
  }
}	
update $\muVec_{1:C}$ and $\sigma^2$ using \cref{EqnGMMMStep} with \cref{EqnQMain};  
}
\caption{GMM with partial truncated E-step and exhaustive cluster neighborhood evaluation.
\label{alg:var-GMM-X}}
\end{algorithm}

\subsection{Exhaustive Cluster Neighborhoods}
To illustrate our approach, first consider \cref{FigVarGMM}.
It visualizes a set $\KKn$ and its distances to the data point $\yVecN$ and its closest neighboring clusters.
We denote the set of clusters consisting of $c$ and its $G-1$ nearest neighboring clusters by $\GGc$ and the union of sets $\GGc$ belonging to clusters $c\in\KKn$ as $\GGn$.
To update $\KKn$ such that it only includes clusters with equal or smaller distances to $\yVecN$ as before, we now won't evaluate all data-to-cluster distances.
Instead, we only calculate the distances of $\yVecN$ to the clusters in $\GGn$ and choose the $C'$ closest clusters from these as new $\KKn$.
Optimization of $\KKn$ then involves at most \mbox{$C'G<C$} distance evaluations, and it can be considered much more likely than random search that clusters~$c$ which improve the free energy are found.
However, we require additional computations to determine the nearest neighbors for each cluster.
These computations are, however, independent of $N$. 

\cref{alg:var-GMM-X} shows the complete algorithm, which consists of three blocks of computation: (1st block)~computation of cluster-to-cluster distances and definition of sets $\GGc$; (2nd block)~computation of data-to-cluster distances and update of $\KKn$; and (3rd block)~update of model parameters $\Theta$.
The complexities are given as follows:

The first block computes the nearest neighbor sets $\GGc$ (where we do not exploit symmetry of $d_{c\ct}$ for simplicity).
The complexity of the required distance evaluations is $\OO(C^2D)$.
%, which is dominated by the $\OO(C^2)$ distance evaluations.
%The following definition of the sets $\GGc$ is less demanding.
The following problem of finding the $G$ smallest elements in an array of $C$ elements is an unordered partial sorting problem.
This is provably solvable with complexity $\OO(C)$, e.g., by applying the median-of-medians algorithm \citep{BlumEtAl1973} or introselect \citep{Musser1997}.

The second block computes for each data point the distances to all clusters in $\GGn$ consisting of all $G$ neighbors $\GGc$ (including $c$) of all $C'$ clusters $c\in\KKn$.
We thus have to compute $\OO(C'G)$ distances for each $n$, which amounts to $\OO(NC'GD)$.
If the clusters in $\KKn$ are close together, they might share some nearest neighbors, which reduces the actual number of distance computations.
Also note that the $\OO(C'G)$ can never exceed the total number of clusters $C$.
To update one set of $\KKn$, we again have to solve an unordered partial sorting problem.
For a given $n$, we have $C'G$ distances, such that finding the $C'$ smallest elements is of complexity $\OO(C'G)$, with $\OO(NC'G)$ total evaluations.
Hence, the evaluations of distances $\dcn$ dominate the second block, which has an overall complexity of~$\OO(NC'GD)$.

Finally, the third block computes the responsibilities $\scn$ and updates the model parameters $\muVec_{1:C}$ and $\sigma^2$.
Both have a complexity of $\OO(NC'D)$.

The overall complexity of one EM iterations of \cref{alg:var-GMM-X} is consequently given by $\OO(NC'GD+C^2D)$.
Storage of sets $\GGc$ requires $\OO(CG)$ additional memory, which results in a total of $\OO(CD+NC'+CG)$ memory demand.
We provide an illustration of the algorithm in \cref{app:IllustrationOFAlgorithms} and a line-by-line complexity analysis in \cref{app:ComplexityAnalysis}.

\subsection{Estimated Cluster Neighborhoods}
For \cref{alg:var-GMM-X}, we introduced new sets $\GG_c$ that require additional evaluations of cluster-to-cluster distances.
This can allow for a complexity reduction in the updates of our variational parameters $\KKn$, but adds a computational cost of $\OO(C^2D)$.
With \Cref{alg:var-GMM-S} we show a new way to further reduce this algorithmic complexity.

In \Cref{alg:var-GMM-S}, the data-to-cluster distance computations and updates of the sets $\KKn$ remain exactly as in \cref{alg:var-GMM-X}.
However, to further improve efficiency, we now {\em estimate} the cluster-to-cluster distances $d_{c\ct}$ using the data-to-cluster distances $\dcn$ which we anyway have to compute.
To do this, the ordering of the first and second blocks of \cref{alg:var-GMM-X} now changes -- \cref{alg:var-GMM-S} first computes the data-to-cluster distances $\dcn$ before the sets $\GGc$ are updated -- and we replace the exhaustive $\GGc$ evaluation of \cref{alg:var-GMM-X} by estimated cluster neighborhoods.

To illustrate the estimation approach, first consider two non-overlapping clusters $c$ and $\ct$ with cluster centers $\muVec_c$ and $\muVec_{\ct}$ that already represent the two clusters well.
If we denote by $\II_c$ the set of all points $\yVecN$ that have $c$ as closest cluster, ${\II_c=\{n\,|\,c=\myargmin{\ct=1:C}\,\|\yVecN-\muVec_{\ct}\|\}}$\vspace{-4pt}, we may
estimate:
\begin{align}
d_{c\ct} \approx \frac{1}{|\II^*_c|} \!\!\! \sum_{\substack{n\in\II_c \\ \ \mathrm{if}\,\ct\in\GGn}}\!\!\!\! \|\yVecN - \muVec_{\ct}\| = \frac{1}{|\II^*_c|}\!\!\!\sum_{ \substack{n\in\II_c \\ \ \mathrm{if}\,\ct\in\GGn}}\!\!\!\! \dctn\,.
\label{EqnDCCEstimation}
\end{align}
If we sum over all $n\in\II_c$ (by ignoring the additional conditions for the sums and assuming $\II_c=\II^*_c$ for now), then \Cref{EqnDCCEstimation} becomes exact for well separated clusters and close to optimal convergence points. % (we discuss the additional conditons in the following).
However, in general not all the $\dctn$ that we require for \cref{EqnDCCEstimation} have been computed before.
We therefore introduce the condition $\ct\in\GGn$, which ensures that only data points $n$ with known $\dctn$ are summed over (see \cref{app:IllustrationOFAlgorithms} for an example).
$|\II^*_c|$ is then simply the number of summands.
For already well-learned $\KKn$ and cluster centers $\muVec_c$, \cref{EqnDCCEstimation} estimates the $d_{c\ct}$ of close-by clusters well (and we are only interested in those close-by distances).
The summation in \cref{EqnDCCEstimation} for distances to far-away clusters may not contain any data points and are ignored in practice.
Away from convergence of $\Theta$ and $\KKn$, the estimation of $d_{c\ct}$ may be very coarse, but because of Prop.\,1 the updates of $\KKn$ still warrant a monotonous increase of the free energy.

Regarding the computational complexity of \cref{alg:var-GMM-S}, the first and last block are identical to the second and third block of \cref{alg:var-GMM-X}.
Thus, we now only have to evaluate the complexity of the new middle part for the $\GGc$ estimation, replacing what was the first block of \cref{alg:var-GMM-X}.

This new middle part of \cref{alg:var-GMM-S} consists of two steps: definition of auxiliary sets $\II_c$ and subsequent estimation of cluster neighborhoods $\GGc$.
The definition of all sets $\II_c$ here requires a complexity of $\OO(NC'G)$.
In the next computational block, the sets $\II_c$ are used to estimate the distances $d_{c\ct}$ according to \cref{EqnDCCEstimation}.
There we loop over all clusters $c$ and $\ct$ and then over all $n\in\II_c$ to estimate the distances. 
But there we only consider already computed $\dctn$ (using the condition of Eq.\,\ref{EqnDCCEstimation}).
The loops over $c=1:C$ and $\ct=1:C$ of the third block may at first seem to suggest that their complexity depends on $C$.
We can however rewrite the loops to show that this is actually not the case.
If we formalize them as sums for computing functions $F^{(n)}_{c\ct}$, we obtain:
\vspace{-14pt}
\begin{align}
\sum_{c=1}^C\sum_{\ct=1}^C\!\sum_{n\in\II_c}\!\delta(\ct\in\GGn)\,F^{(n)}_{c\ct}  =  \sum_{c=1}^C\sum_{n\in\II_c}\!\!\!\!\sum_{\,\,\,\,\,\ct\in\GGn}\!\!\!\!F^{(n)}_{c\ct},
\label{EqnLoops}
\end{align}

\vspace{-8pt}
where $\delta(\ct\in\GGn)$ is an indicator function representing the `if'-condition.
Further note that the sets $\II_c$ contain $(N/C)$ data points on average and that $\sum_{c=1}^C\sum_{n\in\II_c}$ by construction of $\II_c$ then goes over exactly $N$ values.
As a set $\GGn$ contains at most $C'G$ elements, the total summation, i.e., the total cost of $d_{c\ct}$ estimations in \cref{alg:var-GMM-S}, has a complexity of $\OO(NC'G)$.
For each $c$ our procedure estimates on average $(N/C)C'G$ distances $d_{c\ct}$.
For each $\GGc$, finding the $G$ smallest elements (unordered partial sorting) then requires a computational cost of on average $\OO((N/C)C'G)$, with a total cost for defining all sets $\GGc$ of $\OO(NC'G)$.
Storage of all $\dcn$ distances within one iteration results (without additional measures) in an extra $\OO(NC'G)$ memory demand.
% Since the sets $\GGc$ have to be stored across iterations, we now have an additional memory requirement of $\OO(CG)$.

Everything taken together, \cref{alg:var-GMM-S} has a run-time complexity of $\OO(NC'GD)$ and a memory demand of $\OO(CD+NC'G+CG)$ for the storage of all model parameters $\muVec_c$ and $\sigma$, variational parameters $\KKn$ and nearest neighbors $\GGc$, and computed distances $\dcn$.
A line-by-line complexity analysis of a more detailed version of \cref{alg:var-GMM-S} can be found in \cref{app:ComplexityAnalysis}.

Considering \cref{alg:var-GMM-S}, e.g., the estimation of cluster-to-cluster distances may be deemed to cause problems.
%many things may be deemed to cause problems, especially the estimation of cluster-to-cluster distances.
For the sake of complexity reduction, we accepted very coarse estimates.
However, those estimates were far from random and can in principle and finally be good estimates for close-by clusters.
Because of the way we defined the $\KKn$ updates in \cref{alg:var-GMM-S}, a data-driven and finally relatively precise cluster-to-cluster distance estimation is all we require.
The reason is that Prop.\,1 warrants that the $\KKn$ updates will always monotonously increase the free energy, and the M-step with these updated $\KKn$ will in turn
monotonously increase the objective.
Such a provably monotonous increase ensures that learning proceeds in the right direction, and we know that truncated free-energies can result in very tight lower likelihood bounds (see also \cref{app:DetailedResults}).
If \cref{alg:var-GMM-S} really does result in efficient optimization of the clustering objective, remains to be verified and investigated empirically in the next section.

\begin{algorithm}[t]
init $\muVec_{1:C}$ and $\sigma^2$, and init $\GGn$ for all $n$;\algBreak
\Repeat{$\muVec_{1:C}$ and $\sigma^2$ have converged\vspace{0.5mm}}
{
\For{$n=1:N$}{
  $\GGn =\bigcup_{c\in\KKn}\GGc$;\algBreak
  \For{$c\in\GGn$}{
      $d^{(n)}_{c} = \|\yVecN\,-\,\muVec_{c}\|$;
  }
  $\KKn = \{c\,|\,d^{(n)}_{c}$ is among the \linebreak 
  \phantom{$\KKn=\{c\,|$}smallest distances$\}$;
}
\For{$n=1:N$}{
  $\con = \myargmin{c\in\GGn}\,\big\{d^{(n)}_{c}\big\}$;\\%\algBreak
  $\II_{\con} = \II_{\con} \cup \{n\}$;
}
\For{$c=1:C$}{                                      
  \For{$\ct=1:C$}{                                   
    \For{$n\in\II_c$}{
      \If{$\ct\in\GGn$}{
        $d_{c\ct} =  d_{c\ct} + d^{(n)}_{\ct}$;
      }
    }
    $d_{c\ct} =  d_{c\ct} \,/\, |\II^*_c|$;
  }
  $d_{cc}=0;$\\
  $\GGc= \{\ct\,|\,d_{c\ct}$ is among the $G$ \linebreak
  \phantom{$\GGc= \{\ct\,|$}smallest distances $d_{c:}\}$;
}
\For{$n=1:N$}{
%  $\GGn = \cup_{c \in \KKn}\GG_c$;\\
  \For{$c\in\KKn$}{
%    $d_c^{(n)} = \|\yVecN\,-\,\muVec_c\|$;\algBreak
    $s_c^{(n)} = \frac{\exp\!\big(-\frac{1}{2}(\dcn / \sigma)^2\big)}{\sum_{c'\in\KKn}\exp\!\big(-\frac{1}{2}{(\dcpn / \sigma)}^2\big)}$;
  }
}
update $\muVec_{1:C}$ and $\sigma^2$ using \cref{EqnGMMMStep} with \cref{EqnQMain};  
}
\caption{GMM with partial truncated E-step and estimated cluster neighborhood evaluation.\label{alg:var-GMM-S}}
\end{algorithm}

\section{Numerical Experiments}
%Verification of Efficient Free Energy Optimization}
%
Theoretical investigations of convergence are typically very intricate already for standard $k$-means \citep[e.g.,][]{HarPeledSadri2005} or standard GMMs \citep[e.g.,][]{XuEtAl2016}.
We therefore verify sufficiently efficient optimization of the clustering objectives numerically.

\Cref{alg:var-GMM-X,alg:var-GMM-S} both allow for different choices of $G$ and $C'$, which optimized on a variety of data sets would quickly result in a large combinatorics of different algorithms and data sets.
As our primary aim here is to show that a clustering which scales sublinearly with $C$ is possible, we focus on two algorithms: variational GMM and variational $k$-means (see below).
We use a BIRCH artificial data set with $5\times{}5$ clusters to show the viability of the exhaustive and estimated algorithms (\cref{alg:var-GMM-X,alg:var-GMM-S}) compared to their non-variational counterparts.
We then use the more efficient estimated algorithm (\cref{alg:var-GMM-S}) on artificial BIRCH data sets of up to $C=4096$ clusters to show that the gained computational benefits are not traded off by a higher number of necessary training iterations.
Further experiments on KDD2004 ($C=200$) and SONG ($C=2000$) demonstrate large-scale applicability of the algorithms to natural data.

\paragraph{Variational GMM.}
For simplicity we choose $C'=G$ such that we remain with one parameter $G$ that relates to cluster neighborhood relations.
The resulting algorithm is similar to EM for GMM but optimizes a variational free energy instead of the likelihood directly. 
We therefore refer to \cref{alg:var-GMM-X} with $C'=G$ as variational-GMM-exhaustive (var-GMM-X), and to \cref{alg:var-GMM-S} with $C'=G$ as variational-GMM-estimated (var-GMM-S).

\paragraph{Variational $k$-means.}
It was recently shown \citep{LuckeForster2017} that $k$-means is equivalent to truncated variational EM with a full E-step if we choose $|\KKn|=C'=1$.
If we maintain the choice of $C'=1$ but replace the full E-step by a partial E-step with $G < C$ following \cref{alg:var-GMM-X} or \cref{alg:var-GMM-S}, we obtain algorithms for which each iteration is more efficient in terms of required distance evaluations than $k$-means.
Following the naming above, \cref{alg:var-GMM-X} with $C'=1<G<C$ will be referred to as var-$k$-means-X, and \cref{alg:var-GMM-S} with $C'=1<G<C$ we will refer to as var-$k$-means-S.

\paragraph{Complexities.}
The complexities of the four algorithms above directly result from the complexity considerations of \cref{alg:var-GMM-X,alg:var-GMM-S}.
We here summarize the run-time and memory demand of the algorithms.
The main approximation parameter is $G$, while keeping in mind that for var-GMM $C'=G$ and for var-$k$-means $C'=1$.
By inserting into the previously derived complexity formulas for \cref{alg:var-GMM-X,alg:var-GMM-S}, we obtain \cref{tab:complexities}:

%\begin{table}[h]
%	%\caption{Computational complexities of \cref{alg:var-GMM-X,alg:var-GMM-S}.}\vspace{4pt}
%	\centering
%	\small
%\ \\[-2mm]
%	\begin{tabular}{lcccc}
%	\toprule
%%\ \\\hline
%	\ \ Tab.\,1	&  var-GMM-X & var-GMM-S & var-$k$-means-X & var-$k$-means-S \\\hline
%	%\midrule
%	Run-Time &  $\OO(NG^2D+C^2D)$ & $\OO(NG^2D)$ & $\OO(NGD+C^2D)$ & $\OO(NGD)$\phantom{$\int^f$} \\
%	Memory   &  $\OO(CD+NG)$      & $\OO(CD+NG)$ & $\OO(CD+N)$     & $\OO(CD+N+CG)$ \\\hline	
%	%\bottomrule
%	\end{tabular}\vspace{-1mm}
%	\label{tab:complexities}
%\end{table}

\begin{table}[h]
	\vspace{-5pt}
	\caption{Computational complexities, see also \cref{app:ComplexityAnalysis}}
	\vspace{-8pt}
  {
%\fontsize{8}{10} \sffamily \sansmath
\renewcommand{\arraystretch}{1.05}
\newcommand{\midspace}{\hphantom{abc}}
\setlength{\tabcolsep}{3pt}
\small
	\begin{tabular}{@{}lll}
  \toprule
%\ \\\hline
	&  Run-Time & Memory \\
	\midrule
	var-GMM-X       &  $\OO(NG^2D+C^2D)$  & $\OO(CD+NG+CG)$ \\
	var-GMM-S       &  $\OO(NG^2D)$       & $\OO(CD+NG^2+CG)$ \\
	var-$k$-means-X &  $\OO(NGD+C^2D)$    & $\OO(CD+N+CG)$  \\
	var-$k$-means-S &  $\OO(NGD)$         & $\OO(CD+NG+CG)$ \\	
	\bottomrule
	\end{tabular}
}
	\label{tab:complexities}
	\vspace{-10pt}
\end{table}

\paragraph{Data sets.}
We apply the var-GMM and var-$k$-means algorithms to three data sets:
(I)~A `BIRCH' data set \citep[based on][]{ZhangEtAl1997}, which is an artificial $D=2$-dimensional data set of $C$ isotropic Gaussians, arranged in equal distances on a $\sqrt{C}\times\sqrt{C}$ grid.
We investigate settings of $C=25$~to~$4096$ Gaussians with $\sigma^2 = 1$ in nearest neighbor distances of $4\sqrt{2}$ and $100$~samples per cluster, i.e., $N=$~\num{2500}~to~\num{409600} samples in total.
(II)~The KDD-Cup 2004 Protein Homology (KDD2004) data set, which was originally designed as a supervised classification set for the KDD competition 2004 but is also frequently used as a clustering benchmark.
It consists of $N=$~\num{145751} data points with $D=74$ numerical features each.
(III)~The Year Prediction Million Song Dataset (SONG) \citep{BertinMahieuxEtAl2011}, which contains $N=$~\num{515345} data points with $D=$~\num{90} audio features (timbre averages and covariances).
This represents the largest data set in our experiments.

\subsection{Empirical Results}
For \cref{alg:var-GMM-X,alg:var-GMM-S}, we monitor the standard quantization error during training: $\phi = \sum_n \min_{c\in C} \|\yVecN - \muVec_{c}\|^2$.
%\begin{equation}
%\phi = \sum_n \min_{c\in C} \|\yVecN - \muVec_{c}\|^2.
%\end{equation}
On the BIRCH and KDD data sets we allow for a maximum of 200 and on SONG of 500~training iterations for all algorithms.
In all cases, we use \textsc{afk-mc$^2$} \citep{BachemEtAl2016b} for initialization of the means.

\paragraph{Validity.}

\begin{figure}[htb]
  \begin{subfigure}[c]{0.23\textwidth}
    \begin{adjustbox}{trim=5pt 8pt 0pt 5pt}
%      \tikzset{external/remake next}
      %\tikzset{external/remake next}
\tikzsetnextfilename{BIRCH_GMM_exhaustive_QE-main}
\pgfplotsset{
	grid style={dotted,gray},
	minor grid style={dotted,lightgray},
  tick label style = {font=\tiny\sansmath\sffamily},
  legend style = {font=\sansmath\sffamily},
  xlabel style = {font=\sansmath\sffamily},
  ylabel style = {font=\sansmath\sffamily},
	% to match the colors of the markers to the plot cycle list 'exotic'
  legend image code/.code={
    \draw[mark repeat=2,mark phase=2]
    plot coordinates {
      (0cm,0cm)
      (0.25cm,0cm)        %% default is (0.3cm,0cm)
      (0.5cm,0cm)         %% default is (0.6cm,0cm)
    };%	
  }
}

\begin{tikzpicture}%[rotate=-90]
	\tikzset{mark size={1.0}}
	\begin{axis}[
  	title style={yshift=-5pt,xshift=5pt,},
    title = {\small\sffamily var-GMM-X},
		colormap access=direct,
		width = 4.6cm,
		height = 3.5cm,
    	xmin=0,
		xmax=100,
%		xtick = {0,5,...,75},
%		minor xtick = {0,10,...,500},
		scaled x ticks = false,
		xlabel={\scriptsize\sffamily \phantom{Iteration}},
		xlabel near ticks, xticklabel pos=lower,
		ymin=4000,
		ymax=12000,
%		ytick = {-9.3,-9.1,...,-8.5},
%		minor ytick = {5,10,...,95},
		ylabel={\scriptsize\sffamily Quantization Error},
		ylabel near ticks,
		yticklabel pos=left,
    y tick label style={
        /pgf/number format/.cd,
            fixed,
            fixed zerofill,
            precision=1,
        /tikz/.cd
    },
		grid = both,
   legend image code/.code={%
     \draw[solid]  (0cm, 0.025cm) -- (0.52cm, 0.025cm);
%     \draw[dashed] (0cm,0.0cm) -- (0.5cm,0.0cm);
     \draw[dotted, thick] (0cm,-0.025cm) -- (0.52cm,-0.025cm);
    },		
    legend entries={
      \hspace{-4pt}\fontsize{6}{0}\selectfont\sffamily $G=2$,
      \hspace{-4pt}\fontsize{6}{0}\selectfont\sffamily $G=5$,
      \hspace{-4pt}\fontsize{6}{0}\selectfont\sffamily $G=C=25$
    },
		legend style={
			at={(1.6,1.55)},
			legend columns=3,
%			row sep=-2pt,
			column sep=0.2cm,
      inner sep=1.5pt,
		},
		legend cell align=left,
    %reverse legend,
  	]
	\addlegendimage{color=red!70!black}
  \addlegendimage{color=blue!70!black}  	
  \addlegendimage{color=black}
%  \addlegendimage{only marks, mark=*, color=black}
  %
  
  %--- Distances C'=2 ------------------------------------------
  % Phi Error
  % Lower bound (invisible plot)
  \addplot [draw=none, stack plots=y, forget plot] table[
    x=n,
    y expr=\thisrow{phi}-\thisrow{phi_err}
  ] {./figs/BIRCH_GMM_exhaustive_2.txt};	
  
  % Stack twice the error, draw as area plot
  \addplot [draw=none, fill=red!70!black, stack plots=y, fill opacity=0.15] table [
      x=n,
      y expr=2*\thisrow{phi_err}
  ] {./figs/BIRCH_GMM_exhaustive_2.txt}\closedcycle;
  
  % Reset stack using invisible plot
  \addplot [forget plot, stack plots=y,draw=none] table [x=n, y expr=-(\thisrow{phi}+\thisrow{phi_err})] {./figs/BIRCH_GMM_exhaustive_2.txt};
     
	\addplot [solid, mark=None, color=red!70!black] table[x=n, y=phi] {./figs/BIRCH_GMM_exhaustive_2.txt};	
	\addplot [dotted, thick, mark=None, color=red!70!black] table[x=n, y=phi] {./figs/BIRCH_GMM_exhaustive_2_best.txt};	

  %--- Distances C'=3 ------------------------------------------
  % Phi Error
  % Lower bound (invisible plot)
  \addplot [draw=none, stack plots=y, forget plot] table[
    x=n,
    y expr=\thisrow{phi}-\thisrow{phi_err}
  ] {./figs/BIRCH_GMM_exhaustive_5.txt};	
  
  % Stack twice the error, draw as area plot
  \addplot [draw=none, fill=blue!70!black, stack plots=y, fill opacity=0.15] table [
      x=n,
      y expr=2*\thisrow{phi_err}
  ] {./figs/BIRCH_GMM_exhaustive_5.txt}\closedcycle;
  
  % Reset stack using invisible plot
  \addplot [forget plot, stack plots=y,draw=none] table [x=n, y expr=-(\thisrow{phi}+\thisrow{phi_err})] {./figs/BIRCH_GMM_exhaustive_5.txt};
     
	\addplot [solid, mark=None, color=blue!70!black] table[x=n, y=phi] {./figs/BIRCH_GMM_exhaustive_5.txt};	
	\addplot [dotted, thick, mark=None, color=blue!70!black] table[x=n, y=phi] {./figs/BIRCH_GMM_exhaustive_5_best.txt};	

  %--- Distances C'=25 ------------------------------------------
  % Phi Error
  % Lower bound (invisible plot)
  \addplot [draw=none, stack plots=y, forget plot] table[
    x=n,
    y expr=\thisrow{phi}-\thisrow{phi_err}
  ] {./figs/BIRCH_GMM.txt};	
  
  % Stack twice the error, draw as area plot
  \addplot [draw=none, fill=black, stack plots=y, fill opacity=0.15] table [
      x=n,
      y expr=2*\thisrow{phi_err}
  ] {./figs/BIRCH_GMM.txt}\closedcycle;
  
  % Reset stack using invisible plot
  \addplot [forget plot, stack plots=y,draw=none] table [x=n, y expr=-(\thisrow{phi}+\thisrow{phi_err})] {./figs/BIRCH_GMM.txt};
     
	\addplot [solid, mark=None, color=black] table[x=n, y=phi] {./figs/BIRCH_GMM.txt};	
	\addplot [dotted, thick, mark=None, color=black] table[x=n, y=phi] {./figs/BIRCH_GMM_best.txt};	

  \end{axis}
\end{tikzpicture}
    \end{adjustbox}
	\end{subfigure}
  \begin{subfigure}[c]{0.23\textwidth}
    \begin{adjustbox}{trim=3pt 24pt 0pt 5pt}
%      \tikzset{external/remake next}
      %\tikzset{external/remake next}
\tikzsetnextfilename{BIRCH_GMM_estimated_QE-main}
\pgfplotsset{
	grid style={dotted,gray},
	minor grid style={dotted,lightgray},
  tick label style = {font=\tiny\sansmath\sffamily},
  legend style = {font=\sansmath\sffamily},
  xlabel style = {font=\sansmath\sffamily},
  ylabel style = {font=\sansmath\sffamily},
	% to match the colors of the markers to the plot cycle list 'exotic'
  legend image code/.code={
    \draw[mark repeat=2,mark phase=2]
    plot coordinates {
      (0cm,0cm)
      (0.25cm,0cm)        %% default is (0.3cm,0cm)
      (0.5cm,0cm)         %% default is (0.6cm,0cm)
    };%	
  }
}

\begin{tikzpicture}%[rotate=-90]
	\tikzset{mark size={1.0}}
	\begin{axis}[
  	title style={yshift=-5pt,xshift=5pt,},
    title = {\small\sffamily var-GMM-S},
		colormap access=direct,
		width = 4.6cm,
		height = 3.5cm,
    xmin=0,
		xmax=100,
%		xtick = {0,5,...,75},
%		minor xtick = {0,10,...,500},
		scaled x ticks = false,
		xlabel={\scriptsize\sffamily \phantom{Iteration}},
		xlabel near ticks, xticklabel pos=lower,
		ymin=4000,
		ymax=12000,
%		ytick = {-9.3,-9.1,...,-8.5},
%		minor ytick = {5,10,...,95},
		ylabel={\scriptsize\sffamily \phantom{Quantization Error}},
		ylabel near ticks,
		yticklabel pos=left,
    y tick label style={
        /pgf/number format/.cd,
            fixed,
            fixed zerofill,
            precision=1,
        /tikz/.cd
    },
		grid = both,
		%
%   legend image code/.code={%
%     \draw[solid]  (0cm, 0.05cm) -- (0.5cm, 0.05cm);
%     \draw[dashed] (0cm,0.0cm) -- (0.5cm,0.0cm);
%     \draw[dotted] (0cm,-0.05cm) -- (0.5cm,-0.05cm);
%    },		
%    legend entries={
%      \hspace{-4pt}\fontsize{6}{0}\selectfont\sffamily $G=2$,
%      \hspace{-4pt}\fontsize{6}{0}\selectfont\sffamily $G=5$,
%      \hspace{-4pt}\fontsize{6}{0}\selectfont\sffamily $G=C=25$
%    },
%		legend style={
%			draw = none,
%			at={(1.6,1.40)},
%			anchor=north,
%			legend columns=3,
%			row sep=-2pt,
%			column sep=0.2cm,
%      inner sep=1.5pt,
%		},
%		legend cell align=left,
%    reverse legend,
  	]
%	\addlegendimage{color=red!70!black}
%  \addlegendimage{color=blue!70!black}  	
%  \addlegendimage{color=black}
%%  \addlegendimage{only marks, mark=*, color=black}
  %
  
  %--- Distances C'=2 ------------------------------------------
  % Phi Error
  % Lower bound (invisible plot)
  \addplot [draw=none, stack plots=y, forget plot] table[
    x=n,
    y expr=\thisrow{phi}-\thisrow{phi_err}
  ] {./figs/BIRCH_GMM_estimated_2.txt};	
  
  % Stack twice the error, draw as area plot
  \addplot [draw=none, fill=red!70!black, stack plots=y, fill opacity=0.15] table [
      x=n,
      y expr=2*\thisrow{phi_err}
  ] {./figs/BIRCH_GMM_estimated_2.txt}\closedcycle;
  
  % Reset stack using invisible plot
  \addplot [forget plot, stack plots=y,draw=none] table [x=n, y expr=-(\thisrow{phi}+\thisrow{phi_err})] {./figs/BIRCH_GMM_estimated_2.txt};
     
	\addplot [solid, mark=None, color=red!70!black] table[x=n, y=phi] {./figs/BIRCH_GMM_estimated_2.txt};	
	\addplot [dotted, thick, mark=None, color=red!70!black] table[x=n, y=phi] {./figs/BIRCH_GMM_estimated_2_best.txt};	

  %--- Distances C'=5 ------------------------------------------
  % Phi Error
  % Lower bound (invisible plot)
  \addplot [draw=none, stack plots=y, forget plot] table[
    x=n,
    y expr=\thisrow{phi}-\thisrow{phi_err}
  ] {./figs/BIRCH_GMM_estimated_5.txt};	
  
  % Stack twice the error, draw as area plot
  \addplot [draw=none, fill=blue!70!black, stack plots=y, fill opacity=0.15] table [
      x=n,
      y expr=2*\thisrow{phi_err}
  ] {./figs/BIRCH_GMM_estimated_5.txt}\closedcycle;
  
  % Reset stack using invisible plot
  \addplot [forget plot, stack plots=y,draw=none] table [x=n, y expr=-(\thisrow{phi}+\thisrow{phi_err})] {./figs/BIRCH_GMM_estimated_5.txt};
     
	\addplot [solid, mark=None, color=blue!70!black] table[x=n, y=phi] {./figs/BIRCH_GMM_estimated_5.txt};	
	\addplot [dotted, thick, mark=None, color=blue!70!black] table[x=n, y=phi] {./figs/BIRCH_GMM_estimated_5_best.txt};	

  %--- Distances C'=25 ------------------------------------------
  % Phi Error
  % Lower bound (invisible plot)
  \addplot [draw=none, stack plots=y, forget plot] table[
    x=n,
    y expr=\thisrow{phi}-\thisrow{phi_err}
  ] {./figs/BIRCH_GMM.txt};	
  
  % Stack twice the error, draw as area plot
  \addplot [draw=none, fill=black, stack plots=y, fill opacity=0.15] table [
      x=n,
      y expr=2*\thisrow{phi_err}
  ] {./figs/BIRCH_GMM.txt}\closedcycle;
  
  % Reset stack using invisible plot
  \addplot [forget plot, stack plots=y,draw=none] table [x=n, y expr=-(\thisrow{phi}+\thisrow{phi_err})] {./figs/BIRCH_GMM.txt};
     
	\addplot [solid, mark=None, color=black] table[x=n, y=phi] {./figs/BIRCH_GMM.txt};	
	\addplot [dotted, thick, mark=None, color=black] table[x=n, y=phi] {./figs/BIRCH_GMM_best.txt};	

  \end{axis}
\end{tikzpicture}
    \end{adjustbox}
	\end{subfigure}\\
%	\vspace{6pt}\\
	%
  \begin{subfigure}[c]{0.23\textwidth}
    \begin{adjustbox}{trim=3pt 0pt 0pt 5pt}
      \hspace{-5pt}
%      \tikzset{external/remake next}
      %\tikzset{external/remake next}
\tikzsetnextfilename{BIRCH_k-means_exhaustive_QE-main}
\pgfplotsset{
	grid style={dotted,gray},
	minor grid style={dotted,lightgray},
  tick label style = {font=\tiny\sansmath\sffamily},
  legend style = {font=\sansmath\sffamily},
  xlabel style = {font=\sansmath\sffamily},
  ylabel style = {font=\sansmath\sffamily},
	% to match the colors of the markers to the plot cycle list 'exotic'
  legend image code/.code={
    \draw[mark repeat=2,mark phase=2]
    plot coordinates {
      (0cm,0cm)
      (0.25cm,0cm)        %% default is (0.3cm,0cm)
      (0.5cm,0cm)         %% default is (0.6cm,0cm)
    };%	
  }
}

\begin{tikzpicture}%[rotate=-90]
	\tikzset{mark size={1.0}}
	\begin{axis}[
  	title style={yshift=-5pt,xshift=5pt,},
    title = {\small\sffamily var-$k$-means-X},
		colormap access=direct,
		width = 4.6cm,
		height = 3.5cm,
    	xmin=0,
		xmax=50,
		xtick = {0,10,...,50},
%		minor xtick = {0,10,...,500},
		scaled x ticks = false,
		xlabel={\scriptsize\sffamily Iteration},
		xlabel near ticks, xticklabel pos=lower,
    x label style={at={(0.5,-0.15)}},
		ymin= 4000,
		ymax = 12000,
%		ytick = {-9.3,-9.1,...,-8.5},
%		minor ytick = {5,10,...,95},
		ylabel={\scriptsize\sffamily Quantization Error},
		ylabel near ticks,
		yticklabel pos=left,
    y tick label style={
        /pgf/number format/.cd,
            fixed,
            fixed zerofill,
            precision=1,
        /tikz/.cd
    },		
		grid = both,
		%
%   legend image code/.code={%
%     \draw[dash pattern=on 0.175cm off 0.05cm on 0.05cm off 0.05cm on 0.175cm, draw=red!70!black] (0cm,0.05cm) -- (0.5cm,0.05cm);
%     \draw[dashed] (0cm,-0.025cm) -- (0.5cm,-0.025cm);
%     \draw[solid]  (0cm, 0.025cm) -- (0.5cm, 0.025cm);
%    },		
    %legend entries={
      %\fontsize{6}{0}\selectfont\sffamily \, $C'=2$,
      %\fontsize{6}{0}\selectfont\sffamily \, $C'=5$,
      %\fontsize{6}{0}\selectfont\sffamily \, $C'=25$
    %},
		%legend style={
%%			draw = none,
			%at={(0.95,0.95)},
%%			anchor=north,
			%legend columns=1,
			%row sep=-2pt,
			%%column sep=0.5cm,
      %%inner sep=1.5pt,
		%},
		%legend cell align=left,
    %%reverse legend,
  	]
	%\addlegendimage{color=red!70!black}
  %\addlegendimage{color=blue!70!black}  	
  %\addlegendimage{color=black}
%%  \addlegendimage{only marks, mark=*, color=black}
  %
  
  %--- Distances C'=2 ------------------------------------------
  % Phi Error
  % Lower bound (invisible plot)
  \addplot [draw=none, stack plots=y, forget plot] table[
    x=n,
    y expr=\thisrow{phi}-\thisrow{phi_err}
  ] {./figs/BIRCH_k-means_exhaustive_2.txt};	
  
  % Stack twice the error, draw as area plot
  \addplot [draw=none, fill=red!70!black, stack plots=y, fill opacity=0.15] table [
      x=n,
      y expr=2*\thisrow{phi_err}
  ] {./figs/BIRCH_k-means_exhaustive_2.txt}\closedcycle;
  
  % Reset stack using invisible plot
  \addplot [forget plot, stack plots=y,draw=none] table [x=n, y expr=-(\thisrow{phi}+\thisrow{phi_err})] {./figs/BIRCH_k-means_exhaustive_2.txt};
     
	\addplot [solid, mark=None, color=red!70!black] table[x=n, y=phi] {./figs/BIRCH_k-means_exhaustive_2.txt};	
	\addplot [dotted, thick, mark=None, color=red!70!black] table[x=n, y=phi] {./figs/BIRCH_k-means_exhaustive_2_best.txt};	

  %--- Distances C'=5 ------------------------------------------
  % Phi Error
  % Lower bound (invisible plot)
  \addplot [draw=none, stack plots=y, forget plot] table[
    x=n,
    y expr=\thisrow{phi}-\thisrow{phi_err}
  ] {./figs/BIRCH_k-means_exhaustive_5.txt};	
  
  % Stack twice the error, draw as area plot
  \addplot [draw=none, fill=blue!70!black, stack plots=y, fill opacity=0.15] table [
      x=n,
      y expr=2*\thisrow{phi_err}
  ] {./figs/BIRCH_k-means_exhaustive_5.txt}\closedcycle;
  
  % Reset stack using invisible plot
  \addplot [forget plot, stack plots=y,draw=none] table [x=n, y expr=-(\thisrow{phi}+\thisrow{phi_err})] {./figs/BIRCH_k-means_exhaustive_5.txt};
     
	\addplot [solid, mark=None, color=blue!70!black] table[x=n, y=phi] {./figs/BIRCH_k-means_exhaustive_5.txt};	
	\addplot [dotted, thick, mark=None, color=blue!70!black] table[x=n, y=phi] {./figs/BIRCH_k-means_exhaustive_5_best.txt};	

  %--- Distances C'=25 ------------------------------------------
  % Phi Error
  % Lower bound (invisible plot)
  \addplot [draw=none, stack plots=y, forget plot] table[
    x=n,
    y expr=\thisrow{phi}-\thisrow{phi_err}
  ] {./figs/BIRCH_k-means.txt};	
  
  % Stack twice the error, draw as area plot
  \addplot [draw=none, fill=black, stack plots=y, fill opacity=0.15] table [
      x=n,
      y expr=2*\thisrow{phi_err}
  ] {./figs/BIRCH_k-means.txt}\closedcycle;
  
  % Reset stack using invisible plot
  \addplot [forget plot, stack plots=y,draw=none] table [x=n, y expr=-(\thisrow{phi}+\thisrow{phi_err})] {./figs/BIRCH_k-means.txt};
     
	\addplot [solid, mark=None, color=black] table[x=n, y=phi] {./figs/BIRCH_k-means.txt};	
	\addplot [dotted, thick, mark=None, color=black] table[x=n, y=phi] {./figs/BIRCH_k-means_best.txt};	

  \end{axis}
\end{tikzpicture}
    \end{adjustbox}
	\end{subfigure}
  \begin{subfigure}[c]{0.23\textwidth}
    \begin{adjustbox}{trim=3pt 0pt 0pt 5pt}
      \hspace{-5pt}
%      \tikzset{external/remake next}
      %\tikzset{external/remake next}
\tikzsetnextfilename{BIRCH_k-means_estimated_QE-main}
\pgfplotsset{
	grid style={dotted,gray},
	minor grid style={dotted,lightgray},
  tick label style = {font=\tiny\sansmath\sffamily},
  legend style = {font=\sansmath\sffamily},
  xlabel style = {font=\sansmath\sffamily},
  ylabel style = {font=\sansmath\sffamily},
	% to match the colors of the markers to the plot cycle list 'exotic'
  legend image code/.code={
    \draw[mark repeat=2,mark phase=2]
    plot coordinates {
      (0cm,0cm)
      (0.25cm,0cm)        %% default is (0.3cm,0cm)
      (0.5cm,0cm)         %% default is (0.6cm,0cm)
    };%	
  }
}

\begin{tikzpicture}%[rotate=-90]
	\tikzset{mark size={1.0}}
	\begin{axis}[
  	title style={yshift=-5pt,xshift=5pt,},
    title = {\small\sffamily var-$k$-means-S},
		colormap access=direct,
		width = 4.6cm,
		height = 3.5cm,
    xmin=0,
		xmax=50,
		xtick = {0,10,...,50},
%		minor xtick = {0,10,...,500},
		scaled x ticks = false,
		xlabel={\scriptsize\sffamily Iteration},
		xlabel near ticks, xticklabel pos=lower,
    x label style={at={(0.5,-0.15)}},
		ymin= 4000,
		ymax = 12000,
%		ytick = {-9.3,-9.1,...,-8.5},
%		minor ytick = {5,10,...,95},
		ylabel={\scriptsize\sffamily \phantom{Quantization Error}},
		ylabel near ticks,
		yticklabel pos=left,
    y tick label style={
        /pgf/number format/.cd,
            fixed,
            fixed zerofill,
            precision=1,
        /tikz/.cd
    },
		grid = both,
		%
%   legend image code/.code={%
%     \draw[dash pattern=on 0.175cm off 0.05cm on 0.05cm off 0.05cm on 0.175cm, draw=red!70!black] (0cm,0.05cm) -- (0.5cm,0.05cm);
%     \draw[dashed] (0cm,-0.025cm) -- (0.5cm,-0.025cm);
%     \draw[solid]  (0cm, 0.025cm) -- (0.5cm, 0.025cm);
%    },		
    %legend entries={
      %%\fontsize{6}{0}\selectfont\sffamily \, $C'=2$,
      %\fontsize{6}{0}\selectfont\sffamily \, $C'=5$,
      %\fontsize{6}{0}\selectfont\sffamily \, $C'=25$
    %},
		%legend style={
%%			draw = none,
			%at={(0.95,0.95)},
%%			anchor=north,
			%legend columns=1,
			%row sep=-2pt,
			%%column sep=0.5cm,
      %%inner sep=1.5pt,
		%},
		%legend cell align=left,
    %%reverse legend,
  	]
	%%\addlegendimage{color=red!70!black}
  %\addlegendimage{color=blue!70!black}  	
  %\addlegendimage{color=black}
%%  \addlegendimage{only marks, mark=*, color=black}
  %
  
%  %--- Distances C'=3 ------------------------------------------
%  % Phi Error
%  % Lower bound (invisible plot)
%  \addplot [draw=none, stack plots=y, forget plot] table[
%    x=n,
%    y expr=\thisrow{phi}-\thisrow{phi_err}
%  ] {./figs/BIRCH_k-means_estimated_2.txt};	
%  
%  % Stack twice the error, draw as area plot
%  \addplot [draw=none, fill=red!70!black, stack plots=y, fill opacity=0.15] table [
%      x=n,
%      y expr=2*\thisrow{phi_err}
%  ] {./figs/BIRCH_k-means_estimated_2.txt}\closedcycle;
%  
%  % Reset stack using invisible plot
%  \addplot [forget plot, stack plots=y,draw=none] table [x=n, y expr=-(\thisrow{phi}+\thisrow{phi_err})] {./figs/BIRCH_k-means_estimated_2.txt};
%     
%	\addplot [solid, mark=None, color=red!70!black] table[x=n, y=phi] {./figs/BIRCH_k-means_estimated_2.txt};	

  %--- Distances C'=5 ------------------------------------------
  % Phi Error
  % Lower bound (invisible plot)
  \addplot [draw=none, stack plots=y, forget plot] table[
    x=n,
    y expr=\thisrow{phi}-\thisrow{phi_err}
  ] {./figs/BIRCH_k-means_estimated_5.txt};	
  
  % Stack twice the error, draw as area plot
  \addplot [draw=none, fill=blue!70!black, stack plots=y, fill opacity=0.15] table [
      x=n,
      y expr=2*\thisrow{phi_err}
  ] {./figs/BIRCH_k-means_estimated_5.txt}\closedcycle;
  
  % Reset stack using invisible plot
  \addplot [forget plot, stack plots=y,draw=none] table [x=n, y expr=-(\thisrow{phi}+\thisrow{phi_err})] {./figs/BIRCH_k-means_estimated_5.txt};
     
	\addplot [solid, mark=None, color=blue!70!black] table[x=n, y=phi] {./figs/BIRCH_k-means_estimated_5.txt};	
	\addplot [dotted, thick, mark=None, color=blue!70!black] table[x=n, y=phi] {./figs/BIRCH_k-means_estimated_5_best.txt};	

  %--- Distances C'=25 ------------------------------------------
  % Phi Error
  % Lower bound (invisible plot)
  \addplot [draw=none, stack plots=y, forget plot] table[
    x=n,
    y expr=\thisrow{phi}-\thisrow{phi_err}
  ] {./figs/BIRCH_k-means.txt};	
  
  % Stack twice the error, draw as area plot
  \addplot [draw=none, fill=black, stack plots=y, fill opacity=0.15] table [
      x=n,
      y expr=2*\thisrow{phi_err}
  ] {./figs/BIRCH_k-means.txt}\closedcycle;
  
  % Reset stack using invisible plot
  \addplot [forget plot, stack plots=y,draw=none] table [x=n, y expr=-(\thisrow{phi}+\thisrow{phi_err})] {./figs/BIRCH_k-means.txt};
     
	\addplot [solid, mark=None, color=black] table[x=n, y=phi] {./figs/BIRCH_k-means.txt};	
	\addplot [dotted, thick, mark=None, color=black] table[x=n, y=phi] {./figs/BIRCH_k-means_best.txt};	

  \end{axis}
\end{tikzpicture}
    \end{adjustbox}
	\end{subfigure}
\vspace{-8pt}\ \\
  \caption{
Results of \cref{alg:var-GMM-X,alg:var-GMM-S} on the BIRCH~$5\times 5$ data set.
The mean quantization error (solid), shaded with its SEM, as well as the single run with lowest final quantization error (dotted) over 100~independent training runs is given.
Only in the extreme case of var-$k$-means-S with $C'=1$ and $G=2$ the distance estimation fails to recover viable results.}
  \label{fig:BIRCH5x5}
  \vspace{1pt}
\end{figure}
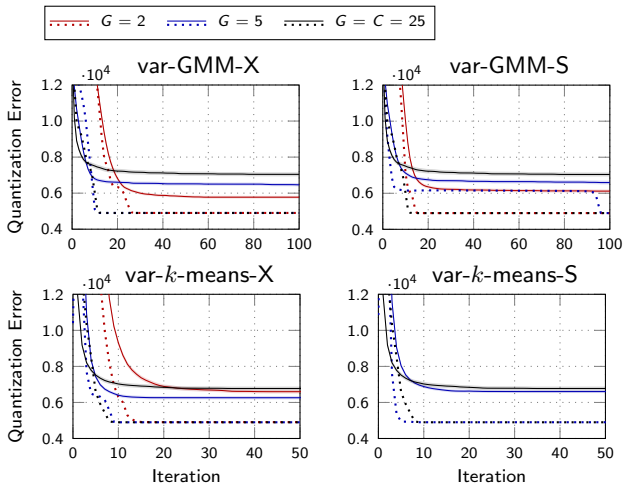

We first consider a small scale $5 \times 5$ BIRCH data set as described above to compare the partial variational algorithms with their respective non-variational counterparts.
Considering \cref{fig:BIRCH5x5}, we note that initially var-GMM and var-$k$-means require more EM iterations than standard GMM or $k$-means to obtain comparable quantization errors.
The primary reason for this is the random $\KKn$ and $\GGn$ initialization, which requires a couple of iterations until these reflect the true neighborhood of a data point.
The number of additional EM iterations is however relatively small, and only more significant for very low values of $G$ (small cluster neighborhoods).
Differences between exhaustive cluster neighborhoods (var-GMM-X and var-$k$-means-X) and estimated neighborhoods (var-GMM-S and var-$k$-means-S) are only observable for very low $G$, which verifies that the estimation of cluster-to-cluster distances does not negatively affect performance at least not above such very low values.
We will therefore from here on focus on the two most run-time efficient algorithms, var-GMM-S and var-$k$-means-S, which scale with $\OO(NG^2D)$ and $\OO(NGD)$, respectively.

%\vspace{3pt}

\paragraph{Neighborhood initialization and exploratory neighbors.}
For large scale data with many clusters in low-dimensional spaces, the random initialization of cluster neighborhoods can be arbitrarily bad and can take many iterations until better neighborhoods are found.
Performing M-steps during this initial period can have unwanted, destructive effects.
For large-scale experiments we therefore allow for a couple of initial E-steps (optimized on the training quantization error) to find better initial neighborhood relations $\KKn$, $\GGc$.
Furthermore, allowing for a single extra randomly picked exploratory cluster per data point within $\GGn$ helps to overcome large gaps between areas of dense clusters.

\begin{figure}[tb]
  \begin{adjustbox}{trim=8pt 6pt 0pt 5pt}
    %\tikzset{external/remake next}
\tikzsetnextfilename{BIRCH-Iterations-C}

\pgfplotsset{
	grid style={dotted,gray},
	minor grid style={dotted,lightgray},
  tick label style = {font=\scriptsize\sansmath\sffamily},
  legend style = {font=\sansmath\sffamily},
	/tikz/font=\sansmath\sffamily,
  xlabel style = {font=\sansmath\sffamily},
  ylabel style = {font=\sansmath\sffamily},
	% to match the colors of the markers to the plot cycle list 'exotic'
}	

\begin{tikzpicture}%[rotate=-90]
	\tikzset{mark size={1.0}}
	% background
	\begin{axis}[
		colormap access=direct,
		width = 7.75cm,
		height = 5.5cm,
		xmin=64,
		xmax=4096,
		xmode=log,
		log basis x={2},
    axis x line=none,
		ymin = 8,
		ymax = 512,
		ymode=log,
	  log basis y={2},
    axis y line=none,
		]
    \newcommand\bgone{gray!3!white}
    \newcommand\bgtwo{gray!8!white}	
    \addplot+ [solid, thin, mark=none, draw=\bgone, fill=\bgone, domain={64:4096}] {256/64*x} \closedcycle;
    \addplot+ [solid, thin, mark=none, draw=\bgtwo, fill=\bgtwo, domain={64:4096}] {128/64*x} \closedcycle;
    \addplot+ [solid, thin, mark=none, draw=\bgone, fill=\bgone, domain={64:4096}] {64/64*x} \closedcycle;
    \addplot+ [solid, thin, mark=none, draw=\bgtwo, fill=\bgtwo, domain={64:4096}] {32/64*x} \closedcycle;
    \addplot+ [solid, thin, mark=none, draw=\bgone, fill=\bgone, domain={64:4096}] {16/64*x} \closedcycle;
    \addplot+ [solid, thin, mark=none, draw=\bgtwo, fill=\bgtwo, domain={64:4096}] {8/64*x} \closedcycle;
    \addplot+ [solid, thin, mark=none, draw=\bgone, fill=\bgone, domain={64:4096}] {8/128*x} \closedcycle;
    \addplot+ [solid, thin, mark=none, draw=\bgtwo, fill=\bgtwo, domain={64:4096}] {8/256*x} \closedcycle;
    \addplot+ [solid, thin, mark=none, draw=\bgone, fill=\bgone, domain={64:4096}] {8/512*x} \closedcycle;
    \addplot+ [solid, thin, mark=none, draw=\bgtwo, fill=\bgtwo, domain={64:4096}] {8/1024*x} \closedcycle;
    \addplot+ [solid, thin, mark=none, draw=\bgone, fill=\bgone, domain={64:4096}] {8/2048*x} \closedcycle;
  ];
  \end{axis} 
  %
  % left axis
	\begin{axis}[
		colormap access=direct,
		width = 7.75cm,
		height = 5.5cm,
		xmin=64,
		xmax=4096,
		xmode=log,
		log basis x={2},
		log ticks with fixed point,
		xtick = {64,128,256,512,1024,2048,4096},
		xticklabels = {64,128,256,512,1024,2048,4096},
		scaled x ticks = false,
%		minor xtick = {0,10,...,500},
		xlabel={\fontsize{7}{0}\selectfont\sffamily Number of clusters $C$},
		ymin = 8,
		ymax = 512,
		ymode=log,
	  log basis y={2},
		scaled y ticks = false,
		axis y line*=left,
		ytick = {16,32,64,128,256,512},
		yticklabels = {16,32,64,128,256,512},
%		minor ytick = {5,10,...,95},
    ylabel style={align=center, font=\fontsize{7}{0}\selectfont\sffamily},
		ylabel={\#Iterations until the variational algorithms \\ reach $k$-means' converged quantization error},
    y label style={at={(-0.08,0.46)}},
%    axis y line*=left,
		grid = both,
		legend style={
			at={(0.635,0.95)},
			legend columns=1,
			row sep=-2.8pt,
			%column sep=0.5cm,
      inner sep=1.3pt,
		},
		legend cell align=left,
		legend entries={\hspace{-20pt}\fontsize{6}{0}\selectfont\itshape\sffamily Algs. independent of $C$ per iteration:,
                    \hspace{1pt}\fontsize{6}{0}\selectfont\sffamily var-$k$-means-S; $G=2+1$,
		                \hspace{1pt}\fontsize{6}{0}\selectfont\sffamily var-$k$-means-S; $G=5+1$,
		                \hspace{1pt}\fontsize{6}{0}\selectfont\sffamily var-GMM-S; $G=2+1$,
		                \hspace{1pt}\fontsize{6}{0}\selectfont\sffamily var-GMM-S; $G=5+1$,
		},		
		]
%    \addplot+ [dashed, thin, mark=none, draw=gray, domain={64:4096}] {8/64*x};
		\addlegendimage{empty legend};
		\addplot+ [thick, mark=x, color=blue!40!white] table[x=C, y=Nkm2] {./figs/BIRCH-Total_Iterations.txt};
		\addplot+ [thick, mark=x, color=blue!70!black] table[x=C, y=Nkm5] {./figs/BIRCH-Total_Iterations.txt};
		\addplot+ [thick, mark=x, color=red!40!white] table[x=C, y=NGMM2] {./figs/BIRCH-Total_Iterations.txt};
		\addplot+ [thick, mark=x, color=red!80!black] table[x=C, y=NGMM5] {./figs/BIRCH-Total_Iterations.txt};
		\draw[thick] (axis cs:64,170)--(axis cs:100,170);
    \filldraw (axis cs:64,170) circle (1.3pt);
  ];
  \end{axis} 
  %
  % right axis
  \begin{axis}[
		width = 7.75cm,
		height = 5.5cm,
		xmin=64,
		xmax=4096,
		xmode=log,
		log basis x={2},
    axis x line=none,
		ymin = 8,
		ymax = 512,
		ymode=log,
	  log basis y={2},
		scaled y ticks = false,
		ytick = {16,32,64,128,256,512},
		yticklabels = {16,32,64,128,256,512},
    axis y line*=right,
    ylabel style={align=center, font=\fontsize{7}{0}\selectfont\sffamily},    
    ylabel={\#Iterations until k-means converges},
    y label style={at={(1.08,0.5)}},
		legend style={
			at={(0.96,0.17)},
			legend columns=1,
			row sep=-2.8pt,
			%column sep=0.5cm,
      inner sep=1.3pt,
		},
		legend cell align=left,
		legend entries={\hspace{1pt}\fontsize{6}{0}\selectfont\sffamily standard $k$-means,
		},		
    ]
    \addplot+ [thick, dashed, mark=x, color=black!80!white] table[x=C, y=km] {./figs/BIRCH-Total_Iterations.txt};
		\draw[thick] (axis cs:3200,12.7)--(axis cs:4096,12.7);
    \filldraw (axis cs:4096,12.7) circle (1.3pt);
  \end{axis}  
  
\end{tikzpicture}
  \end{adjustbox}
  \caption{Total number of iterations (for initialization and training) until the variational algorithms or surpass the converged quantization error of standard $k$-means on BIRCH data sets of different sizes, given by the number of clusters $C$.
  The number of iterations for $k$-means until convergence is given as dashed line.
  A slope of 1 is indicated by the background for reference.
  }
  \label{fig:BIRCH-Iterations-C}
  \vspace{2pt}
\end{figure}
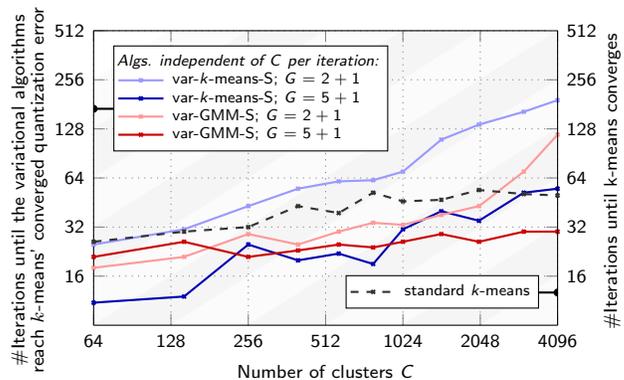

\paragraph{Absolute vs. relative complexity.}
Per iteration, the var-$k$-means and var-GMM algorithms save a significant amount of required distance evaluations.
However, if this would result in an equally higher amount of required training iterations, the absolute gain of this relative reduction would vanish again.
Already in \cref{fig:BIRCH5x5} we saw that training times will not necessarily increase (equally) with $G < C$.
We now further investigate this behavior on larger data sets by using systematically increasing sizes of BIRCH.
As measure for training time we count the initial E-step iterations (see previous paragraph) and following EM iterations until the mean quantization error of var-$k$-means-S and var-GMM-S surpasses the mean converged quantization error of standard $k$-means. %(evaluation of standard GMMs as baseline for var-GMM-S was not computationally feasible for the large scale experiments).
For both algorithms, we use $G=2$ as well as $G=5$, and for each $G$ an additional randomly picked cluster in $\GGn$ (as described before).
The compared means were taken over 5~independent training runs for each setting.
As seen in \cref{fig:BIRCH-Iterations-C}, the reduction of distance evaluations per iteration in the variational algorithms does here not result in an equally increasing number of necessary iterations with respect to $k$-means.
In most cases even less iterations were necessary than for $k$-means while the same quantization error was reached (all plots below the black dashed line).
Importantly, the number of iterations increases less than linearly with $C$ (which would in the log-log plot for small offsets result in linear graphs with slopes of $1$, parallel to the lines of the background shading).

In all here investigated cases, the quantization error of var-$k$-means-S and var-GMM-S improved on standard $k$-means.
Similar observations of better avoidance of local optima through truncated variational approaches were reported e.g.\ by \citet{ForsterLucke2017} for Poisson Mixtures and by \citet{LuckeForster2017} for GMMs (however on a much smaller scale). %ADD HUGHES AND SUDDERTH?
The optimal degree of truncation is thereby primarily dependent on the geometric structure in the data.
However, using smaller than optimal neighborhoods can still lead to viable results with potentially huge computational savings for large scale data.

\paragraph{Performance on large scale and natural data.}
The BIRCH data set is well suited to controllably increase the number of clusters within the data without changing their overall geometric structure.
However, the low dimensionality, high regularity and good cluster separation within this data set does not necessarily reflect natural data very well.
Real world practical implications are therefore demonstrated on two additional, large scale, high-dimensional, natural data sets: KDD and SONG.
%We allow for 500~training iterations on both data sets for all algorithms and report the means over 5~independent runs.
As for the BIRCH data sets we again also for SONG use initial E-steps and one single additional randomly selected cluster in $\GGn$ (denoted by the `$+1$' for $G$), and we now report mean quantization errors over 5~independent runs.
\Cref{tab:speedup} shows the trade-off between computational gain and quantization error for these two natural as well as high scale BIRCH data sets.
Especially where the number of clusters is very large, var-$k$-means shows significant speedups of two to three orders of magnitude, reaching comparable quantization errors on the natural data sets and even improving on $k$-means on the BIRCH data sets.
On the other hand, var-GMM-S consistently improves on the $k$-means baseline for $G \geq 5$ on the natural data sets, and in all settings on BIRCH.
Furthermore, while results on KDD for the full GMM were still obtainable with high computational effort (see \cref{app:DetailedResults} for results), evaluation of GMMs on SONG was only possible due to the significant computational savings of the var-GMM-S algorithm of up to two orders of magnitude.

\begin{table}[bt]
  \caption{
Relative speedup in terms of saved distance evaluations per iteration  with respect to the respective non-variational variants and relative quantization error with respect to standard $k$-means for both algorithms (as evaluation with full GMMs was computationally not feasible for these large data sets).
For var-GMM-S both the theoretical minimal (first value) as well as the experimentally measured (second value) speedup is given, as in praxis overlaps of cluster neighborhoods $\GGc$ can reduce the number of distance evaluations significantly.}
  \vspace{-7pt}
  {
\fontsize{8}{10} \sffamily \sansmath
\renewcommand{\arraystretch}{1.3}
\newcommand{\midspace}{\hphantom{abc}}
\setlength{\tabcolsep}{3pt}
\scriptsize
\begin{tabular}{lrrr r @{\hspace{3pt}}r @{\hspace{11pt}}r@{}c@{}l @{\hspace{3pt}}r}
\toprule
& & & & \multicolumn{2}{@{}c}{var-$k$-means-S}  & \multicolumn{4}{@{}c}{var-GMM-S}  \\
data set & D & C & G &  \multicolumn{1}{c}{speedup}  &  \multicolumn{1}{@{}c}{q.\ error} &  \multicolumn{3}{@{}c}{speedup}  &  q.\ error \\
\midrule
KDD & $74$ & $200$ & $2$  &  $100 \times$ & $+101.8\%$ & $50 \times$ &/& $55 \times$ & $+4.2\%$ \\[-2pt]
\multicolumn{3}{l}{$(N=145\,751)$} & $5$  &   $40 \times$ & $+5.8\%$ & $8 \times$ &/& $15 \times$ & $-0.0\%$ \\[-2pt]
    & &  & $20$ &   $10 \times$ & $+0.3\%$ & $1 \times$ &/& $3.3 \times$ & $-0.0\%$ \\[2pt]
SONG & $90$ & $2000$ & $2+1$  &  $667 \times$ & $+1.0\%$ & $400 \times$  &/& $412 \times$ & $+1.8\%$ \\[-2pt]
\multicolumn{3}{l}{$(N=515\,345)$} & $5+1$  &  $333 \times$  & $+0.5\%$ & $77 \times$ &/& $105 \times$ & $-0.2\%$ \\[-2pt]
     & & & $10+1$ &  $182 \times$  & $+0.2\%$ & $20 \times$ &/& $40 \times$ & $-0.3\%$ \\[-2pt]
     & & & $20+1$ &  $95 \times$  & $+0.1\%$ & $5 \times$    &/& $15 \times$ & $-0.3\%$ \\[-2pt]
     & & & $50+1$ &  $39 \times$   & $+0.0\%$ & % $1 \times$    &/& $6.4 \times$ & - 
     \\[2pt]
BIRCH & $2$ & $2025$ & $2+1$ & $675 \times$  & $-2.8\%$ & $405 \times$ &/& $458 \times$ & $-4.6\%$ \\[-2pt]
\multicolumn{3}{l}{$N=(202\,500)$} & $5+1$ & $338\times$ & $-4.3\%$ & $78 \times$  &/& $143 \times$ & $-9.1\%$ \\[2pt]
BIRCH & $2$ & $4096$ & $2+1$ & $1365 \times$  & $-3.7\%$ & $819 \times$ &/& $927 \times$ & $-4.4\%$ \\[-2pt]
\multicolumn{3}{l}{$(N=409\,600)$} & $5+1$ & $683 \times$ & $-4.0\%$  & $158 \times$ &/& $287 \times$ & $ -11.7\%$ \\[2pt]
\bottomrule
\end{tabular}
\setlength{\tabcolsep}{6pt}
}
  \label{tab:speedup}
%  \vspace{2pt}
\end{table}

\section{Discussion}
Using partial variational EM, we derived efficient clustering algorithms which require less than $\OO(NCD)$ numerical operations per iteration.
Most notably, algorithm var-GMM-S reduces the number of required distance evaluations per iteration from $\OO(NC)$ to $\OO(NG^2)$ where $G$ is a small constant which relates to the cluster neighborhood relationship.
\Cref{fig:BIRCH-Iterations-C} shows that the clustering objective is effectively and efficiently increased within the same number of iterations as $k$-means even if $G$ is kept at a constant value while $C$ is increased.
%Large $C$ (we used up to $C=4096$) require more care for initialization but the computational effort for initialization remains small compared to the EM iteration.
%
The strong reduction of complexity per iteration with a sublinear scaling of required iterations with $C$ (e.g.\,\cref{fig:BIRCH-Iterations-C}) is evidence for clustering being scalable sublinearly with $C$.
For real data with many clusters, the complexity reduction amounted to a reduction of distance evaluations by two to almost three orders of magnitude (\cref{tab:speedup}) compared to $k$-means, while comparable (or better) clustering results were obtained.
And distance evaluations dominate the computational demand (see \cref{app:ComplexityAnalysis}). 
%
%The independence of $C$ per iteration with no or no significant increase of the number of iterations with $C$ is evidence for clustering being scalable sublinearly with $C$. 
%
But in what sense can clustering feature such a sublinear scaling? In one sense, all clustering algorithms scale
with $C$ as implicitly we demand $N>C$, and only a value of $N$ several times larger than $C$ is sensible \citep[but see][]{KlamiJitta2016}.
A sublinear dependency on $C$ may also be perceived as counter-intuitive because we want to update each of the $C$ clusters in one iteration.
Sublinear scaling with $C$, therefore, has to be interpreted as a sublinear scaling with $NC$. In this case, our reduction to $\OO(NG^2)$ (or $\OO(NG)$)
distance evaluations per iteration still provides sufficient information (in the form of non-zero $\scn$) to update all clusters because $NG$ or $NG^2$ is greater than $C$ (typically several times). 
The sublinearity suggested by our empirical results also relies on specific properties of the data, especially certain neighborhood relationships among clusters.
Data sets and initial conditions {\em can} be constructed that will result in a poor performance.
These situations can typically be obtained using well separated areas with many close-by clusters, and by deliberately not assigning any cluster center to one area.
With careful seeding (we used \citet{BachemEtAl2016b}) and for typical data sets, such situations are unlikely, however. In practice, seeding (which may
itself depend on $C$ but not necessarily on $NC$, \citet{BachemEtAl2016b}) also limits the number of EM iterations required for convergence although it remains very difficult
to obtain theoretical results for convergence times. % remain very difficult to obtain.
%
%to obtain limits on the number of iterations are typically difficult to obtain.
%
Finally, our results clearly show that memory requirement does linearly scale with $C$ (but not with $NC$), and increases for algorithms with decreasing run-time complexity (Tab.\,1).
For memory reduction and more generally, combinations with coresets may represent promising future research. Coresets are complementary as they focus on the dependence on $N$, and as
their weighted likelihood \citep[compare, e.g.,][]{LucicEtAl2017} can be treated by our approach very similarly to (\ref{EqnLikelihood}). Coresets are also available for general GMMs, and such generalizations
likewise represents future research in the context of this work.

%ADD DISCUSSION OF CURTIN
%
To summarize, we can (under the conditions discussed above) state that our empirical results suggest that a run-time scaling of clustering sublinear with its clusters is possible.
In combination with recent advances in complementary lines of research, especially seeding \citep{BachemEtAl2016b}, efficient distance evaluations \citep{Elkan2003}, or coresets \citep{HarPeledMazumdat2004,FeldmanEtAl2011,BachemEtAl2017}, the significance of such a scaling being in principle and in practice possible may be very substantial.
%
%
%

%{\footnotesize 
\subsubsection*{Acknowledgments}
%{\bf Acknowledgments.}
We acknowledge funding by DFG grants \mbox{LU 1196/5-1} (DF) and \mbox{EXC 1077/1} (JL), and thank Florian Hirschberger for very helpful feedback.
%}
% on an earlier version of this manuscript.

{
\renewcommand\baselinestretch{0.98}
\small
\bibliographystyle{abbrvnat}
%\bibliography{../../ml_bibs/cnml-all}
\bibliography{./bibliography.bib}
\renewcommand\baselinestretch{1.0}
}

\clearpage
%\ \\[5mm]
\appendix
\crefalias{section}{appendix}
\crefalias{subsection}{appendix}
\counterwithin{figure}{section}
\counterwithin{table}{section}

\belowdisplayskip=5pt
\abovedisplayskip=5pt

\noindent{\Large \bfseries Supplementary Appendix}

\section{Proof of Proposition 1}
\label{app:ProofProp1}

Here we provide for completeness the proof of Prop.\,1 following \citet{LuckeForster2017}.
First, note that the free energy \cref{EqnFreeEnergy} is maximized w.r.t.\,$\ThetaHat$ by setting $\ThetaHat=\Theta$. This can be shown \citep[][]{Lucke2016} by generalizing a similar result for full posteriors \citep[][]{NealHinton1998}.
After setting $\ThetaHat=\Theta$ the free energy $\FF(\KK,\Theta,\Theta)=:\FF(\KK,\Theta)$ can be simplified to take on the following form \citep[][]{Lucke2016}:
 \begin{equation}
\FF(\KK,\Theta)\,=\,\sum_{n} \log\!\Big(\!\sum_{c\in\KKn} p(c,\yVecN\,|\,\Theta)\Big)\,.
\label{EqnFTrunc}
 \end{equation}
Let us now repeat Prop.\,1 in more detail before we reiterate the proof: %the proposition in a bit more detail:\\

%
%mmmmmmmmmmmmmmmmm
\noindent{}{\bfseries Proposition 1}\\
Consider the GMM of \cref{EqnGMMIso} and the free energy \cref{EqnFTrunc} for 
$n=1:N$ data points $\yVec^{(n)}\in\RRR^D$.
Furthermore, consider for a fixed $n$ the replacement of a cluster $c\in\KKn$ by a cluster $\ct\not\in\KKn$.
Then the free energy $\FF(\KK,\Theta)$ increases if and only if
\begin{align}
\|\yVecN-\muVec_{\ct}\| < \|\yVecN-\muVec_{c}\|\,. \label{EqnEuclid}
\end{align}

{\bfseries Proof}\\
By considering the specific functional form of \cref{EqnFTrunc} 
%
%First consider the free energy \cref{EqnFTrunc}:
%
%\begin{align}
%\FF(\KK,\Theta) &= \sum_{n=1}^N \log\!\Big(\! \sum_{\ct\in\KKn} p(\ct,\yVecN\,|\,\Theta)\Big) 
%                &=  \sum_{n=1}^N \log\Big( \sum_{\ct\in\KKn}\disT\frac{1}{C}(2\pi\sigma^2)^{-\frac{D}{2}}\exp\big(\hspace{-0.8mm}-\frac{1}{2\sigma^2}\|{}\yVecN-\muVec_{\ct}\|{}^2\big)\Big).
% \label{EqnEuclidF}
%\end{align}
%
we can observe that the free energy is increased by replacing $c\in\KKn$ with $\ct\not\in\KKn$ if $p(\ct,\yVecN\,|\,\Theta)>p(c,\yVecN\,|\,\Theta)$.
This applies because of the summation over $c$ in \cref{EqnFTrunc} and because of the concavity of the logarithm.
Analogously, the free energy stays constant or decreases for $p(\ct,\yVecN\,|\,\Theta)\leq{}p(c,\yVecN\,|\,\Theta)$.
If we insert for the joint probability $p(\ct,\yVec\,|\,\Theta)$ the GMM (\ref{EqnGMMIso}), we obtain:
\begin{align}
p(\ct,\yVec\,|\,\Theta)=\disT\frac{1}{C}(2\pi\sigma^2)^{-\frac{D}{2}}\exp\!\big(\hspace{-0.8mm}-\frac{1}{2\sigma^2}\|{}\yVec-\muVec_{\ct}\|{}^2\big).
\end{align}
The first two factors are independent of the data point and cluster.
The criterion for an increase of the free energy can therefore be reformulated as follows:
\begin{align}
&& p(\ct,\yVec\,|\,\Theta) &>  p(c,\yVec\,|\,\Theta) \nonumber \\
\Leftrightarrow && \exp\!\big(-\frac{1}{2\sigma^2}\|\yVec-\muVec_{\ct}\|^2\big) & > \exp\!\big(-\frac{1}{2\sigma^2}\|\yVec-\muVec_{c}\|^2\big) \nonumber \\
\Leftrightarrow && -\frac{1}{2\sigma^2}\|\yVec-\muVec_{\ct}\|^2 &> -\frac{1}{2\sigma^2}\|\yVec-\muVec_{c}\|^2 \nonumber \\
\Leftrightarrow && \|\yVec-\muVec_{\ct}\| &< \|\yVec-\muVec_{c}\|\,.
\end{align}

%which is the inequality in (\ref{EqnEuclid}).\\
$\square$
%\ \\
%As can be observed the proof relies on the very compact form of the truncated free energy (\ref{EqnFTrunc}), which was itself derived in \citet{Lucke2016}.
%Prop.\,1 is non-trivial for GMMs (and other models) and 

\section{Illustration of Algorithms 1 and 2}
\label{app:IllustrationOFAlgorithms}

\begin{figure}[bht]
	\resizebox{0.235\textwidth}{!}{
		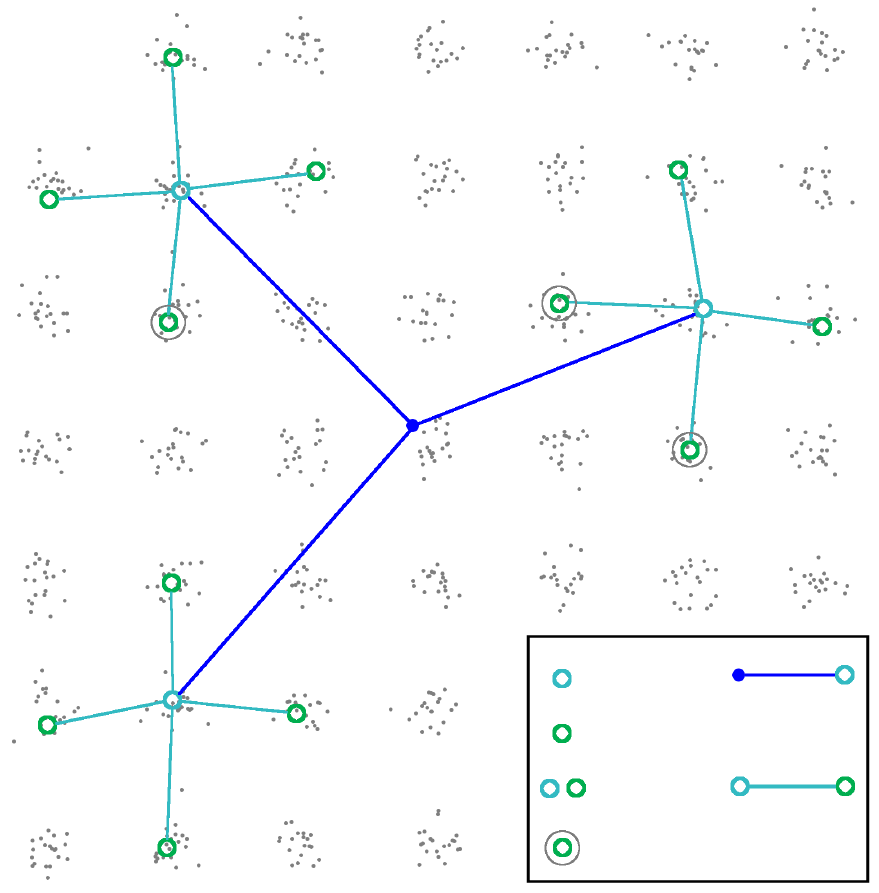
	}\hfill
	\resizebox{0.235\textwidth}{!}{
		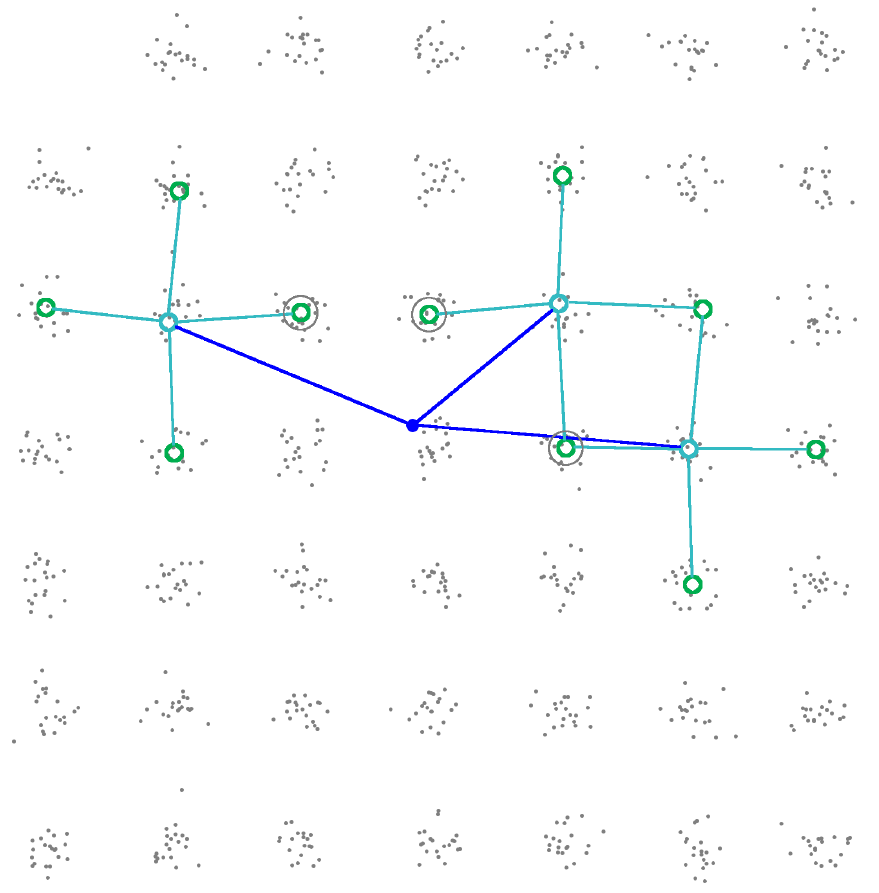
	}
	\\[4mm]
	\resizebox{0.235\textwidth}{!}{
		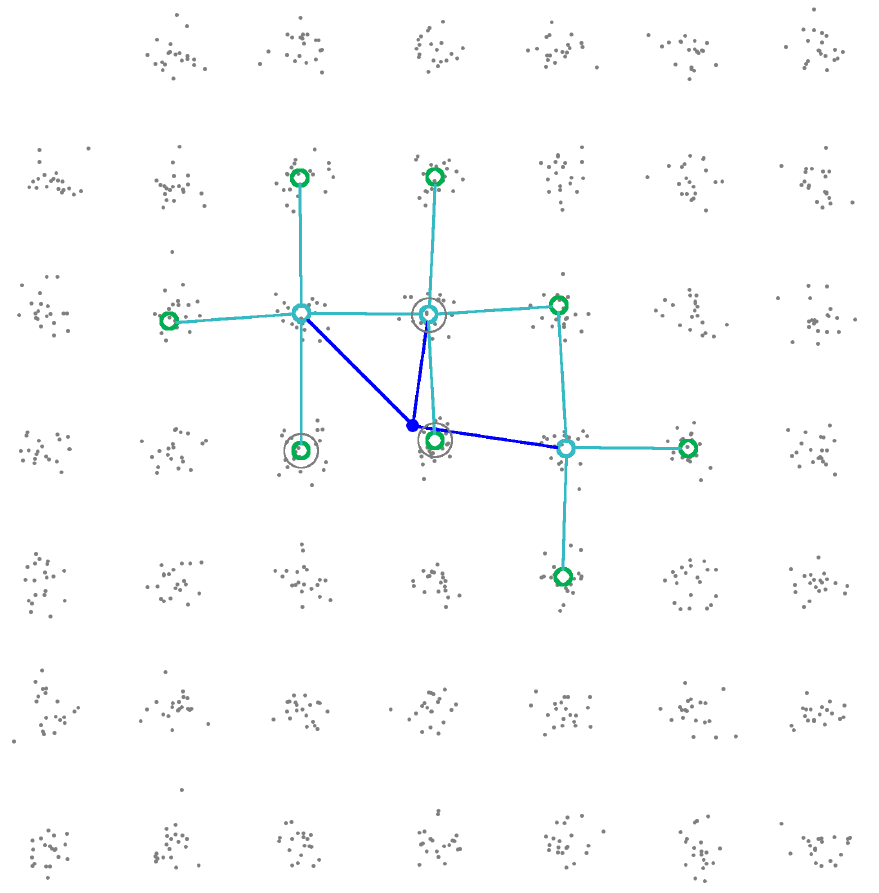
	}\hfill
	\resizebox{0.235\textwidth}{!}{
		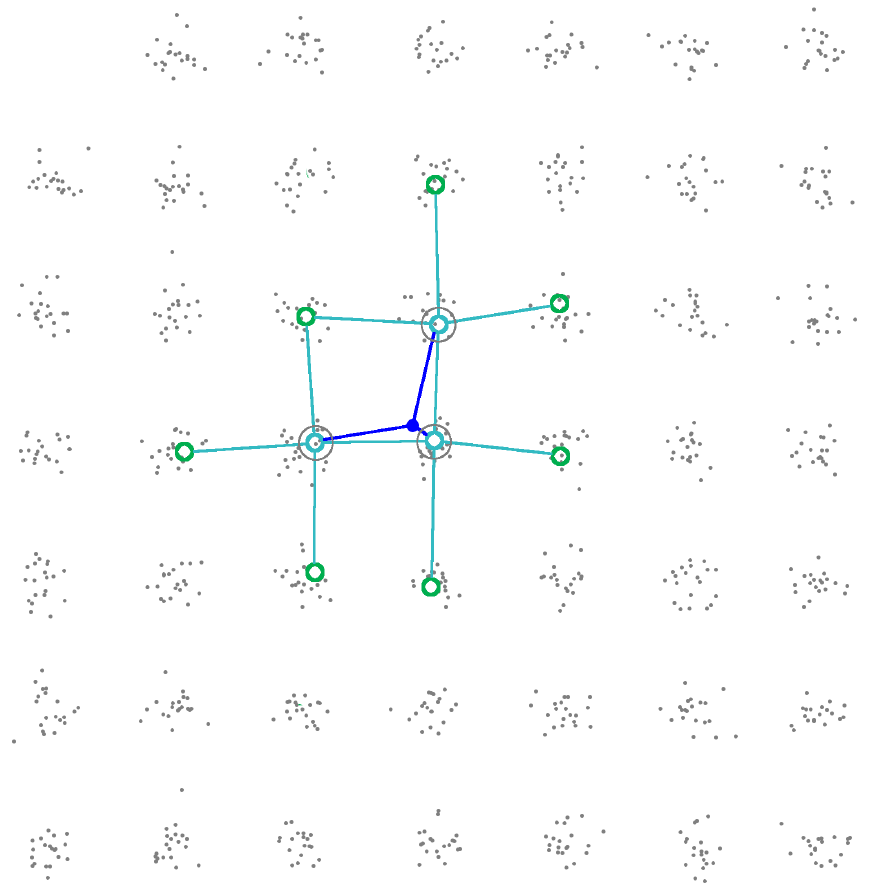
	}
  \caption{
Visualization of four E-step iterations of \cref{alg:var-GMM-X}.
In the fourth iteration the nearest clusters for $\KKn$ are found.}
\label{fig:FourIts}
\end{figure}

The four sub-figures of \cref{fig:FourIts} illustrate how the variational E-steps of \cref{alg:var-GMM-X} find clusters increasingly close to $\yVecN$.
We use $C'=3$ and $G=5$ for this example (the same as for \cref{FigVarGMM}).
For illustration purposes, we assume that good cluster centers have already been found and that they remain fixed across iterations.
{\bfseries A}~The subfigure replicates \cref{FigVarGMM}, i.e., it shows for one data point $\yVecN$ its set $\KKn$ and the sets $\GGc$ for all $c\in\KKn$.
The $\GGc$ of all clusters are computed in the first block of \cref{alg:var-GMM-X}.
The $\GGn$ are (in the second block of Alg.\,1) defined as the union over all $c\in\KKn$.
Then the distances between $\yVecN$ and all clusters in $\GGn$ (green and turquoise rings as cluster centers) are computed. 
From the clusters in $\GGn$, the $C'=3$ clusters with the smallest distances to $\yVecN$ are selected
to define the new $\KKn$ (which concludes the computations in the second block of Alg.\,1).
{\bfseries B}~Next iteration starting with the three previously selected clusters, and again showing $\KKn$ and the search space $\GGn$ for the new closest clusters.
Due to cluster neighborhood overlaps, $|\GGn|$ is here (and for the following iterations) smaller than the maximum of $C'G$.
{\bfseries C}~Third iteration.
{\bfseries D} Fourth and final iteration, any further update does not improve $\KKn$.

Note, \cref{alg:var-GMM-S} finds improved clusters for $\KKn$ similarly to \cref{alg:var-GMM-X}.
The only difference is that the sets $\GGc$ will look less ordered.
For \cref{alg:var-GMM-X} the $\GGc$ in the example of \cref{fig:FourIts} form crosses (turquoise lines)  because the algorithm always uses the four nearest clusters to construct sets $\GGc$.
Instead, \cref{alg:var-GMM-S} estimates the $\GGc$ which do hence not necessarily contain the nearest neighbors.
The $\GGc$ do therefore not look like crosses for this example but also \cref{alg:var-GMM-S} always decreases
the distances (and thus improves the free energy).

\begin{figure}[tb]
\begin{center}
  \begin{adjustbox}{trim=0pt 10pt 0pt 0pt}
  	\resizebox{0.5\textwidth}{!}{
	  	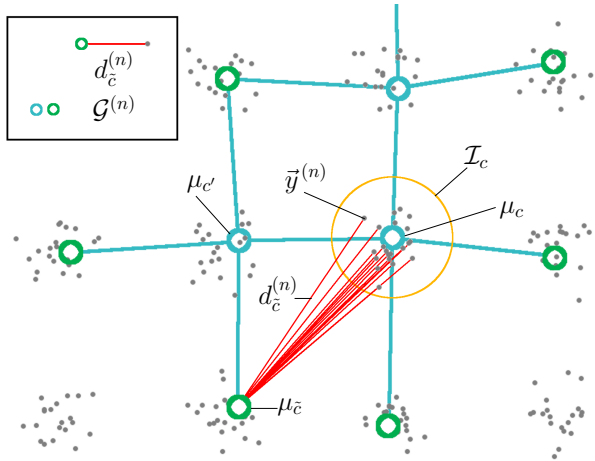
  	}
  \end{adjustbox}
\end{center}
\caption{
Visualization of the estimation of cluster-to-cluster distances $\dcct$ with already computed cluster-to-point distances $\dnct$ as used by \cref{alg:var-GMM-S} in \cref{EqnDCCEstimation}.
The mean of the read lines (distances $\dnct$ of data points closest to cluster $c$ that have cluster $\ct$ in their own neighborhood search space) gives an estimate of the true distance between cluster $c$ and $\ct$.
}
\label{FigAlgTwo}
\end{figure}

Finally, \cref{FigAlgTwo} shows an illustration of how the cluster-to-cluster distances $\dcct$ are estimated by \cref{alg:var-GMM-S}.
For the example, let us consider cluster centers and search spaces per data point $\GGn$ as in the last iteration of \cref{fig:FourIts}.
In the first computational block, \cref{alg:var-GMM-S} computes for each data point $\yVecN$ the distances $\dcn$ to all clusters in the search space $\GGn$.
From these distances, the algorithm first estimates the set of data points $\II_c$ closest to each cluster.
The illustration now shows how the distance $\dcct$ between clusters $c$ and $\ct$ is estimated using \cref{EqnDCCEstimation}. 
For cluster $c$ first the set of data points $\II_c$ is considered. For most of the points $n\in\II_c$ the distances $\dnct$ to cluster $\ct$ have been computed in the first computational block.
For instance, data point $\yVecN$ of $\II_c$ (which was used for the illustration in \cref{fig:FourIts}) has a search space $\GGn$ which includes cluster $\ct$ (we illustrate the same $\GGn$ as for \cref{fig:FourIts}D).
However, data points in the upper-right corner of $\II_c$ have search spaces which do not contain $\ct$ and are therefore not considered for the summation in \cref{EqnDCCEstimation}.
Those data points that are considered for the summation are, however, sufficient to provide a reasonable estimate for the distance $\dcct$.

For the example, we have chosen clusters $c$ and $\ct$ for which the condition in \cref{EqnDCCEstimation} results in neither trivial nor perfect estimation of $\dcct$.
If we instead consider clusters $c$ and $c'$ (see \cref{FigAlgTwo}), then distances $d^{(n)}_{c'}$ for all data points of $\II_c$ would be available, and the estimation of $d_{cc'}$ would be very accurate.
In general, the closer the clusters or the larger the $\GGn$ the better are the estimates of cluster-to-cluster distances by \cref{EqnDCCEstimation}.
For clusters very distant to each other, there may be no data points available from which the cluster-to-cluster distance could be estimated.
\Cref{alg:var-GMM-S} never estimates these distances, which essentially means that they are considered as infinite.
However, in $k$-means these clusters would likewise never contribute to their respective updates, and for the GMM their contribution would exponentially approach zero with higher distance.

%\begin{figure*}[p]
%\caption{Visualization of four E-step iterations of \cref{alg:var-GMM-X}. In the fourth iteration the nearest clusters
%for $\KKn$ are found. \cref{alg:var-GMM-S} works similarly for this data but the sets $\GGc$ are less regular (no crosses).}
%\label{FigFourIts}
%\end{figure*}

%\ \\
%\cleardoublepage

%\begin{figure}[h]
%		\input{./FigVarGMMTrunc01.pdf_t}
%\label{FigVarGMMAppOne}
%%
%\end{figure}
%%
%%
%\begin{figure}[h]%
		%\i%nput{./FigVarGMMTrunc01.pdf_t}
%\label{FigVarGMMAppTwo}
%
%\end{figure}

\section{Further Numerical Results}
\label{app:DetailedResults}

\begin{figure*}[t!]
  \begin{subfigure}[c]{0.24\textwidth}
%			\tikzset{external/remake next}
			\vspace{-4pt}
    	%\tikzset{external/remake next}
\tikzsetnextfilename{BIRCH_GMM_exhaustive_LF}
\pgfplotsset{
	grid style={dotted,gray},
	minor grid style={dotted,lightgray},
  tick label style = {font=\tiny\sansmath\sffamily},
  legend style = {font=\sansmath\sffamily},
  xlabel style = {font=\sansmath\sffamily},
  ylabel style = {font=\sansmath\sffamily},
	% to match the colors of the markers to the plot cycle list 'exotic'
  legend image code/.code={
    \draw[mark repeat=2,mark phase=2]
    plot coordinates {
      (0cm,0cm)
      (0.25cm,0cm)        %% default is (0.3cm,0cm)
      (0.5cm,0cm)         %% default is (0.6cm,0cm)
    };%	
  } 
}

\begin{tikzpicture}%[rotate=-90]
	\tikzset{mark size={1.0}}
	\begin{axis}[
		colormap access=direct,
		width = 4.9cm,
		height = 3.5cm,
    	xmin=0,
		xmax=100,
%		xtick = {0,5,...,75},
%		minor xtick = {0,10,...,500},
		scaled x ticks = false,
		xlabel near ticks, xticklabel pos=lower,
		ymin=-7.0,
		ymax = -6.0,
%		ytick = {-9.3,-9.1,...,-8.5},
%		minor ytick = {5,10,...,95},
		ylabel={\scriptsize\sffamily Log-Likelihood / Free Energy},
		ylabel near ticks,
		yticklabel pos=left,
    y label style={at={(-0.18,0.5)}},
    y tick label style={
        /pgf/number format/.cd,
            fixed,
            fixed zerofill,
            precision=1,
        /tikz/.cd
    },
		grid = both,
   legend image code/.code={%
%     \draw[dash pattern=on 0.175cm off 0.05cm on 0.05cm off 0.05cm on 0.175cm, draw=red!70!black] (0cm,0.05cm) -- (0.5cm,0.05cm);
     \draw[solid]  (0cm, 0.05cm) -- (0.52cm, 0.05cm);
     \draw[dashed] (0cm,0.0cm) -- (0.52cm,0.0cm);
     \draw[dotted, thick] (0cm,-0.05cm) -- (0.52cm,-0.05cm);
    },		
    legend entries={
      \hspace{-4pt}\fontsize{6}{0}\selectfont\sffamily $G=2$,
      \hspace{-4pt}\fontsize{6}{0}\selectfont\sffamily $G=5$,
      \hspace{-4pt}\fontsize{6}{0}\selectfont\sffamily $G=C=25$
    },
		legend style={
			at={(1.45,1.35)},
			legend columns=3,
%			row sep=-2pt,
			column sep=0.2cm,
      inner sep=1.5pt,
		},
		legend cell align=left,
    %reverse legend,
  	]
	\addlegendimage{color=red!70!black}
  \addlegendimage{color=blue!70!black}  	
  \addlegendimage{color=black}
%  \addlegendimage{only marks, mark=*, color=black}

  %--- Distances C'=2 ------------------------------------------
  % Likelihood Error
  % Lower bound (invisible plot)
  \addplot [draw=none, stack plots=y, forget plot] table[
    x=n,
    y expr=\thisrow{F}-\thisrow{F_err}
  ] {./figs/BIRCH_GMM_exhaustive_2.txt};	
  
  % Stack twice the error, draw as area plot
  \addplot [draw=none, fill=red!70!black, stack plots=y, fill opacity=0.15] table [
      x=n,
      y expr=2*\thisrow{F_err}
  ] {./figs/BIRCH_GMM_exhaustive_2.txt}\closedcycle;
  
  % Reset stack using invisible plot
  \addplot [forget plot, stack plots=y,draw=none] table [x=n, y expr=-(\thisrow{F}+\thisrow{F_err})] {./figs/BIRCH_GMM_exhaustive_2.txt};
    
  % Free Energy Error
  % Lower bound (invisible plot)
  \addplot [draw=none, stack plots=y, forget plot] table[
    x=n,
    y expr=\thisrow{L}-\thisrow{L_err}
  ] {./figs/BIRCH_GMM_exhaustive_2.txt};	
  
  % Stack twice the error, draw as area plot
  \addplot [draw=none, fill=red!70!black, stack plots=y, fill opacity=0.15] table [
      x=n,
      y expr=2*\thisrow{L_err}
  ] {./figs/BIRCH_GMM_exhaustive_2.txt}\closedcycle;
  
  % Reset stack using invisible plot
  \addplot [forget plot, stack plots=y,draw=none] table [x=n, y expr=-(\thisrow{L}+\thisrow{L_err})] {./figs/BIRCH_GMM_exhaustive_2.txt};

	\addplot [solid, mark=None, color=red!70!black] table[x=n, y=L] {./figs/BIRCH_GMM_exhaustive_2.txt};	
	\addplot [dashed, mark=None, color=red!70!black] table[x=n, y=F] {./figs/BIRCH_GMM_exhaustive_2.txt};

  %--- Distances C'=3 ------------------------------------------
  % Likelihood Error
  % Lower bound (invisible plot)
  \addplot [draw=none, stack plots=y, forget plot] table[
    x=n,
    y expr=\thisrow{F}-\thisrow{F_err}
  ] {./figs/BIRCH_GMM_exhaustive_5.txt};	
  
  % Stack twice the error, draw as area plot
  \addplot [draw=none, fill=blue!70!black, stack plots=y, fill opacity=0.15] table [
      x=n,
      y expr=2*\thisrow{F_err}
  ] {./figs/BIRCH_GMM_exhaustive_5.txt}\closedcycle;
  
  % Reset stack using invisible plot
  \addplot [forget plot, stack plots=y,draw=none] table [x=n, y expr=-(\thisrow{F}+\thisrow{F_err})] {./figs/BIRCH_GMM_exhaustive_5.txt};
    
  % Free Energy Error
  % Lower bound (invisible plot)
  \addplot [draw=none, stack plots=y, forget plot] table[
    x=n,
    y expr=\thisrow{L}-\thisrow{L_err}
  ] {./figs/BIRCH_GMM_exhaustive_5.txt};	
  
  % Stack twice the error, draw as area plot
  \addplot [draw=none, fill=blue!70!black, stack plots=y, fill opacity=0.15] table [
      x=n,
      y expr=2*\thisrow{L_err}
  ] {./figs/BIRCH_GMM_exhaustive_5.txt}\closedcycle;
  
  % Reset stack using invisible plot
  \addplot [forget plot, stack plots=y,draw=none] table [x=n, y expr=-(\thisrow{L}+\thisrow{L_err})] {./figs/BIRCH_GMM_exhaustive_5.txt};

	\addplot [solid, mark=None, color=blue!70!black] table[x=n, y=L] {./figs/BIRCH_GMM_exhaustive_5.txt};	
	\addplot [dashed, mark=None, color=blue!70!black] table[x=n, y=F] {./figs/BIRCH_GMM_exhaustive_5.txt};

  %--- Distances C'=25 ------------------------------------------
  % Likelihood Error
  % Lower bound (invisible plot)
  \addplot [draw=none, stack plots=y, forget plot] table[
    x=n,
    y expr=\thisrow{F}-\thisrow{F_err}
  ] {./figs/BIRCH_GMM.txt};	
  
  % Stack twice the error, draw as area plot
  \addplot [draw=none, fill=black, stack plots=y, fill opacity=0.15] table [
      x=n,
      y expr=2*\thisrow{F_err}
  ] {./figs/BIRCH_GMM.txt}\closedcycle;
  
  % Reset stack using invisible plot
  \addplot [forget plot, stack plots=y,draw=none] table [x=n, y expr=-(\thisrow{F}+\thisrow{F_err})] {./figs/BIRCH_GMM.txt};
    
  % Free Energy Error
  % Lower bound (invisible plot)
  \addplot [draw=none, stack plots=y, forget plot] table[
    x=n,
    y expr=\thisrow{L}-\thisrow{L_err}
  ] {./figs/BIRCH_GMM.txt};	
  
  % Stack twice the error, draw as area plot
  \addplot [draw=none, fill=black, stack plots=y, fill opacity=0.15] table [
      x=n,
      y expr=2*\thisrow{L_err}
  ] {./figs/BIRCH_GMM.txt}\closedcycle;
  
  % Reset stack using invisible plot
  \addplot [forget plot, stack plots=y,draw=none] table [x=n, y expr=-(\thisrow{L}+\thisrow{L_err})] {./figs/BIRCH_GMM.txt};

	\addplot [solid, mark=None, color=black] table[x=n, y=L] {./figs/BIRCH_GMM.txt};	
	\addplot [dashed, mark=None, color=black] table[x=n, y=F] {./figs/BIRCH_GMM.txt};

  \end{axis}
\end{tikzpicture}
	\end{subfigure}
  \begin{subfigure}[c]{0.24\textwidth}
%			\tikzset{external/remake next}
    	%\tikzset{external/remake next}
\tikzsetnextfilename{BIRCH_GMM_estimated_LF}
\pgfplotsset{
	grid style={dotted,gray},
	minor grid style={dotted,lightgray},
  tick label style = {font=\tiny\sansmath\sffamily},
  legend style = {font=\sansmath\sffamily},
  xlabel style = {font=\sansmath\sffamily},
  ylabel style = {font=\sansmath\sffamily},
	% to match the colors of the markers to the plot cycle list 'exotic'
  legend image code/.code={
    \draw[mark repeat=2,mark phase=2]
    plot coordinates {
      (0cm,0cm)
      (0.25cm,0cm)        %% default is (0.3cm,0cm)
      (0.5cm,0cm)         %% default is (0.6cm,0cm)
    };%	
  }
}

\begin{tikzpicture}%[rotate=-90]
	\tikzset{mark size={1.0}}
	\begin{axis}[
		colormap access=direct,
		width = 4.9cm,
		height = 3.5cm,
    xmin=0,
		xmax=100,
%		xtick = {0,5,...,75},
%		minor xtick = {0,10,...,500},
		scaled x ticks = false,
		xlabel near ticks, xticklabel pos=lower,
		ymin=-7.0,
		ymax = -6.0,
%		ytick = {-9.3,-9.1,...,-8.5},
%		minor ytick = {5,10,...,95},
		ylabel={\scriptsize\sffamily \phantom{Likelihood / Free Energy}},
		ylabel near ticks,
		yticklabel pos=left,
    y tick label style={
        /pgf/number format/.cd,
            fixed,
            fixed zerofill,
            precision=1,
        /tikz/.cd
    },
		grid = both,
		%
   %legend image code/.code={%
%%     \draw[dash pattern=on 0.175cm off 0.05cm on 0.05cm off 0.05cm on 0.175cm, draw=red!70!black] (0cm,0.05cm) -- (0.5cm,0.05cm);
     %\draw[dashed] (0cm,-0.025cm) -- (0.5cm,-0.025cm);
     %\draw[solid]  (0cm, 0.025cm) -- (0.5cm, 0.025cm);
    %},		
    %legend entries={
      %\fontsize{6}{0}\selectfont\sffamily \, $C'=2$,
      %\fontsize{6}{0}\selectfont\sffamily \, $C'=5$,
      %\fontsize{6}{0}\selectfont\sffamily \, $C'=25$
    %},
		%legend style={
%%			draw = none,
			%at={(0.95,0.35)},
%%			anchor=north,
			%legend columns=1,
			%row sep=-2pt,
			%%column sep=0.5cm,
      %%inner sep=1.5pt,
		%},
		%legend cell align=left,
    %%reverse legend,
  	]
	%\addlegendimage{color=red!70!black}
  %\addlegendimage{color=blue!70!black}  	
  %\addlegendimage{color=black}
%  \addlegendimage{only marks, mark=*, color=black}
  %
  
  %--- Distances C'=2 ------------------------------------------
  % Likelihood Error
  % Lower bound (invisible plot)
  \addplot [draw=none, stack plots=y, forget plot] table[
    x=n,
    y expr=\thisrow{F}-\thisrow{F_err}
  ] {./figs/BIRCH_GMM_estimated_2.txt};	
  
  % Stack twice the error, draw as area plot
  \addplot [draw=none, fill=red!70!black, stack plots=y, fill opacity=0.15] table [
      x=n,
      y expr=2*\thisrow{F_err}
  ] {./figs/BIRCH_GMM_estimated_2.txt}\closedcycle;
  
  % Reset stack using invisible plot
  \addplot [forget plot, stack plots=y,draw=none] table [x=n, y expr=-(\thisrow{F}+\thisrow{F_err})] {./figs/BIRCH_GMM_estimated_2.txt};
    
  % Free Energy Error
  % Lower bound (invisible plot)
  \addplot [draw=none, stack plots=y, forget plot] table[
    x=n,
    y expr=\thisrow{L}-\thisrow{L_err}
  ] {./figs/BIRCH_GMM_estimated_2.txt};	
  
  % Stack twice the error, draw as area plot
  \addplot [draw=none, fill=red!70!black, stack plots=y, fill opacity=0.15] table [
      x=n,
      y expr=2*\thisrow{L_err}
  ] {./figs/BIRCH_GMM_estimated_2.txt}\closedcycle;
  
  % Reset stack using invisible plot
  \addplot [forget plot, stack plots=y,draw=none] table [x=n, y expr=-(\thisrow{L}+\thisrow{L_err})] {./figs/BIRCH_GMM_estimated_2.txt};

	\addplot [solid, mark=None, color=red!70!black] table[x=n, y=L] {./figs/BIRCH_GMM_estimated_2.txt};	
	\addplot [dashed, mark=None, color=red!70!black] table[x=n, y=F] {./figs/BIRCH_GMM_estimated_2.txt};

  %--- Distances C'=3 ------------------------------------------
  % Likelihood Error
  % Lower bound (invisible plot)
  \addplot [draw=none, stack plots=y, forget plot] table[
    x=n,
    y expr=\thisrow{F}-\thisrow{F_err}
  ] {./figs/BIRCH_GMM_estimated_5.txt};	
  
  % Stack twice the error, draw as area plot
  \addplot [draw=none, fill=blue!70!black, stack plots=y, fill opacity=0.15] table [
      x=n,
      y expr=2*\thisrow{F_err}
  ] {./figs/BIRCH_GMM_estimated_5.txt}\closedcycle;
  
  % Reset stack using invisible plot
  \addplot [forget plot, stack plots=y,draw=none] table [x=n, y expr=-(\thisrow{F}+\thisrow{F_err})] {./figs/BIRCH_GMM_estimated_5.txt};
    
  % Free Energy Error
  % Lower bound (invisible plot)
  \addplot [draw=none, stack plots=y, forget plot] table[
    x=n,
    y expr=\thisrow{L}-\thisrow{L_err}
  ] {./figs/BIRCH_GMM_estimated_5.txt};	
  
  % Stack twice the error, draw as area plot
  \addplot [draw=none, fill=blue!70!black, stack plots=y, fill opacity=0.15] table [
      x=n,
      y expr=2*\thisrow{L_err}
  ] {./figs/BIRCH_GMM_estimated_5.txt}\closedcycle;
  
  % Reset stack using invisible plot
  \addplot [forget plot, stack plots=y,draw=none] table [x=n, y expr=-(\thisrow{L}+\thisrow{L_err})] {./figs/BIRCH_GMM_estimated_5.txt};

	\addplot [solid, mark=None, color=blue!70!black] table[x=n, y=L] {./figs/BIRCH_GMM_estimated_5.txt};	
	\addplot [dashed, mark=None, color=blue!70!black] table[x=n, y=F] {./figs/BIRCH_GMM_estimated_5.txt};

  %--- Distances C'=25 ------------------------------------------
  % Likelihood Error
  % Lower bound (invisible plot)
  \addplot [draw=none, stack plots=y, forget plot] table[
    x=n,
    y expr=\thisrow{F}-\thisrow{F_err}
  ] {./figs/BIRCH_GMM.txt};	
  
  % Stack twice the error, draw as area plot
  \addplot [draw=none, fill=black, stack plots=y, fill opacity=0.15] table [
      x=n,
      y expr=2*\thisrow{F_err}
  ] {./figs/BIRCH_GMM.txt}\closedcycle;
  
  % Reset stack using invisible plot
  \addplot [forget plot, stack plots=y,draw=none] table [x=n, y expr=-(\thisrow{F}+\thisrow{F_err})] {./figs/BIRCH_GMM.txt};
    
  % Free Energy Error
  % Lower bound (invisible plot)
  \addplot [draw=none, stack plots=y, forget plot] table[
    x=n,
    y expr=\thisrow{L}-\thisrow{L_err}
  ] {./figs/BIRCH_GMM.txt};	
  
  % Stack twice the error, draw as area plot
  \addplot [draw=none, fill=black, stack plots=y, fill opacity=0.15] table [
      x=n,
      y expr=2*\thisrow{L_err}
  ] {./figs/BIRCH_GMM.txt}\closedcycle;
  
  % Reset stack using invisible plot
  \addplot [forget plot, stack plots=y,draw=none] table [x=n, y expr=-(\thisrow{L}+\thisrow{L_err})] {./figs/BIRCH_GMM.txt};

	\addplot [solid, mark=None, color=black] table[x=n, y=L] {./figs/BIRCH_GMM.txt};	
	\addplot [dashed, mark=None, color=black] table[x=n, y=F] {./figs/BIRCH_GMM.txt};

  \end{axis}
\end{tikzpicture}
	\end{subfigure}
  \begin{subfigure}[c]{0.24\textwidth}
%			\tikzset{external/remake next}
    	%\tikzset{external/remake next}
\tikzsetnextfilename{BIRCH_k-means_exhaustive_LF}
\pgfplotsset{
	grid style={dotted,gray},
	minor grid style={dotted,lightgray},
  tick label style = {font=\tiny\sansmath\sffamily},
  legend style = {font=\sansmath\sffamily},
  xlabel style = {font=\sansmath\sffamily},
  ylabel style = {font=\sansmath\sffamily},
	% to match the colors of the markers to the plot cycle list 'exotic'
  legend image code/.code={
    \draw[mark repeat=2,mark phase=2]
    plot coordinates {
      (0cm,0cm)
      (0.25cm,0cm)        %% default is (0.3cm,0cm)
      (0.5cm,0cm)         %% default is (0.6cm,0cm)
    };%	
  }
}

\begin{tikzpicture}%[rotate=-90]
	\tikzset{mark size={1.0}}
	\begin{axis}[
		colormap access=direct,
		width = 4.9cm,
		height = 3.5cm,
    xmin=0,
		xmax=50,
		xtick = {0,10,...,50},
%		minor xtick = {0,10,...,500},
		scaled x ticks = false,
		xlabel near ticks, xticklabel pos=lower,
		ymin=-7.0,
		ymax = -6.0,
%		ytick = {-9.3,-9.1,...,-8.5},
%		minor ytick = {5,10,...,95},
		ylabel={\scriptsize\sffamily \phantom{Likelihood / Free Energy}},
		ylabel near ticks,
		yticklabel pos=left,
    y tick label style={
        /pgf/number format/.cd,
            fixed,
            fixed zerofill,
            precision=1,
        /tikz/.cd
    },
		grid = both,
		%
   %legend image code/.code={%
%%     \draw[dash pattern=on 0.175cm off 0.05cm on 0.05cm off 0.05cm on 0.175cm, draw=red!70!black] (0cm,0.05cm) -- (0.5cm,0.05cm);
     %\draw[dashed] (0cm,-0.025cm) -- (0.5cm,-0.025cm);
     %\draw[solid]  (0cm, 0.025cm) -- (0.5cm, 0.025cm);
    %},		
    %legend entries={
      %\fontsize{6}{0}\selectfont\sffamily \, $C'=2$,
      %\fontsize{6}{0}\selectfont\sffamily \, $C'=5$,
      %\fontsize{6}{0}\selectfont\sffamily \, $C'=25$
    %},
		%legend style={
%%			draw = none,
			%at={(0.95,0.35)},
%%			anchor=north,
			%legend columns=1,
			%row sep=-2pt,
			%%column sep=0.5cm,
      %%inner sep=1.5pt,
		%},
		%legend cell align=left,
    %%reverse legend,
  	]
	%\addlegendimage{color=red!70!black}
  %\addlegendimage{color=blue!70!black}  	
  %\addlegendimage{color=black}
%  \addlegendimage{only marks, mark=*, color=black}
  %
  
  %--- Distances C'=2 ------------------------------------------
  % Likelihood Error
  % Lower bound (invisible plot)
  \addplot [draw=none, stack plots=y, forget plot] table[
    x=n,
    y expr=\thisrow{F}-\thisrow{F_err}
  ] {./figs/BIRCH_k-means_exhaustive_2.txt};	
  
  % Stack twice the error, draw as area plot
  \addplot [draw=none, fill=red!70!black, stack plots=y, fill opacity=0.15] table [
      x=n,
      y expr=2*\thisrow{F_err}
  ] {./figs/BIRCH_k-means_exhaustive_2.txt}\closedcycle;
  
  % Reset stack using invisible plot
  \addplot [forget plot, stack plots=y,draw=none] table [x=n, y expr=-(\thisrow{F}+\thisrow{F_err})] {./figs/BIRCH_k-means_exhaustive_2.txt};
    
  % Free Energy Error
  % Lower bound (invisible plot)
  \addplot [draw=none, stack plots=y, forget plot] table[
    x=n,
    y expr=\thisrow{L}-\thisrow{L_err}
  ] {./figs/BIRCH_k-means_exhaustive_2.txt};	
  
  % Stack twice the error, draw as area plot
  \addplot [draw=none, fill=red!70!black, stack plots=y, fill opacity=0.15] table [
      x=n,
      y expr=2*\thisrow{L_err}
  ] {./figs/BIRCH_k-means_exhaustive_2.txt}\closedcycle;
  
  % Reset stack using invisible plot
  \addplot [forget plot, stack plots=y,draw=none] table [x=n, y expr=-(\thisrow{L}+\thisrow{L_err})] {./figs/BIRCH_k-means_exhaustive_2.txt};

	\addplot [solid, mark=None, color=red!70!black] table[x=n, y=L] {./figs/BIRCH_k-means_exhaustive_2.txt};	
	\addplot [dashed, mark=None, color=red!70!black] table[x=n, y=F] {./figs/BIRCH_k-means_exhaustive_2.txt};

  %--- Distances C'=5 ------------------------------------------
  % Likelihood Error
  % Lower bound (invisible plot)
  \addplot [draw=none, stack plots=y, forget plot] table[
    x=n,
    y expr=\thisrow{F}-\thisrow{F_err}
  ] {./figs/BIRCH_k-means_exhaustive_5.txt};	
  
  % Stack twice the error, draw as area plot
  \addplot [draw=none, fill=blue!70!black, stack plots=y, fill opacity=0.15] table [
      x=n,
      y expr=2*\thisrow{F_err}
  ] {./figs/BIRCH_k-means_exhaustive_5.txt}\closedcycle;
  
  % Reset stack using invisible plot
  \addplot [forget plot, stack plots=y,draw=none] table [x=n, y expr=-(\thisrow{F}+\thisrow{F_err})] {./figs/BIRCH_k-means_exhaustive_5.txt};
    
  % Free Energy Error
  % Lower bound (invisible plot)
  \addplot [draw=none, stack plots=y, forget plot] table[
    x=n,
    y expr=\thisrow{L}-\thisrow{L_err}
  ] {./figs/BIRCH_k-means_exhaustive_5.txt};	
  
  % Stack twice the error, draw as area plot
  \addplot [draw=none, fill=blue!70!black, stack plots=y, fill opacity=0.15] table [
      x=n,
      y expr=2*\thisrow{L_err}
  ] {./figs/BIRCH_k-means_exhaustive_5.txt}\closedcycle;
  
  % Reset stack using invisible plot
  \addplot [forget plot, stack plots=y,draw=none] table [x=n, y expr=-(\thisrow{L}+\thisrow{L_err})] {./figs/BIRCH_k-means_exhaustive_5.txt};

	\addplot [solid, mark=None, color=blue!70!black] table[x=n, y=L] {./figs/BIRCH_k-means_exhaustive_5.txt};	
	\addplot [dashed, mark=None, color=blue!70!black] table[x=n, y=F] {./figs/BIRCH_k-means_exhaustive_5.txt};

  %--- Distances C'=25 ------------------------------------------
  % Likelihood Error
  % Lower bound (invisible plot)
  \addplot [draw=none, stack plots=y, forget plot] table[
    x=n,
    y expr=\thisrow{F}-\thisrow{F_err}
  ] {./figs/BIRCH_k-means.txt};	
  
  % Stack twice the error, draw as area plot
  \addplot [draw=none, fill=black, stack plots=y, fill opacity=0.15] table [
      x=n,
      y expr=2*\thisrow{F_err}
  ] {./figs/BIRCH_k-means.txt}\closedcycle;
  
  % Reset stack using invisible plot
  \addplot [forget plot, stack plots=y,draw=none] table [x=n, y expr=-(\thisrow{F}+\thisrow{F_err})] {./figs/BIRCH_k-means.txt};
    
  % Free Energy Error
  % Lower bound (invisible plot)
  \addplot [draw=none, stack plots=y, forget plot] table[
    x=n,
    y expr=\thisrow{L}-\thisrow{L_err}
  ] {./figs/BIRCH_k-means.txt};	
  
  % Stack twice the error, draw as area plot
  \addplot [draw=none, fill=black, stack plots=y, fill opacity=0.15] table [
      x=n,
      y expr=2*\thisrow{L_err}
  ] {./figs/BIRCH_k-means.txt}\closedcycle;
  
  % Reset stack using invisible plot
  \addplot [forget plot, stack plots=y,draw=none] table [x=n, y expr=-(\thisrow{L}+\thisrow{L_err})] {./figs/BIRCH_k-means.txt};

	\addplot [solid, mark=None, color=black] table[x=n, y=L] {./figs/BIRCH_k-means.txt};	
	\addplot [dashed, mark=None, color=black] table[x=n, y=F] {./figs/BIRCH_k-means.txt};

  \end{axis}
\end{tikzpicture}
	\end{subfigure}
  \begin{subfigure}[c]{0.24\textwidth}
%      \tikzset{external/remake next}
    	%\tikzset{external/remake next}
\tikzsetnextfilename{BIRCH_k-means_estimated_LF}
\pgfplotsset{
	grid style={dotted,gray},
	minor grid style={dotted,lightgray},
  tick label style = {font=\tiny\sansmath\sffamily},
  legend style = {font=\sansmath\sffamily},
  xlabel style = {font=\sansmath\sffamily},
  ylabel style = {font=\sansmath\sffamily},
	% to match the colors of the markers to the plot cycle list 'exotic'
  legend image code/.code={
    \draw[mark repeat=2,mark phase=2]
    plot coordinates {
      (0cm,0cm)
      (0.25cm,0cm)        %% default is (0.3cm,0cm)
      (0.5cm,0cm)         %% default is (0.6cm,0cm)
    };%	
  }
}

\begin{tikzpicture}%[rotate=-90]
	\tikzset{mark size={1.0}}
	\begin{axis}[
		colormap access=direct,
		width = 4.9cm,
		height = 3.5cm,
    	xmin=0,
		xmax=50,
		xtick = {0,10,...,50},
%		minor xtick = {0,10,...,500},
		scaled x ticks = false,
		xlabel near ticks, xticklabel pos=lower,
		ymin=-7.0,
		ymax = -6.0,
%		ytick = {-9.3,-9.1,...,-8.5},
%		minor ytick = {5,10,...,95},
		ylabel={\scriptsize\sffamily \phantom{Likelihood / Free Energy}},
		ylabel near ticks,
		yticklabel pos=left,
    y tick label style={
        /pgf/number format/.cd,
            fixed,
            fixed zerofill,
            precision=1,
        /tikz/.cd
    },
		grid = both,
		%%
   %legend image code/.code={%
%%     \draw[dash pattern=on 0.175cm off 0.05cm on 0.05cm off 0.05cm on 0.175cm, draw=red!70!black] (0cm,0.05cm) -- (0.5cm,0.05cm);
     %\draw[dashed] (0cm,-0.025cm) -- (0.5cm,-0.025cm);
     %\draw[solid]  (0cm, 0.025cm) -- (0.5cm, 0.025cm);
    %},		
    %legend entries={
      %%\fontsize{6}{0}\selectfont\sffamily \, $C'=2$,
      %\fontsize{6}{0}\selectfont\sffamily \, $C'=5$,
      %\fontsize{6}{0}\selectfont\sffamily \, $C'=25$
    %},
		%legend style={
%%			draw = none,
			%at={(0.95,0.35)},
%%			anchor=north,
			%legend columns=1,
			%row sep=-2pt,
			%%column sep=0.5cm,
      %%inner sep=1.5pt,
		%},
		%legend cell align=left,
    %%reverse legend,
  	]
  \addplot [draw=none, stack plots=y, forget plot] table[
    x=n,
    y expr=\thisrow{F}-\thisrow{F_err}
  ] {./figs/BIRCH_k-means_estimated_5.txt};	
  
  % Stack twice the error, draw as area plot
  \addplot [draw=none, fill=blue!70!black, stack plots=y, fill opacity=0.15] table [
      x=n,
      y expr=2*\thisrow{F_err}
  ] {./figs/BIRCH_k-means_estimated_5.txt}\closedcycle;
  
  % Reset stack using invisible plot
  \addplot [forget plot, stack plots=y,draw=none] table [x=n, y expr=-(\thisrow{F}+\thisrow{F_err})] {./figs/BIRCH_k-means_estimated_5.txt};
    
  % Free Energy Error
  % Lower bound (invisible plot)
  \addplot [draw=none, stack plots=y, forget plot] table[
    x=n,
    y expr=\thisrow{L}-\thisrow{L_err}
  ] {./figs/BIRCH_k-means_estimated_5.txt};	
  
  % Stack twice the error, draw as area plot
  \addplot [draw=none, fill=blue!70!black, stack plots=y, fill opacity=0.15] table [
      x=n,
      y expr=2*\thisrow{L_err}
  ] {./figs/BIRCH_k-means_estimated_5.txt}\closedcycle;
  
  % Reset stack using invisible plot
  \addplot [forget plot, stack plots=y,draw=none] table [x=n, y expr=-(\thisrow{L}+\thisrow{L_err})] {./figs/BIRCH_k-means_estimated_5.txt};

	\addplot [solid, mark=None, color=blue!70!black] table[x=n, y=L] {./figs/BIRCH_k-means_estimated_5.txt};	
	\addplot [dashed, mark=None, color=blue!70!black] table[x=n, y=F] {./figs/BIRCH_k-means_estimated_5.txt};

  %--- Distances C'=25 ------------------------------------------
  % Likelihood Error
  % Lower bound (invisible plot)
  \addplot [draw=none, stack plots=y, forget plot] table[
    x=n,
    y expr=\thisrow{F}-\thisrow{F_err}
  ] {./figs/BIRCH_k-means.txt};	
  
  % Stack twice the error, draw as area plot
  \addplot [draw=none, fill=black, stack plots=y, fill opacity=0.15] table [
      x=n,
      y expr=2*\thisrow{F_err}
  ] {./figs/BIRCH_k-means.txt}\closedcycle;
  
  % Reset stack using invisible plot
  \addplot [forget plot, stack plots=y,draw=none] table [x=n, y expr=-(\thisrow{F}+\thisrow{F_err})] {./figs/BIRCH_k-means.txt};
    
  % Free Energy Error
  % Lower bound (invisible plot)
  \addplot [draw=none, stack plots=y, forget plot] table[
    x=n,
    y expr=\thisrow{L}-\thisrow{L_err}
  ] {./figs/BIRCH_k-means.txt};	
  
  % Stack twice the error, draw as area plot
  \addplot [draw=none, fill=black, stack plots=y, fill opacity=0.15] table [
      x=n,
      y expr=2*\thisrow{L_err}
  ] {./figs/BIRCH_k-means.txt}\closedcycle;
  
  % Reset stack using invisible plot
  \addplot [forget plot, stack plots=y,draw=none] table [x=n, y expr=-(\thisrow{L}+\thisrow{L_err})] {./figs/BIRCH_k-means.txt};

	\addplot [solid, mark=None, color=black] table[x=n, y=L] {./figs/BIRCH_k-means.txt};	
	\addplot [dashed, mark=None, color=black] table[x=n, y=F] {./figs/BIRCH_k-means.txt};

  \end{axis}
\end{tikzpicture}
	\end{subfigure}\\
	%	
	%\begin{picture}(0,0)
		%\put(35,-40){\fontsize{8}{0}\selectfont\sffamily (a) var-GMM-X}
	%\end{picture}
  \begin{subfigure}[c]{0.24\textwidth}
    \begin{adjustbox}{trim=6pt 0pt 0pt 14pt}
%      \tikzset{external/remake next}
      %\tikzset{external/remake next}
\tikzsetnextfilename{BIRCH_GMM_exhaustive_QE}
\pgfplotsset{
	grid style={dotted,gray},
	minor grid style={dotted,lightgray},
  tick label style = {font=\tiny\sansmath\sffamily},
  legend style = {font=\sansmath\sffamily},
  xlabel style = {font=\sansmath\sffamily},
  ylabel style = {font=\sansmath\sffamily},
	% to match the colors of the markers to the plot cycle list 'exotic'
  legend image code/.code={
    \draw[mark repeat=2,mark phase=2]
    plot coordinates {
      (0cm,0cm)
      (0.25cm,0cm)        %% default is (0.3cm,0cm)
      (0.5cm,0cm)         %% default is (0.6cm,0cm)
    };%	
  }
}

\begin{tikzpicture}%[rotate=-90]
	\tikzset{mark size={1.0}}
	\begin{axis}[
		colormap access=direct,
		width = 4.9cm,
		height = 3.5cm,
    	xmin=0,
		xmax=100,
%		xtick = {0,5,...,75},
%		minor xtick = {0,10,...,500},
		scaled x ticks = false,
		xlabel={\scriptsize\sffamily \phantom{Iteration}},
		xlabel near ticks, xticklabel pos=lower,
		ymin=4000,
		ymax=12000,
%		ytick = {-9.3,-9.1,...,-8.5},
%		minor ytick = {5,10,...,95},
		ylabel={\scriptsize\sffamily Quantization Error},
		ylabel near ticks,
		yticklabel pos=left,
    y label style={at={(-0.2,0.5)}},
    y tick label style={
        /pgf/number format/.cd,
            fixed,
            fixed zerofill,
            precision=1,
        /tikz/.cd
    },
		grid = both,
		%
%   legend image code/.code={%
%     \draw[dash pattern=on 0.175cm off 0.05cm on 0.05cm off 0.05cm on 0.175cm, draw=red!70!black] (0cm,0.05cm) -- (0.5cm,0.05cm);
%     \draw[dashed] (0cm,-0.025cm) -- (0.5cm,-0.025cm);
%     \draw[solid]  (0cm, 0.025cm) -- (0.5cm, 0.025cm);
%    },		
    %legend entries={
      %\fontsize{6}{0}\selectfont\sffamily \, $C'=2$,
      %\fontsize{6}{0}\selectfont\sffamily \, $C'=5$,
      %\fontsize{6}{0}\selectfont\sffamily \, $C'=25$
    %},
		%legend style={
%%			draw = none,
			%at={(0.95,0.95)},
%%			anchor=north,
			%legend columns=1,
			%row sep=-2pt,
			%%column sep=0.5cm,
      %%inner sep=1.5pt,
		%},
		%legend cell align=left,
    %%reverse legend,
  	]
	%\addlegendimage{color=red!70!black}
  %\addlegendimage{color=blue!70!black}  	
  %\addlegendimage{color=black}
%  \addlegendimage{only marks, mark=*, color=black}
  %
  
  %--- Distances C'=2 ------------------------------------------
  % Phi Error
  % Lower bound (invisible plot)
  \addplot [draw=none, stack plots=y, forget plot] table[
    x=n,
    y expr=\thisrow{phi}-\thisrow{phi_err}
  ] {./figs/BIRCH_GMM_exhaustive_2.txt};	
  
  % Stack twice the error, draw as area plot
  \addplot [draw=none, fill=red!70!black, stack plots=y, fill opacity=0.15] table [
      x=n,
      y expr=2*\thisrow{phi_err}
  ] {./figs/BIRCH_GMM_exhaustive_2.txt}\closedcycle;
  
  % Reset stack using invisible plot
  \addplot [forget plot, stack plots=y,draw=none] table [x=n, y expr=-(\thisrow{phi}+\thisrow{phi_err})] {./figs/BIRCH_GMM_exhaustive_2.txt};
     
	\addplot [solid, mark=None, color=red!70!black] table[x=n, y=phi] {./figs/BIRCH_GMM_exhaustive_2.txt};	
	\addplot [dotted, thick, mark=None, color=red!70!black] table[x=n, y=phi] {./figs/BIRCH_GMM_exhaustive_2_best.txt};	

  %--- Distances C'=3 ------------------------------------------
  % Phi Error
  % Lower bound (invisible plot)
  \addplot [draw=none, stack plots=y, forget plot] table[
    x=n,
    y expr=\thisrow{phi}-\thisrow{phi_err}
  ] {./figs/BIRCH_GMM_exhaustive_5.txt};	
  
  % Stack twice the error, draw as area plot
  \addplot [draw=none, fill=blue!70!black, stack plots=y, fill opacity=0.15] table [
      x=n,
      y expr=2*\thisrow{phi_err}
  ] {./figs/BIRCH_GMM_exhaustive_5.txt}\closedcycle;
  
  % Reset stack using invisible plot
  \addplot [forget plot, stack plots=y,draw=none] table [x=n, y expr=-(\thisrow{phi}+\thisrow{phi_err})] {./figs/BIRCH_GMM_exhaustive_5.txt};
     
	\addplot [solid, mark=None, color=blue!70!black] table[x=n, y=phi] {./figs/BIRCH_GMM_exhaustive_5.txt};	
	\addplot [dotted, thick, mark=None, color=blue!70!black] table[x=n, y=phi] {./figs/BIRCH_GMM_exhaustive_5_best.txt};	

  %--- Distances C'=25 ------------------------------------------
  % Phi Error
  % Lower bound (invisible plot)
  \addplot [draw=none, stack plots=y, forget plot] table[
    x=n,
    y expr=\thisrow{phi}-\thisrow{phi_err}
  ] {./figs/BIRCH_GMM.txt};	
  
  % Stack twice the error, draw as area plot
  \addplot [draw=none, fill=black, stack plots=y, fill opacity=0.15] table [
      x=n,
      y expr=2*\thisrow{phi_err}
  ] {./figs/BIRCH_GMM.txt}\closedcycle;
  
  % Reset stack using invisible plot
  \addplot [forget plot, stack plots=y,draw=none] table [x=n, y expr=-(\thisrow{phi}+\thisrow{phi_err})] {./figs/BIRCH_GMM.txt};
     
	\addplot [solid, mark=None, color=black] table[x=n, y=phi] {./figs/BIRCH_GMM.txt};	
	\addplot [dotted, thick, mark=None, color=black] table[x=n, y=phi] {./figs/BIRCH_GMM_best.txt};	

  \end{axis}
\end{tikzpicture}
    \end{adjustbox}
	\end{subfigure}
	\begin{picture}(0,0)
		\put(-130,-20){\rotatebox{90}{\rlap{\makebox[5.80cm]{BIRCH~$5\times 5$}}}}
	\end{picture}
	%
	%\begin{picture}(0,0)
		%\put(13,-25){\rotatebox{90}{\rlap{\makebox[5.30cm]{\hrulefill}}}}
		%\put(35,-40){\fontsize{8}{0}\selectfont\sffamily (b) var-GMM-S}
	%\end{picture}
	%
  \begin{subfigure}[c]{0.24\textwidth}
    \begin{adjustbox}{trim=3pt 0pt 0pt 14pt}
%      \tikzset{external/remake next}
      %\tikzset{external/remake next}
\tikzsetnextfilename{BIRCH_GMM_estimated_QE}
\pgfplotsset{
	grid style={dotted,gray},
	minor grid style={dotted,lightgray},
  tick label style = {font=\tiny\sansmath\sffamily},
  legend style = {font=\sansmath\sffamily},
  xlabel style = {font=\sansmath\sffamily},
  ylabel style = {font=\sansmath\sffamily},
	% to match the colors of the markers to the plot cycle list 'exotic'
  legend image code/.code={
    \draw[mark repeat=2,mark phase=2]
    plot coordinates {
      (0cm,0cm)
      (0.25cm,0cm)        %% default is (0.3cm,0cm)
      (0.5cm,0cm)         %% default is (0.6cm,0cm)
    };%	
  }
}

\begin{tikzpicture}%[rotate=-90]
	\tikzset{mark size={1.0}}
	\begin{axis}[
		colormap access=direct,
		width = 4.9cm,
		height = 3.5cm,
    xmin=0,
		xmax=100,
%		xtick = {0,5,...,75},
%		minor xtick = {0,10,...,500},
		scaled x ticks = false,
		xlabel={\scriptsize\sffamily \phantom{Iteration}},
		xlabel near ticks, xticklabel pos=lower,
		ymin=4000,
		ymax=12000,
%		ytick = {-9.3,-9.1,...,-8.5},
%		minor ytick = {5,10,...,95},
		ylabel={\scriptsize\sffamily \phantom{Quantization Error}},
		ylabel near ticks,
		yticklabel pos=left,
    y tick label style={
        /pgf/number format/.cd,
            fixed,
            fixed zerofill,
            precision=1,
        /tikz/.cd
    },
		grid = both,
		%
%   legend image code/.code={%
%     \draw[dash pattern=on 0.175cm off 0.05cm on 0.05cm off 0.05cm on 0.175cm, draw=red!70!black] (0cm,0.05cm) -- (0.5cm,0.05cm);
%     \draw[dashed] (0cm,-0.025cm) -- (0.5cm,-0.025cm);
%     \draw[solid]  (0cm, 0.025cm) -- (0.5cm, 0.025cm);
%    },		
    %legend entries={
      %\fontsize{6}{0}\selectfont\sffamily \, $C'=2$,
      %\fontsize{6}{0}\selectfont\sffamily \, $C'=5$,
      %\fontsize{6}{0}\selectfont\sffamily \, $C'=25$
    %},
		%legend style={
%%			draw = none,
			%at={(0.95,0.95)},
%%			anchor=north,
			%legend columns=1,
			%row sep=-2pt,
			%%column sep=0.5cm,
      %%inner sep=1.5pt,
		%},
		%legend cell align=left,
    %%reverse legend,
  	]
	%\addlegendimage{color=red!70!black}
  %\addlegendimage{color=blue!70!black}  	
  %\addlegendimage{color=black}
%  \addlegendimage{only marks, mark=*, color=black}
  %
  
  %--- Distances C'=2 ------------------------------------------
  % Phi Error
  % Lower bound (invisible plot)
  \addplot [draw=none, stack plots=y, forget plot] table[
    x=n,
    y expr=\thisrow{phi}-\thisrow{phi_err}
  ] {./figs/BIRCH_GMM_estimated_2.txt};	
  
  % Stack twice the error, draw as area plot
  \addplot [draw=none, fill=red!70!black, stack plots=y, fill opacity=0.15] table [
      x=n,
      y expr=2*\thisrow{phi_err}
  ] {./figs/BIRCH_GMM_estimated_2.txt}\closedcycle;
  
  % Reset stack using invisible plot
  \addplot [forget plot, stack plots=y,draw=none] table [x=n, y expr=-(\thisrow{phi}+\thisrow{phi_err})] {./figs/BIRCH_GMM_estimated_2.txt};
     
	\addplot [solid, mark=None, color=red!70!black] table[x=n, y=phi] {./figs/BIRCH_GMM_estimated_2.txt};	
	\addplot [dotted, thick, mark=None, color=red!70!black] table[x=n, y=phi] {./figs/BIRCH_GMM_estimated_2_best.txt};	

  %--- Distances C'=5 ------------------------------------------
  % Phi Error
  % Lower bound (invisible plot)
  \addplot [draw=none, stack plots=y, forget plot] table[
    x=n,
    y expr=\thisrow{phi}-\thisrow{phi_err}
  ] {./figs/BIRCH_GMM_estimated_5.txt};	
  
  % Stack twice the error, draw as area plot
  \addplot [draw=none, fill=blue!70!black, stack plots=y, fill opacity=0.15] table [
      x=n,
      y expr=2*\thisrow{phi_err}
  ] {./figs/BIRCH_GMM_estimated_5.txt}\closedcycle;
  
  % Reset stack using invisible plot
  \addplot [forget plot, stack plots=y,draw=none] table [x=n, y expr=-(\thisrow{phi}+\thisrow{phi_err})] {./figs/BIRCH_GMM_estimated_5.txt};
     
	\addplot [solid, mark=None, color=blue!70!black] table[x=n, y=phi] {./figs/BIRCH_GMM_estimated_5.txt};	
	\addplot [dotted, thick, mark=None, color=blue!70!black] table[x=n, y=phi] {./figs/BIRCH_GMM_estimated_5_best.txt};	

  %--- Distances C'=25 ------------------------------------------
  % Phi Error
  % Lower bound (invisible plot)
  \addplot [draw=none, stack plots=y, forget plot] table[
    x=n,
    y expr=\thisrow{phi}-\thisrow{phi_err}
  ] {./figs/BIRCH_GMM.txt};	
  
  % Stack twice the error, draw as area plot
  \addplot [draw=none, fill=black, stack plots=y, fill opacity=0.15] table [
      x=n,
      y expr=2*\thisrow{phi_err}
  ] {./figs/BIRCH_GMM.txt}\closedcycle;
  
  % Reset stack using invisible plot
  \addplot [forget plot, stack plots=y,draw=none] table [x=n, y expr=-(\thisrow{phi}+\thisrow{phi_err})] {./figs/BIRCH_GMM.txt};
     
	\addplot [solid, mark=None, color=black] table[x=n, y=phi] {./figs/BIRCH_GMM.txt};	
	\addplot [dotted, thick, mark=None, color=black] table[x=n, y=phi] {./figs/BIRCH_GMM_best.txt};	

  \end{axis}
\end{tikzpicture}
    \end{adjustbox}
	\end{subfigure}
	%
	%\begin{picture}(0,0)
		%\put(11,-25){\rotatebox{90}{\rlap{\makebox[5.30cm]{\hrulefill}}}}
		%\put(10,-25){\rotatebox{90}{\rlap{\makebox[5.30cm]{\hrulefill}}}}
		%\put(31,-40){\fontsize{8}{0}\selectfont\sffamily (c) var-$k$-means-X}
	%\end{picture}
	%
  \begin{subfigure}[c]{0.24\textwidth}
    \begin{adjustbox}{trim=3pt 0pt 0pt 14pt}
%      \tikzset{external/remake next}
      %\tikzset{external/remake next}
\tikzsetnextfilename{BIRCH_k-means_exhaustive_QE}
\pgfplotsset{
	grid style={dotted,gray},
	minor grid style={dotted,lightgray},
  tick label style = {font=\tiny\sansmath\sffamily},
  legend style = {font=\sansmath\sffamily},
  xlabel style = {font=\sansmath\sffamily},
  ylabel style = {font=\sansmath\sffamily},
	% to match the colors of the markers to the plot cycle list 'exotic'
  legend image code/.code={
    \draw[mark repeat=2,mark phase=2]
    plot coordinates {
      (0cm,0cm)
      (0.25cm,0cm)        %% default is (0.3cm,0cm)
      (0.5cm,0cm)         %% default is (0.6cm,0cm)
    };%	
  }
}

\begin{tikzpicture}%[rotate=-90]
	\tikzset{mark size={1.0}}
	\begin{axis}[
		colormap access=direct,
		width = 4.9cm,
		height = 3.5cm,
    	xmin=0,
		xmax=50,
		xtick = {0,10,...,50},
%		minor xtick = {0,10,...,500},
		scaled x ticks = false,
		xlabel={\scriptsize\sffamily \phantom{Iteration}},
		xlabel near ticks, xticklabel pos=lower,
		ymin= 4000,
		ymax = 12000,
%		ytick = {-9.3,-9.1,...,-8.5},
%		minor ytick = {5,10,...,95},
		ylabel={\scriptsize\sffamily \phantom{Quantization Error}},
		ylabel near ticks,
		yticklabel pos=left,
    y tick label style={
        /pgf/number format/.cd,
            fixed,
            fixed zerofill,
            precision=1,
        /tikz/.cd
    },
		grid = both,
		%
%   legend image code/.code={%
%     \draw[dash pattern=on 0.175cm off 0.05cm on 0.05cm off 0.05cm on 0.175cm, draw=red!70!black] (0cm,0.05cm) -- (0.5cm,0.05cm);
%     \draw[dashed] (0cm,-0.025cm) -- (0.5cm,-0.025cm);
%     \draw[solid]  (0cm, 0.025cm) -- (0.5cm, 0.025cm);
%    },		
    %legend entries={
      %\fontsize{6}{0}\selectfont\sffamily \, $C'=2$,
      %\fontsize{6}{0}\selectfont\sffamily \, $C'=5$,
      %\fontsize{6}{0}\selectfont\sffamily \, $C'=25$
    %},
		%legend style={
%%			draw = none,
			%at={(0.95,0.95)},
%%			anchor=north,
			%legend columns=1,
			%row sep=-2pt,
			%%column sep=0.5cm,
      %%inner sep=1.5pt,
		%},
		%legend cell align=left,
    %%reverse legend,
  	]
	%\addlegendimage{color=red!70!black}
  %\addlegendimage{color=blue!70!black}  	
  %\addlegendimage{color=black}
%%  \addlegendimage{only marks, mark=*, color=black}
  %
  
  %--- Distances C'=2 ------------------------------------------
  % Phi Error
  % Lower bound (invisible plot)
  \addplot [draw=none, stack plots=y, forget plot] table[
    x=n,
    y expr=\thisrow{phi}-\thisrow{phi_err}
  ] {./figs/BIRCH_k-means_exhaustive_2.txt};	
  
  % Stack twice the error, draw as area plot
  \addplot [draw=none, fill=red!70!black, stack plots=y, fill opacity=0.15] table [
      x=n,
      y expr=2*\thisrow{phi_err}
  ] {./figs/BIRCH_k-means_exhaustive_2.txt}\closedcycle;
  
  % Reset stack using invisible plot
  \addplot [forget plot, stack plots=y,draw=none] table [x=n, y expr=-(\thisrow{phi}+\thisrow{phi_err})] {./figs/BIRCH_k-means_exhaustive_2.txt};
     
	\addplot [solid, mark=None, color=red!70!black] table[x=n, y=phi] {./figs/BIRCH_k-means_exhaustive_2.txt};	
	\addplot [dotted, thick, mark=None, color=red!70!black] table[x=n, y=phi] {./figs/BIRCH_k-means_exhaustive_2_best.txt};	

  %--- Distances C'=5 ------------------------------------------
  % Phi Error
  % Lower bound (invisible plot)
  \addplot [draw=none, stack plots=y, forget plot] table[
    x=n,
    y expr=\thisrow{phi}-\thisrow{phi_err}
  ] {./figs/BIRCH_k-means_exhaustive_5.txt};	
  
  % Stack twice the error, draw as area plot
  \addplot [draw=none, fill=blue!70!black, stack plots=y, fill opacity=0.15] table [
      x=n,
      y expr=2*\thisrow{phi_err}
  ] {./figs/BIRCH_k-means_exhaustive_5.txt}\closedcycle;
  
  % Reset stack using invisible plot
  \addplot [forget plot, stack plots=y,draw=none] table [x=n, y expr=-(\thisrow{phi}+\thisrow{phi_err})] {./figs/BIRCH_k-means_exhaustive_5.txt};
     
	\addplot [solid, mark=None, color=blue!70!black] table[x=n, y=phi] {./figs/BIRCH_k-means_exhaustive_5.txt};	
	\addplot [dotted, thick, mark=None, color=blue!70!black] table[x=n, y=phi] {./figs/BIRCH_k-means_exhaustive_5_best.txt};	

  %--- Distances C'=25 ------------------------------------------
  % Phi Error
  % Lower bound (invisible plot)
  \addplot [draw=none, stack plots=y, forget plot] table[
    x=n,
    y expr=\thisrow{phi}-\thisrow{phi_err}
  ] {./figs/BIRCH_k-means.txt};	
  
  % Stack twice the error, draw as area plot
  \addplot [draw=none, fill=black, stack plots=y, fill opacity=0.15] table [
      x=n,
      y expr=2*\thisrow{phi_err}
  ] {./figs/BIRCH_k-means.txt}\closedcycle;
  
  % Reset stack using invisible plot
  \addplot [forget plot, stack plots=y,draw=none] table [x=n, y expr=-(\thisrow{phi}+\thisrow{phi_err})] {./figs/BIRCH_k-means.txt};
     
	\addplot [solid, mark=None, color=black] table[x=n, y=phi] {./figs/BIRCH_k-means.txt};	
	\addplot [dotted, thick, mark=None, color=black] table[x=n, y=phi] {./figs/BIRCH_k-means_best.txt};	

  \end{axis}
\end{tikzpicture}
    \end{adjustbox}
	\end{subfigure}
	%
	%\begin{picture}(0,0)
		%\put(9,-25){\rotatebox{90}{\rlap{\makebox[5.30cm]{\hrulefill}}}}
		%\put(29,-40){\fontsize{8}{0}\selectfont\sffamily (d) var-$k$-means-S}
	%\end{picture}
	%
  \begin{subfigure}[c]{0.24\textwidth}
    \begin{adjustbox}{trim=3pt 0pt 0pt 14pt}
%      \tikzset{external/remake next}
      %\tikzset{external/remake next}
\tikzsetnextfilename{BIRCH_k-means_estimated_QE}
\pgfplotsset{
	grid style={dotted,gray},
	minor grid style={dotted,lightgray},
  tick label style = {font=\tiny\sansmath\sffamily},
  legend style = {font=\sansmath\sffamily},
  xlabel style = {font=\sansmath\sffamily},
  ylabel style = {font=\sansmath\sffamily},
	% to match the colors of the markers to the plot cycle list 'exotic'
  legend image code/.code={
    \draw[mark repeat=2,mark phase=2]
    plot coordinates {
      (0cm,0cm)
      (0.25cm,0cm)        %% default is (0.3cm,0cm)
      (0.5cm,0cm)         %% default is (0.6cm,0cm)
    };%	
  }
}

\begin{tikzpicture}%[rotate=-90]
	\tikzset{mark size={1.0}}
	\begin{axis}[
		colormap access=direct,
		width = 4.9cm,
		height = 3.5cm,
    	xmin=0,
		xmax=50,
		xtick = {0,10,...,50},
%		minor xtick = {0,10,...,500},
		scaled x ticks = false,
		xlabel={\scriptsize\sffamily \phantom{Iteration}},
		xlabel near ticks, xticklabel pos=lower,
		ymin= 4000,
		ymax = 12000,
%		ytick = {-9.3,-9.1,...,-8.5},
%		minor ytick = {5,10,...,95},
		ylabel={\scriptsize\sffamily \phantom{Quantization Error}},
		ylabel near ticks,
		yticklabel pos=left,
    y tick label style={
        /pgf/number format/.cd,
            fixed,
            fixed zerofill,
            precision=1,
        /tikz/.cd
    },
		grid = both,
		%
%   legend image code/.code={%
%     \draw[dash pattern=on 0.175cm off 0.05cm on 0.05cm off 0.05cm on 0.175cm, draw=red!70!black] (0cm,0.05cm) -- (0.5cm,0.05cm);
%     \draw[dashed] (0cm,-0.025cm) -- (0.5cm,-0.025cm);
%     \draw[solid]  (0cm, 0.025cm) -- (0.5cm, 0.025cm);
%    },		
    %legend entries={
      %%\fontsize{6}{0}\selectfont\sffamily \, $C'=2$,
      %\fontsize{6}{0}\selectfont\sffamily \, $C'=5$,
      %\fontsize{6}{0}\selectfont\sffamily \, $C'=25$
    %},
		%legend style={
%%			draw = none,
			%at={(0.95,0.95)},
%%			anchor=north,
			%legend columns=1,
			%row sep=-2pt,
			%%column sep=0.5cm,
      %%inner sep=1.5pt,
		%},
		%legend cell align=left,
    %%reverse legend,
  	]
	%%\addlegendimage{color=red!70!black}
  %\addlegendimage{color=blue!70!black}  	
  %\addlegendimage{color=black}
%%  \addlegendimage{only marks, mark=*, color=black}
  %
  
%  %--- Distances C'=3 ------------------------------------------
%  % Phi Error
%  % Lower bound (invisible plot)
%  \addplot [draw=none, stack plots=y, forget plot] table[
%    x=n,
%    y expr=\thisrow{phi}-\thisrow{phi_err}
%  ] {./figs/BIRCH_k-means_estimated_2.txt};	
%  
%  % Stack twice the error, draw as area plot
%  \addplot [draw=none, fill=red!70!black, stack plots=y, fill opacity=0.15] table [
%      x=n,
%      y expr=2*\thisrow{phi_err}
%  ] {./figs/BIRCH_k-means_estimated_2.txt}\closedcycle;
%  
%  % Reset stack using invisible plot
%  \addplot [forget plot, stack plots=y,draw=none] table [x=n, y expr=-(\thisrow{phi}+\thisrow{phi_err})] {./figs/BIRCH_k-means_estimated_2.txt};
%     
%	\addplot [solid, mark=None, color=red!70!black] table[x=n, y=phi] {./figs/BIRCH_k-means_estimated_2.txt};	

  %--- Distances C'=5 ------------------------------------------
  % Phi Error
  % Lower bound (invisible plot)
  \addplot [draw=none, stack plots=y, forget plot] table[
    x=n,
    y expr=\thisrow{phi}-\thisrow{phi_err}
  ] {./figs/BIRCH_k-means_estimated_5.txt};	
  
  % Stack twice the error, draw as area plot
  \addplot [draw=none, fill=blue!70!black, stack plots=y, fill opacity=0.15] table [
      x=n,
      y expr=2*\thisrow{phi_err}
  ] {./figs/BIRCH_k-means_estimated_5.txt}\closedcycle;
  
  % Reset stack using invisible plot
  \addplot [forget plot, stack plots=y,draw=none] table [x=n, y expr=-(\thisrow{phi}+\thisrow{phi_err})] {./figs/BIRCH_k-means_estimated_5.txt};
     
	\addplot [solid, mark=None, color=blue!70!black] table[x=n, y=phi] {./figs/BIRCH_k-means_estimated_5.txt};	
	\addplot [dotted, thick, mark=None, color=blue!70!black] table[x=n, y=phi] {./figs/BIRCH_k-means_estimated_5_best.txt};	

  %--- Distances C'=25 ------------------------------------------
  % Phi Error
  % Lower bound (invisible plot)
  \addplot [draw=none, stack plots=y, forget plot] table[
    x=n,
    y expr=\thisrow{phi}-\thisrow{phi_err}
  ] {./figs/BIRCH_k-means.txt};	
  
  % Stack twice the error, draw as area plot
  \addplot [draw=none, fill=black, stack plots=y, fill opacity=0.15] table [
      x=n,
      y expr=2*\thisrow{phi_err}
  ] {./figs/BIRCH_k-means.txt}\closedcycle;
  
  % Reset stack using invisible plot
  \addplot [forget plot, stack plots=y,draw=none] table [x=n, y expr=-(\thisrow{phi}+\thisrow{phi_err})] {./figs/BIRCH_k-means.txt};
     
	\addplot [solid, mark=None, color=black] table[x=n, y=phi] {./figs/BIRCH_k-means.txt};	
	\addplot [dotted, thick, mark=None, color=black] table[x=n, y=phi] {./figs/BIRCH_k-means_best.txt};	

  \end{axis}
\end{tikzpicture}
    \end{adjustbox}
	\end{subfigure}\\
%	\vspace{6pt}\\
	%
  \begin{subfigure}[c]{0.24\textwidth}
%			\tikzset{external/remake next}
			\vspace{-3pt}
			%\tikzset{external/remake next}
\tikzsetnextfilename{KDD_GMM_exhaustive_LF}
\pgfplotsset{
	grid style={dotted,gray},
	minor grid style={dotted,lightgray},
  tick label style = {font=\tiny\sansmath\sffamily},
  legend style = {font=\sansmath\sffamily},
  xlabel style = {font=\sansmath\sffamily},
  ylabel style = {font=\sansmath\sffamily},
	% to match the colors of the markers to the plot cycle list 'exotic'
  legend image code/.code={
    \draw[mark repeat=2,mark phase=2]
    plot coordinates {
      (0cm,0cm)
      (0.25cm,0cm)        %% default is (0.3cm,0cm)
      (0.5cm,0cm)         %% default is (0.6cm,0cm)
    };%	
  }
}

\begin{tikzpicture}%[rotate=-90]
	\tikzset{mark size={1.0}}
	\begin{axis}[
		colormap access=direct,
		width = 4.9cm,
		height = 3.5cm,
    	xmin=0,
		xmax=50,
		xtick = {0,10,...,50},
%		minor xtick = {0,10,...,500},
		scaled x ticks = false,
		xlabel near ticks, xticklabel pos=lower,
		ymin=-500,
		ymax = -450,
%		ytick = {-9.3,-9.1,...,-8.5},
%		minor ytick = {5,10,...,95},
		ylabel={\scriptsize\sffamily Log-Likelihood / Free Energy},
		ylabel near ticks,
		yticklabel pos=left,
    y label style={at={(-0.18,0.5)}},
    grid = both,
   legend image code/.code={%
%     \draw[dash pattern=on 0.175cm off 0.05cm on 0.05cm off 0.05cm on 0.175cm, draw=red!70!black] (0cm,0.05cm) -- (0.5cm,0.05cm);
     \draw[solid]  (0cm, 0.05cm) -- (0.52cm, 0.05cm);
     \draw[dashed] (0cm,0.0cm) -- (0.52cm,0.0cm);
     \draw[dotted, thick] (0cm,-0.05cm) -- (0.52cm,-0.05cm);
    },		
    legend entries={
      \hspace{-4pt}\fontsize{6}{0}\selectfont\sffamily $G=2$,
      \hspace{-4pt}\fontsize{6}{0}\selectfont\sffamily $G=5$,
      \hspace{-4pt}\fontsize{6}{0}\selectfont\sffamily $G=20$,
      \hspace{-4pt}\fontsize{6}{0}\selectfont\sffamily $G=C=200$
    },
		legend style={
			at={(1.98,1.35)},
			legend columns=4,
%			row sep=-2pt,
			column sep=0.2cm,
      inner sep=1.5pt,
		},
		legend cell align=left,
    %reverse legend,
  	]
	\addlegendimage{color=red!70!black}
  \addlegendimage{color=blue!70!black}  	
  \addlegendimage{color=green!50!black}  	
  \addlegendimage{color=black}
  
  %--- Distances C'=2 ------------------------------------------
  \newcommand\filename{./figs/KDD_GMM_exhaustive_2.txt}
  % Likelihood Error
  % Lower bound (invisible plot)
  \addplot [draw=none, stack plots=y, forget plot] table[
    x=n,
    y expr=\thisrow{F}-\thisrow{F_err}
  ] {\filename};	
  
  % Stack twice the error, draw as area plot
  \addplot [draw=none, fill=red!70!black, stack plots=y, fill opacity=0.15] table [
      x=n,
      y expr=2*\thisrow{F_err}
  ] {\filename}\closedcycle;
  
  % Reset stack using invisible plot
  \addplot [forget plot, stack plots=y,draw=none] table [x=n, y expr=-(\thisrow{F}+\thisrow{F_err})] {\filename};
    
  % Free Energy Error
  % Lower bound (invisible plot)
  \addplot [draw=none, stack plots=y, forget plot] table[
    x=n,
    y expr=\thisrow{L}-\thisrow{L_err}
  ] {\filename};	
  
  % Stack twice the error, draw as area plot
  \addplot [draw=none, fill=red!70!black, stack plots=y, fill opacity=0.15] table [
      x=n,
      y expr=2*\thisrow{L_err}
  ] {\filename}\closedcycle;
  
  % Reset stack using invisible plot
  \addplot [forget plot, stack plots=y,draw=none] table [x=n, y expr=-(\thisrow{L}+\thisrow{L_err})] {\filename};

	\addplot [solid, mark=None, color=red!70!black] table[x=n, y=L] {\filename};	
	\addplot [dashed, mark=None, color=red!70!black] table[x=n, y=F] {\filename};

  %--- Distances C'=5 ------------------------------------------
  \renewcommand\filename{./figs/KDD_GMM_exhaustive_5.txt}
  % Likelihood Error
  % Lower bound (invisible plot)
  \addplot [draw=none, stack plots=y, forget plot] table[
    x=n,
    y expr=\thisrow{F}-\thisrow{F_err}
  ] {\filename};	
  
  % Stack twice the error, draw as area plot
  \addplot [draw=none, fill=blue!70!black, stack plots=y, fill opacity=0.15] table [
      x=n,
      y expr=2*\thisrow{F_err}
  ] {\filename}\closedcycle;
  
  % Reset stack using invisible plot
  \addplot [forget plot, stack plots=y,draw=none] table [x=n, y expr=-(\thisrow{F}+\thisrow{F_err})] {\filename};
    
  % Free Energy Error
  % Lower bound (invisible plot)
  \addplot [draw=none, stack plots=y, forget plot] table[
    x=n,
    y expr=\thisrow{L}-\thisrow{L_err}
  ] {\filename};	
  
  % Stack twice the error, draw as area plot
  \addplot [draw=none, fill=blue!70!black, stack plots=y, fill opacity=0.15] table [
      x=n,
      y expr=2*\thisrow{L_err}
  ] {\filename}\closedcycle;
  
  % Reset stack using invisible plot
  \addplot [forget plot, stack plots=y,draw=none] table [x=n, y expr=-(\thisrow{L}+\thisrow{L_err})] {\filename};

	\addplot [solid, mark=None, color=blue!70!black] table[x=n, y=L] {\filename};	
	\addplot [dashed, mark=None, color=blue!70!black] table[x=n, y=F] {\filename};

  %--- Distances C'=20 ------------------------------------------
  \renewcommand\filename{./figs/KDD_GMM_exhaustive_20.txt}
  % Likelihood Error
  % Lower bound (invisible plot)
  \addplot [draw=none, stack plots=y, forget plot] table[
    x=n,
    y expr=\thisrow{F}-\thisrow{F_err}
  ] {\filename};	
  
  % Stack twice the error, draw as area plot
  \addplot [draw=none, fill=green!50!black, stack plots=y, fill opacity=0.15] table [
      x=n,
      y expr=2*\thisrow{F_err}
  ] {\filename}\closedcycle;
  
  % Reset stack using invisible plot
  \addplot [forget plot, stack plots=y,draw=none] table [x=n, y expr=-(\thisrow{F}+\thisrow{F_err})] {\filename};
    
  % Free Energy Error
  % Lower bound (invisible plot)
  \addplot [draw=none, stack plots=y, forget plot] table[
    x=n,
    y expr=\thisrow{L}-\thisrow{L_err}
  ] {\filename};	
  
  % Stack twice the error, draw as area plot
  \addplot [draw=none, fill=green!50!black, stack plots=y, fill opacity=0.15] table [
      x=n,
      y expr=2*\thisrow{L_err}
  ] {\filename}\closedcycle;
  
  % Reset stack using invisible plot
  \addplot [forget plot, stack plots=y,draw=none] table [x=n, y expr=-(\thisrow{L}+\thisrow{L_err})] {\filename};

	\addplot [solid, mark=None, color=green!50!black] table[x=n, y=L] {\filename};	
	\addplot [dashed, mark=None, color=green!50!black] table[x=n, y=F] {\filename};

  %--- Distances C'=200 ------------------------------------------
  \renewcommand\filename{./figs/KDD_GMM.txt}
  % Likelihood Error
  % Lower bound (invisible plot)
  \addplot [draw=none, stack plots=y, forget plot] table[
    x=n,
    y expr=\thisrow{F}-\thisrow{F_err}
  ] {\filename};	
  
  % Stack twice the error, draw as area plot
  \addplot [draw=none, fill=black, stack plots=y, fill opacity=0.15] table [
      x=n,
      y expr=2*\thisrow{F_err}
  ] {\filename}\closedcycle;
  
  % Reset stack using invisible plot
  \addplot [forget plot, stack plots=y,draw=none] table [x=n, y expr=-(\thisrow{F}+\thisrow{F_err})] {\filename};
    
  % Free Energy Error
  % Lower bound (invisible plot)
  \addplot [draw=none, stack plots=y, forget plot] table[
    x=n,
    y expr=\thisrow{L}-\thisrow{L_err}
  ] {\filename};	
  
  % Stack twice the error, draw as area plot
  \addplot [draw=none, fill=black, stack plots=y, fill opacity=0.15] table [
      x=n,
      y expr=2*\thisrow{L_err}
  ] {\filename}\closedcycle;
  
  % Reset stack using invisible plot
  \addplot [forget plot, stack plots=y,draw=none] table [x=n, y expr=-(\thisrow{L}+\thisrow{L_err})] {\filename};

	\addplot [solid, mark=None, color=black] table[x=n, y=L] {\filename};	
	\addplot [dashed, mark=None, color=black] table[x=n, y=F] {\filename};

  \end{axis}
%	\fill[white] (-0.7,-0.5) rectangle (2.7,2.05);
\end{tikzpicture}
			\hfill
	\end{subfigure}
  \begin{subfigure}[c]{0.24\textwidth}
%			\tikzset{external/remake next}
    	%\tikzset{external/remake next}
\tikzsetnextfilename{KDD_GMM_estimated_LF}
\pgfplotsset{
	grid style={dotted,gray},
	minor grid style={dotted,lightgray},
  tick label style = {font=\tiny\sansmath\sffamily},
  legend style = {font=\sansmath\sffamily},
  xlabel style = {font=\sansmath\sffamily},
  ylabel style = {font=\sansmath\sffamily},
	% to match the colors of the markers to the plot cycle list 'exotic'
  legend image code/.code={
    \draw[mark repeat=2,mark phase=2]
    plot coordinates {
      (0cm,0cm)
      (0.25cm,0cm)        %% default is (0.3cm,0cm)
      (0.5cm,0cm)         %% default is (0.6cm,0cm)
    };%	
  }
}

\begin{tikzpicture}%[rotate=-90]
	\tikzset{mark size={1.0}}
	\begin{axis}[
		colormap access=direct,
		width = 4.9cm,
		height = 3.5cm,
    	xmin=0,
		xmax=50,
		xtick = {0,10,...,50},
%		minor xtick = {0,10,...,500},
		scaled x ticks = false,
		xlabel near ticks, xticklabel pos=lower,
		ymin=-500,
		ymax = -450,
%		ytick = {-9.3,-9.1,...,-8.5},
%		minor ytick = {5,10,...,95},
		ylabel={\scriptsize\sffamily \phantom{Likelihood / Free Energy}},
		ylabel near ticks,
		yticklabel pos=left,
		grid = both,
		%
   %legend image code/.code={%
%%     \draw[dash pattern=on 0.175cm off 0.05cm on 0.05cm off 0.05cm on 0.175cm, draw=red!70!black] (0cm,0.05cm) -- (0.5cm,0.05cm);
     %\draw[dashed] (0cm,-0.025cm) -- (0.5cm,-0.025cm);
     %\draw[solid]  (0cm, 0.025cm) -- (0.5cm, 0.025cm);
    %},		
    %legend entries={
      %\hspace{-6pt}\fontsize{6}{0}\selectfont\sffamily \, $C'=2$,
      %\hspace{-4pt}\fontsize{6}{0}\selectfont\sffamily $C'=5$,
      %\hspace{-4pt}\fontsize{6}{0}\selectfont\sffamily $C'=25$
    %},
		%legend style={
%%			draw = none,
			%at={(1.75,1.4)},
%%			anchor=north,
			%legend columns=3,
%%			row sep=-2pt,
			%column sep=0.2cm,
      %inner sep=1.5pt,
		%},
		%legend cell align=left,
    %%reverse legend,
  	]
	%\addlegendimage{color=red!70!black}
  %\addlegendimage{color=blue!70!black}  	
  %\addlegendimage{color=black}
%%  \addlegendimage{only marks, mark=*, color=black}

  %--- Distances C'=2 ------------------------------------------
  \newcommand\filename{./figs/KDD_GMM_estimated_2.txt}
  % Likelihood Error
  % Lower bound (invisible plot)
  \addplot [draw=none, stack plots=y, forget plot] table[
    x=n,
    y expr=\thisrow{F}-\thisrow{F_err}
  ] {\filename};	
  
  % Stack twice the error, draw as area plot
  \addplot [draw=none, fill=red!70!black, stack plots=y, fill opacity=0.15] table [
      x=n,
      y expr=2*\thisrow{F_err}
  ] {\filename}\closedcycle;
  
  % Reset stack using invisible plot
  \addplot [forget plot, stack plots=y,draw=none] table [x=n, y expr=-(\thisrow{F}+\thisrow{F_err})] {\filename};
    
  % Free Energy Error
  % Lower bound (invisible plot)
  \addplot [draw=none, stack plots=y, forget plot] table[
    x=n,
    y expr=\thisrow{L}-\thisrow{L_err}
  ] {\filename};	
  
  % Stack twice the error, draw as area plot
  \addplot [draw=none, fill=red!70!black, stack plots=y, fill opacity=0.15] table [
      x=n,
      y expr=2*\thisrow{L_err}
  ] {\filename}\closedcycle;
  
  % Reset stack using invisible plot
  \addplot [forget plot, stack plots=y,draw=none] table [x=n, y expr=-(\thisrow{L}+\thisrow{L_err})] {\filename};

	\addplot [solid, mark=None, color=red!70!black] table[x=n, y=L] {\filename};	
	\addplot [dashed, mark=None, color=red!70!black] table[x=n, y=F] {\filename};

  %--- Distances C'=5 ------------------------------------------
  \renewcommand\filename{./figs/KDD_GMM_estimated_5.txt}
  % Likelihood Error
  % Lower bound (invisible plot)
  \addplot [draw=none, stack plots=y, forget plot] table[
    x=n,
    y expr=\thisrow{F}-\thisrow{F_err}
  ] {\filename};	
  
  % Stack twice the error, draw as area plot
  \addplot [draw=none, fill=blue!70!black, stack plots=y, fill opacity=0.15] table [
      x=n,
      y expr=2*\thisrow{F_err}
  ] {\filename}\closedcycle;
  
  % Reset stack using invisible plot
  \addplot [forget plot, stack plots=y,draw=none] table [x=n, y expr=-(\thisrow{F}+\thisrow{F_err})] {\filename};
    
  % Free Energy Error
  % Lower bound (invisible plot)
  \addplot [draw=none, stack plots=y, forget plot] table[
    x=n,
    y expr=\thisrow{L}-\thisrow{L_err}
  ] {\filename};	
  
  % Stack twice the error, draw as area plot
  \addplot [draw=none, fill=blue!70!black, stack plots=y, fill opacity=0.15] table [
      x=n,
      y expr=2*\thisrow{L_err}
  ] {\filename}\closedcycle;
  
  % Reset stack using invisible plot
  \addplot [forget plot, stack plots=y,draw=none] table [x=n, y expr=-(\thisrow{L}+\thisrow{L_err})] {\filename};

	\addplot [solid, mark=None, color=blue!70!black] table[x=n, y=L] {\filename};	
	\addplot [dashed, mark=None, color=blue!70!black] table[x=n, y=F] {\filename};

  %--- Distances C'=20 ------------------------------------------
  \renewcommand\filename{./figs/KDD_GMM_estimated_20.txt}
  % Likelihood Error
  % Lower bound (invisible plot)
  \addplot [draw=none, stack plots=y, forget plot] table[
    x=n,
    y expr=\thisrow{F}-\thisrow{F_err}
  ] {\filename};	
  
  % Stack twice the error, draw as area plot
  \addplot [draw=none, fill=green!50!black, stack plots=y, fill opacity=0.15] table [
      x=n,
      y expr=2*\thisrow{F_err}
  ] {\filename}\closedcycle;
  
  % Reset stack using invisible plot
  \addplot [forget plot, stack plots=y,draw=none] table [x=n, y expr=-(\thisrow{F}+\thisrow{F_err})] {\filename};
    
  % Free Energy Error
  % Lower bound (invisible plot)
  \addplot [draw=none, stack plots=y, forget plot] table[
    x=n,
    y expr=\thisrow{L}-\thisrow{L_err}
  ] {\filename};	
  
  % Stack twice the error, draw as area plot
  \addplot [draw=none, fill=green!50!black, stack plots=y, fill opacity=0.15] table [
      x=n,
      y expr=2*\thisrow{L_err}
  ] {\filename}\closedcycle;
  
  % Reset stack using invisible plot
  \addplot [forget plot, stack plots=y,draw=none] table [x=n, y expr=-(\thisrow{L}+\thisrow{L_err})] {\filename};

	\addplot [solid, mark=None, color=green!50!black] table[x=n, y=L] {\filename};	
	\addplot [dashed, mark=None, color=green!50!black] table[x=n, y=F] {\filename};

  %--- Distances C'=200 ------------------------------------------
  \renewcommand\filename{./figs/KDD_GMM.txt}
  % Likelihood Error
  % Lower bound (invisible plot)
  \addplot [draw=none, stack plots=y, forget plot] table[
    x=n,
    y expr=\thisrow{F}-\thisrow{F_err}
  ] {\filename};	
  
  % Stack twice the error, draw as area plot
  \addplot [draw=none, fill=black, stack plots=y, fill opacity=0.15] table [
      x=n,
      y expr=2*\thisrow{F_err}
  ] {\filename}\closedcycle;
  
  % Reset stack using invisible plot
  \addplot [forget plot, stack plots=y,draw=none] table [x=n, y expr=-(\thisrow{F}+\thisrow{F_err})] {\filename};
    
  % Free Energy Error
  % Lower bound (invisible plot)
  \addplot [draw=none, stack plots=y, forget plot] table[
    x=n,
    y expr=\thisrow{L}-\thisrow{L_err}
  ] {\filename};	
  
  % Stack twice the error, draw as area plot
  \addplot [draw=none, fill=black, stack plots=y, fill opacity=0.15] table [
      x=n,
      y expr=2*\thisrow{L_err}
  ] {\filename}\closedcycle;
  
  % Reset stack using invisible plot
  \addplot [forget plot, stack plots=y,draw=none] table [x=n, y expr=-(\thisrow{L}+\thisrow{L_err})] {\filename};

	\addplot [solid, mark=None, color=black] table[x=n, y=L] {\filename};	
	\addplot [dashed, mark=None, color=black] table[x=n, y=F] {\filename};

  \end{axis}
%	\fill[white] (-0.7,-0.5) rectangle (2.7,2.05);
\end{tikzpicture}
			\hfill
	\end{subfigure}
  \begin{subfigure}[c]{0.24\textwidth}
%			\tikzset{external/remake next}
    	%\tikzset{external/remake next}
\tikzsetnextfilename{KDD_k-means_exhaustive_LF}
\pgfplotsset{
	grid style={dotted,gray},
	minor grid style={dotted,lightgray},
  tick label style = {font=\tiny\sansmath\sffamily},
  legend style = {font=\sansmath\sffamily},
  xlabel style = {font=\sansmath\sffamily},
  ylabel style = {font=\sansmath\sffamily},
	% to match the colors of the markers to the plot cycle list 'exotic'
  legend image code/.code={
    \draw[mark repeat=2,mark phase=2]
    plot coordinates {
      (0cm,0cm)
      (0.25cm,0cm)        %% default is (0.3cm,0cm)
      (0.5cm,0cm)         %% default is (0.6cm,0cm)
    };%	
  }
}

\begin{tikzpicture}%[rotate=-90]
	\tikzset{mark size={1.0}}
	\begin{axis}[
		colormap access=direct,
		width = 4.9cm,
		height = 3.5cm,
    	xmin=0,
		xmax=50,
		xtick = {0,10,...,50},
%		minor xtick = {0,10,...,500},
		scaled x ticks = false,
		xlabel near ticks, xticklabel pos=lower,
		ymin=-500,
		ymax = -450,
%		ytick = {-9.3,-9.1,...,-8.5},
%		minor ytick = {5,10,...,95},
		ylabel={\scriptsize\sffamily \phantom{Likelihood / Free Energy}},
		ylabel near ticks,
		yticklabel pos=left,
		grid = both,
		%
   %legend image code/.code={%
%%     \draw[dash pattern=on 0.175cm off 0.05cm on 0.05cm off 0.05cm on 0.175cm, draw=red!70!black] (0cm,0.05cm) -- (0.5cm,0.05cm);
     %\draw[dashed] (0cm,-0.025cm) -- (0.5cm,-0.025cm);
     %\draw[solid]  (0cm, 0.025cm) -- (0.5cm, 0.025cm);
    %},		
    %legend entries={
      %\fontsize{6}{0}\selectfont\sffamily \, $C'=2$,
      %\fontsize{6}{0}\selectfont\sffamily \, $C'=5$,
      %\fontsize{6}{0}\selectfont\sffamily \, $C'=20$,
      %\fontsize{6}{0}\selectfont\sffamily \, $C'=200$
    %},
		%legend style={
%%			draw = none,
			%at={(0.95,0.45)},
%%			anchor=north,
			%legend columns=1,
			%row sep=-2pt,
			%%column sep=0.5cm,
      %%inner sep=1.5pt,
		%},
		%legend cell align=left,
    %%reverse legend,
  	]
	%\addlegendimage{color=red!70!black}
  %\addlegendimage{color=blue!70!black}  	
  %\addlegendimage{color=green!50!black}  	
  %\addlegendimage{color=black}
%%  \addlegendimage{only marks, mark=*, color=black}
  %
  
  %--- Distances C'=2 ------------------------------------------
  % Likelihood Error
  % Lower bound (invisible plot)
  \addplot [draw=none, stack plots=y, forget plot] table[
    x=n,
    y expr=\thisrow{F}-\thisrow{F_err}
  ] {./figs/KDD_k-means_exhaustive_2.txt};	
  
  % Stack twice the error, draw as area plot
  \addplot [draw=none, fill=red!70!black, stack plots=y, fill opacity=0.15] table [
      x=n,
      y expr=2*\thisrow{F_err}
  ] {./figs/KDD_k-means_exhaustive_2.txt}\closedcycle;
  
  % Reset stack using invisible plot
  \addplot [forget plot, stack plots=y,draw=none] table [x=n, y expr=-(\thisrow{F}+\thisrow{F_err})] {./figs/KDD_k-means_exhaustive_2.txt};
    
  % Free Energy Error
  % Lower bound (invisible plot)
  \addplot [draw=none, stack plots=y, forget plot] table[
    x=n,
    y expr=\thisrow{L}-\thisrow{L_err}
  ] {./figs/KDD_k-means_exhaustive_2.txt};	
  
  % Stack twice the error, draw as area plot
  \addplot [draw=none, fill=red!70!black, stack plots=y, fill opacity=0.15] table [
      x=n,
      y expr=2*\thisrow{L_err}
  ] {./figs/KDD_k-means_exhaustive_2.txt}\closedcycle;
  
  % Reset stack using invisible plot
  \addplot [forget plot, stack plots=y,draw=none] table [x=n, y expr=-(\thisrow{L}+\thisrow{L_err})] {./figs/KDD_k-means_exhaustive_2.txt};

	\addplot [solid, mark=None, color=red!70!black] table[x=n, y=L] {./figs/KDD_k-means_exhaustive_2.txt};	
	\addplot [dashed, mark=None, color=red!70!black] table[x=n, y=F] {./figs/KDD_k-means_exhaustive_2.txt};

%  %--- Distances C'=5 ------------------------------------------
  % Likelihood Error
  % Lower bound (invisible plot)
  \addplot [draw=none, stack plots=y, forget plot] table[
    x=n,
    y expr=\thisrow{F}-\thisrow{F_err}
  ] {./figs/KDD_k-means_exhaustive_5.txt};	
  
  % Stack twice the error, draw as area plot
  \addplot [draw=none, fill=blue!70!black, stack plots=y, fill opacity=0.15] table [
      x=n,
      y expr=2*\thisrow{F_err}
  ] {./figs/KDD_k-means_exhaustive_5.txt}\closedcycle;
  
  % Reset stack using invisible plot
  \addplot [forget plot, stack plots=y,draw=none] table [x=n, y expr=-(\thisrow{F}+\thisrow{F_err})] {./figs/KDD_k-means_exhaustive_5.txt};
    
  % Free Energy Error
  % Lower bound (invisible plot)
  \addplot [draw=none, stack plots=y, forget plot] table[
    x=n,
    y expr=\thisrow{L}-\thisrow{L_err}
  ] {./figs/KDD_k-means_exhaustive_5.txt};	
  
  % Stack twice the error, draw as area plot
  \addplot [draw=none, fill=blue!70!black, stack plots=y, fill opacity=0.15] table [
      x=n,
      y expr=2*\thisrow{L_err}
  ] {./figs/KDD_k-means_exhaustive_5.txt}\closedcycle;
  
  % Reset stack using invisible plot
  \addplot [forget plot, stack plots=y,draw=none] table [x=n, y expr=-(\thisrow{L}+\thisrow{L_err})] {./figs/KDD_k-means_exhaustive_5.txt};

	\addplot [solid, mark=None, color=blue!70!black] table[x=n, y=L] {./figs/KDD_k-means_exhaustive_5.txt};	
	\addplot [dashed, mark=None, color=blue!70!black] table[x=n, y=F] {./figs/KDD_k-means_exhaustive_5.txt};

%  %--- Distances C'=20 ------------------------------------------
  % Likelihood Error
  % Lower bound (invisible plot)
  \addplot [draw=none, stack plots=y, forget plot] table[
    x=n,
    y expr=\thisrow{F}-\thisrow{F_err}
  ] {./figs/KDD_k-means_exhaustive_20.txt};	
  
  % Stack twice the error, draw as area plot
  \addplot [draw=none, fill=green!50!black, stack plots=y, fill opacity=0.15] table [
      x=n,
      y expr=2*\thisrow{F_err}
  ] {./figs/KDD_k-means_exhaustive_20.txt}\closedcycle;
  
  % Reset stack using invisible plot
  \addplot [forget plot, stack plots=y,draw=none] table [x=n, y expr=-(\thisrow{F}+\thisrow{F_err})] {./figs/KDD_k-means_exhaustive_20.txt};
    
  % Free Energy Error
  % Lower bound (invisible plot)
  \addplot [draw=none, stack plots=y, forget plot] table[
    x=n,
    y expr=\thisrow{L}-\thisrow{L_err}
  ] {./figs/KDD_k-means_exhaustive_20.txt};	
  
  % Stack twice the error, draw as area plot
  \addplot [draw=none, fill=green!50!black, stack plots=y, fill opacity=0.15] table [
      x=n,
      y expr=2*\thisrow{L_err}
  ] {./figs/KDD_k-means_exhaustive_20.txt}\closedcycle;
  
  % Reset stack using invisible plot
  \addplot [forget plot, stack plots=y,draw=none] table [x=n, y expr=-(\thisrow{L}+\thisrow{L_err})] {./figs/KDD_k-means_exhaustive_20.txt};

	\addplot [solid, mark=None, color=green!50!black] table[x=n, y=L] {./figs/KDD_k-means_exhaustive_20.txt};	
	\addplot [dashed, mark=None, color=green!50!black] table[x=n, y=F] {./figs/KDD_k-means_exhaustive_20.txt};

  %--- Distances C'=200 ------------------------------------------
  % Likelihood Error
  % Lower bound (invisible plot)
  \addplot [draw=none, stack plots=y, forget plot] table[
    x=n,
    y expr=\thisrow{F}-\thisrow{F_err}
  ] {./figs/KDD_k-means.txt};	
  
  % Stack twice the error, draw as area plot
  \addplot [draw=none, fill=black, stack plots=y, fill opacity=0.15] table [
      x=n,
      y expr=2*\thisrow{F_err}
  ] {./figs/KDD_k-means.txt}\closedcycle;
  
  % Reset stack using invisible plot
  \addplot [forget plot, stack plots=y,draw=none] table [x=n, y expr=-(\thisrow{F}+\thisrow{F_err})] {./figs/KDD_k-means.txt};
    
  % Free Energy Error
  % Lower bound (invisible plot)
  \addplot [draw=none, stack plots=y, forget plot] table[
    x=n,
    y expr=\thisrow{L}-\thisrow{L_err}
  ] {./figs/KDD_k-means.txt};	
  
  % Stack twice the error, draw as area plot
  \addplot [draw=none, fill=black, stack plots=y, fill opacity=0.15] table [
      x=n,
      y expr=2*\thisrow{L_err}
  ] {./figs/KDD_k-means.txt}\closedcycle;
  
  % Reset stack using invisible plot
  \addplot [forget plot, stack plots=y,draw=none] table [x=n, y expr=-(\thisrow{L}+\thisrow{L_err})] {./figs/KDD_k-means.txt};

	\addplot [solid, mark=None, color=black] table[x=n, y=L] {./figs/KDD_k-means.txt};	
	\addplot [dashed, mark=None, color=black] table[x=n, y=F] {./figs/KDD_k-means.txt};

  \end{axis}
\end{tikzpicture}
	\end{subfigure}
  \begin{subfigure}[c]{0.24\textwidth}
%      \tikzset{external/remake next}
    	%\tikzset{external/remake next}
\tikzsetnextfilename{KDD_k-means_estimated_LF}
\pgfplotsset{
	grid style={dotted,gray},
	minor grid style={dotted,lightgray},
  tick label style = {font=\tiny\sansmath\sffamily},
  legend style = {font=\sansmath\sffamily},
  xlabel style = {font=\sansmath\sffamily},
  ylabel style = {font=\sansmath\sffamily},
	% to match the colors of the markers to the plot cycle list 'exotic'
  legend image code/.code={
    \draw[mark repeat=2,mark phase=2]
    plot coordinates {
      (0cm,0cm)
      (0.25cm,0cm)        %% default is (0.3cm,0cm)
      (0.5cm,0cm)         %% default is (0.6cm,0cm)
    };%	
  }
}

\begin{tikzpicture}%[rotate=-90]
	\tikzset{mark size={1.0}}
	\begin{axis}[
		colormap access=direct,
		width = 4.9cm,
		height = 3.5cm,
    	xmin=0,
		xmax=50,
		xtick = {0,10,...,50},
%		minor xtick = {0,10,...,500},
		scaled x ticks = false,
		xlabel near ticks, xticklabel pos=lower,
		ymin=-500,
		ymax = -450,
%		ytick = {-9.3,-9.1,...,-8.5},
%		minor ytick = {5,10,...,95},
		ylabel={\scriptsize\sffamily \phantom{Likelihood / Free Energy}},
		ylabel near ticks,
		yticklabel pos=left,
		grid = both,
		%
   %legend image code/.code={%
%%     \draw[dash pattern=on 0.175cm off 0.05cm on 0.05cm off 0.05cm on 0.175cm, draw=red!70!black] (0cm,0.05cm) -- (0.5cm,0.05cm);
     %\draw[dashed] (0cm,-0.025cm) -- (0.5cm,-0.025cm);
     %\draw[solid]  (0cm, 0.025cm) -- (0.5cm, 0.025cm);
    %},		
    %legend entries={
%%      \fontsize{6}{0}\selectfont\sffamily \, $C'=2$,
      %\fontsize{6}{0}\selectfont\sffamily \, $C'=5$,
      %\fontsize{6}{0}\selectfont\sffamily \, $C'=20$,
      %\fontsize{6}{0}\selectfont\sffamily \, $C'=200$
    %},
		%legend style={
%%			draw = none,
			%at={(0.95,0.35)},
%%			anchor=north,
			%legend columns=1,
			%row sep=-2pt,
			%%column sep=0.5cm,
      %%inner sep=1.5pt,
		%},
		%legend cell align=left,
    %%reverse legend,
  	]
  \addplot [draw=none, stack plots=y, forget plot] table[
    x=n,
    y expr=\thisrow{F}-\thisrow{F_err}
  ] {./figs/KDD_k-means_estimated_5.txt};	
  
  % Stack twice the error, draw as area plot
  \addplot [draw=none, fill=blue!70!black, stack plots=y, fill opacity=0.15] table [
      x=n,
      y expr=2*\thisrow{F_err}
  ] {./figs/KDD_k-means_estimated_5.txt}\closedcycle;
  
  % Reset stack using invisible plot
  \addplot [forget plot, stack plots=y,draw=none] table [x=n, y expr=-(\thisrow{F}+\thisrow{F_err})] {./figs/KDD_k-means_estimated_5.txt};
    
  % Free Energy Error
  % Lower bound (invisible plot)
  \addplot [draw=none, stack plots=y, forget plot] table[
    x=n,
    y expr=\thisrow{L}-\thisrow{L_err}
  ] {./figs/KDD_k-means_estimated_5.txt};	
  
  % Stack twice the error, draw as area plot
  \addplot [draw=none, fill=blue!70!black, stack plots=y, fill opacity=0.15] table [
      x=n,
      y expr=2*\thisrow{L_err}
  ] {./figs/KDD_k-means_estimated_5.txt}\closedcycle;
  
  % Reset stack using invisible plot
  \addplot [forget plot, stack plots=y,draw=none] table [x=n, y expr=-(\thisrow{L}+\thisrow{L_err})] {./figs/KDD_k-means_estimated_5.txt};

	\addplot [solid, mark=None, color=blue!70!black] table[x=n, y=L] {./figs/KDD_k-means_estimated_5.txt};	
	\addplot [dashed, mark=None, color=blue!70!black] table[x=n, y=F] {./figs/KDD_k-means_estimated_5.txt};

  %--- Distances C'=20 ------------------------------------------
  % Likelihood Error
  % Lower bound (invisible plot)
  \addplot [draw=none, stack plots=y, forget plot] table[
    x=n,
    y expr=\thisrow{F}-\thisrow{F_err}
  ] {./figs/KDD_k-means_estimated_20.txt};	
  
  % Stack twice the error, draw as area plot
  \addplot [draw=none, fill=green!50!black, stack plots=y, fill opacity=0.15] table [
      x=n,
      y expr=2*\thisrow{F_err}
  ] {./figs/KDD_k-means_estimated_20.txt}\closedcycle;
  
  % Reset stack using invisible plot
  \addplot [forget plot, stack plots=y,draw=none] table [x=n, y expr=-(\thisrow{F}+\thisrow{F_err})] {./figs/KDD_k-means_estimated_20.txt};
    
  % Free Energy Error
  % Lower bound (invisible plot)
  \addplot [draw=none, stack plots=y, forget plot] table[
    x=n,
    y expr=\thisrow{L}-\thisrow{L_err}
  ] {./figs/KDD_k-means_estimated_20.txt};	
  
  % Stack twice the error, draw as area plot
  \addplot [draw=none, fill=green!50!black, stack plots=y, fill opacity=0.15] table [
      x=n,
      y expr=2*\thisrow{L_err}
  ] {./figs/KDD_k-means_estimated_20.txt}\closedcycle;
  
  % Reset stack using invisible plot
  \addplot [forget plot, stack plots=y,draw=none] table [x=n, y expr=-(\thisrow{L}+\thisrow{L_err})] {./figs/KDD_k-means_estimated_20.txt};

	\addplot [solid, mark=None, color=green!50!black] table[x=n, y=L] {./figs/KDD_k-means_estimated_20.txt};	
	\addplot [dashed, mark=None, color=green!50!black] table[x=n, y=F] {./figs/KDD_k-means_estimated_20.txt};

  %--- Distances C'=200 ------------------------------------------
  % Likelihood Error
  % Lower bound (invisible plot)
  \addplot [draw=none, stack plots=y, forget plot] table[
    x=n,
    y expr=\thisrow{F}-\thisrow{F_err}
  ] {./figs/KDD_k-means.txt};	
  
  % Stack twice the error, draw as area plot
  \addplot [draw=none, fill=black, stack plots=y, fill opacity=0.15] table [
      x=n,
      y expr=2*\thisrow{F_err}
  ] {./figs/KDD_k-means.txt}\closedcycle;
  
  % Reset stack using invisible plot
  \addplot [forget plot, stack plots=y,draw=none] table [x=n, y expr=-(\thisrow{F}+\thisrow{F_err})] {./figs/KDD_k-means.txt};
    
  % Free Energy Error
  % Lower bound (invisible plot)
  \addplot [draw=none, stack plots=y, forget plot] table[
    x=n,
    y expr=\thisrow{L}-\thisrow{L_err}
  ] {./figs/KDD_k-means.txt};	
  
  % Stack twice the error, draw as area plot
  \addplot [draw=none, fill=black, stack plots=y, fill opacity=0.15] table [
      x=n,
      y expr=2*\thisrow{L_err}
  ] {./figs/KDD_k-means.txt}\closedcycle;
  
  % Reset stack using invisible plot
  \addplot [forget plot, stack plots=y,draw=none] table [x=n, y expr=-(\thisrow{L}+\thisrow{L_err})] {./figs/KDD_k-means.txt};

	\addplot [solid, mark=None, color=black] table[x=n, y=L] {./figs/KDD_k-means.txt};	
	\addplot [dashed, mark=None, color=black] table[x=n, y=F] {./figs/KDD_k-means.txt};

  \end{axis}
\end{tikzpicture}
	\end{subfigure}\\
	\begin{picture}(0,0)
		\put(35,-42){\fontsize{8}{0}\selectfont\sffamily (a) var-GMM-X}
	\end{picture}
  \begin{subfigure}[c]{0.24\textwidth}
    \begin{adjustbox}{trim=9pt 0pt 0pt 14pt}
%      \tikzset{external/remake next}
      %\tikzset{external/remake next}
\tikzsetnextfilename{KDD_GMM_exhaustive_QE}
\pgfplotsset{
	grid style={dotted,gray},
	minor grid style={dotted,lightgray},
  tick label style = {font=\tiny\sansmath\sffamily},
  legend style = {font=\sansmath\sffamily},
  xlabel style = {font=\sansmath\sffamily},
  ylabel style = {font=\sansmath\sffamily},
	% to match the colors of the markers to the plot cycle list 'exotic'
  legend image code/.code={
    \draw[mark repeat=2,mark phase=2]
    plot coordinates {
      (0cm,0cm)
      (0.25cm,0cm)        %% default is (0.3cm,0cm)
      (0.5cm,0cm)         %% default is (0.6cm,0cm)
    };%	
  }
}

\begin{tikzpicture}%[rotate=-90]
	\tikzset{mark size={1.0}}
	\begin{axis}[
		colormap access=direct,
		width = 4.9cm,
		height = 3.5cm,
    xmin=0,
		xmax=50,
		xtick = {0,10,...,50},
%		minor xtick = {0,10,...,500},
		scaled x ticks = false,
		xlabel={\scriptsize\sffamily Iteration},
		xlabel near ticks, xticklabel pos=lower,
		ymin= 0.13e12,
		ymax = 0.2e12,
%		ytick = {-9.3,-9.1,...,-8.5},
%		minor ytick = {5,10,...,95},
		ylabel={\scriptsize\sffamily Quantization Error},
		ylabel near ticks,
		yticklabel pos=left,
    y label style={at={(-0.2,0.5)}},
    y tick label style={
        /pgf/number format/.cd,
            fixed,
            fixed zerofill,
            precision=1,
        /tikz/.cd
    },
		grid = both,
		%
%   legend image code/.code={%
%     \draw[dash pattern=on 0.175cm off 0.05cm on 0.05cm off 0.05cm on 0.175cm, draw=red!70!black] (0cm,0.05cm) -- (0.5cm,0.05cm);
%     \draw[dashed] (0cm,-0.025cm) -- (0.5cm,-0.025cm);
%     \draw[solid]  (0cm, 0.025cm) -- (0.5cm, 0.025cm);
%    },		
    %legend entries={
      %\fontsize{6}{0}\selectfont\sffamily \, $C'=2$,
      %\fontsize{6}{0}\selectfont\sffamily \, $C'=5$,
      %\fontsize{6}{0}\selectfont\sffamily \, $C'=20$,
      %\fontsize{6}{0}\selectfont\sffamily \, $C'=200$
    %},
		%legend style={
%%			draw = none,
			%at={(0.95,0.95)},
%%			anchor=north,
			%legend columns=1,
			%row sep=-2pt,
			%%column sep=0.5cm,
      %%inner sep=1.5pt,
		%},
		%legend cell align=left,
    %%reverse legend,
  	]
	%\addlegendimage{color=red!70!black}
  %\addlegendimage{color=blue!70!black}  	
  %\addlegendimage{color=green!50!black}  	
  %\addlegendimage{color=black}
%%  \addlegendimage{only marks, mark=*, color=black}
  %
  
  %--- Distances C'=2 ------------------------------------------
  % Phi Error
  % Lower bound (invisible plot)
  \addplot [draw=none, stack plots=y, forget plot] table[
    x=n,
    y expr=\thisrow{phi}-\thisrow{phi_err}
  ] {./figs/KDD_GMM_exhaustive_2.txt};	
  
  % Stack twice the error, draw as area plot
  \addplot [draw=none, fill=red!70!black, stack plots=y, fill opacity=0.15] table [
      x=n,
      y expr=2*\thisrow{phi_err}
  ] {./figs/KDD_GMM_exhaustive_2.txt}\closedcycle;
  
  % Reset stack using invisible plot
  \addplot [forget plot, stack plots=y,draw=none] table [x=n, y expr=-(\thisrow{phi}+\thisrow{phi_err})] {./figs/KDD_GMM_exhaustive_2.txt};
     
	\addplot [solid, mark=None, color=red!70!black] table[x=n, y=phi] {./figs/KDD_GMM_exhaustive_2.txt};	
	\addplot [dotted, thick, mark=None, color=red!70!black] table[x=n, y=phi] {./figs/KDD_GMM_exhaustive_2_best.txt};	

%  %--- Distances C'=5 ------------------------------------------
  % Phi Error
  % Lower bound (invisible plot)
  \addplot [draw=none, stack plots=y, forget plot] table[
    x=n,
    y expr=\thisrow{phi}-\thisrow{phi_err}
  ] {./figs/KDD_GMM_exhaustive_5.txt};	
  
  % Stack twice the error, draw as area plot
  \addplot [draw=none, fill=blue!70!black, stack plots=y, fill opacity=0.15] table [
      x=n,
      y expr=2*\thisrow{phi_err}
  ] {./figs/KDD_GMM_exhaustive_5.txt}\closedcycle;
  
  % Reset stack using invisible plot
  \addplot [forget plot, stack plots=y,draw=none] table [x=n, y expr=-(\thisrow{phi}+\thisrow{phi_err})] {./figs/KDD_GMM_exhaustive_5.txt};
     
	\addplot [solid, mark=None, color=blue!70!black] table[x=n, y=phi] {./figs/KDD_GMM_exhaustive_5.txt};	
	\addplot [dotted, thick, mark=None, color=blue!70!black] table[x=n, y=phi] {./figs/KDD_GMM_exhaustive_5_best.txt};

  %--- Distances C'=20 ------------------------------------------
  % Phi Error
  % Lower bound (invisible plot)
  \addplot [draw=none, stack plots=y, forget plot] table[
    x=n,
    y expr=\thisrow{phi}-\thisrow{phi_err}
  ] {./figs/KDD_GMM_exhaustive_20.txt};	
  
  % Stack twice the error, draw as area plot
  \addplot [draw=none, fill=green!50!black, stack plots=y, fill opacity=0.15] table [
      x=n,
      y expr=2*\thisrow{phi_err}
  ] {./figs/KDD_GMM_exhaustive_20.txt}\closedcycle;
  
  % Reset stack using invisible plot
  \addplot [forget plot, stack plots=y,draw=none] table [x=n, y expr=-(\thisrow{phi}+\thisrow{phi_err})] {./figs/KDD_GMM_exhaustive_20.txt};
     
	\addplot [solid, mark=None, color=green!50!black] table[x=n, y=phi] {./figs/KDD_GMM_exhaustive_20.txt};	
	\addplot [dotted, thick, mark=None, color=green!50!black] table[x=n, y=phi] {./figs/KDD_GMM_exhaustive_20_best.txt};	

  %--- Distances C'=200 ------------------------------------------
  % Phi Error
  % Lower bound (invisible plot)
  \addplot [draw=none, stack plots=y, forget plot] table[
    x=n,
    y expr=\thisrow{phi}-\thisrow{phi_err}
  ] {./figs/KDD_GMM.txt};	
  
  % Stack twice the error, draw as area plot
  \addplot [draw=none, fill=black, stack plots=y, fill opacity=0.15] table [
      x=n,
      y expr=2*\thisrow{phi_err}
  ] {./figs/KDD_GMM.txt}\closedcycle;
  
  % Reset stack using invisible plot
  \addplot [forget plot, stack plots=y,draw=none] table [x=n, y expr=-(\thisrow{phi}+\thisrow{phi_err})] {./figs/KDD_GMM.txt};
     
	\addplot [solid, mark=None, color=black] table[x=n, y=phi] {./figs/KDD_GMM.txt};	
	\addplot [dotted, thick, mark=None, color=black] table[x=n, y=phi] {./figs/KDD_GMM_best.txt};	

  \end{axis}
\end{tikzpicture}
    \end{adjustbox}
	\end{subfigure}
	\begin{picture}(0,0)
		\put(-130,-20){\rotatebox{90}{\rlap{\makebox[5.80cm]{KDD2004}}}}
	\end{picture}	
	\begin{picture}(0,0)
		\put(13,-38){\rotatebox{90}{\rlap{\makebox[5.80cm]{\hrulefill}}}}
		\put(13,155){\rotatebox{90}{\rlap{\makebox[5.40cm]{\hrulefill}}}}
		\put(35,-42){\fontsize{8}{0}\selectfont\sffamily (b) var-GMM-S}
	\end{picture}
  \begin{subfigure}[c]{0.24\textwidth}
    \begin{adjustbox}{trim=5pt 0pt 0pt 14pt}
%      \tikzset{external/remake next}
      %\tikzset{external/remake next}
\tikzsetnextfilename{KDD_GMM_estimated_QE}
\pgfplotsset{
	grid style={dotted,gray},
	minor grid style={dotted,lightgray},
  tick label style = {font=\tiny\sansmath\sffamily},
  legend style = {font=\sansmath\sffamily},
  xlabel style = {font=\sansmath\sffamily},
  ylabel style = {font=\sansmath\sffamily},
	% to match the colors of the markers to the plot cycle list 'exotic'
  legend image code/.code={
    \draw[mark repeat=2,mark phase=2]
    plot coordinates {
      (0cm,0cm)
      (0.25cm,0cm)        %% default is (0.3cm,0cm)
      (0.5cm,0cm)         %% default is (0.6cm,0cm)
    };%	
  }
}

\begin{tikzpicture}%[rotate=-90]
	\tikzset{mark size={1.0}}
	\begin{axis}[
		colormap access=direct,
		width = 4.9cm,
		height = 3.5cm,
    xmin=0,
		xmax=50,
		xtick = {0,10,...,50},
%		minor xtick = {0,10,...,500},
		scaled x ticks = false,
		xlabel={\scriptsize\sffamily Iteration},
		xlabel near ticks, xticklabel pos=lower,
		ymin= 0.13e12,
		ymax = 0.2e12,
%		ytick = {-9.3,-9.1,...,-8.5},
%		minor ytick = {5,10,...,95},
		ylabel={\scriptsize\sffamily \phantom{Quantization Error}},
		ylabel near ticks,
		yticklabel pos=left,
    y tick label style={
        /pgf/number format/.cd,
            fixed,
            fixed zerofill,
            precision=1,
        /tikz/.cd
    },
		grid = both,
		%
%   legend image code/.code={%
%     \draw[dash pattern=on 0.175cm off 0.05cm on 0.05cm off 0.05cm on 0.175cm, draw=red!70!black] (0cm,0.05cm) -- (0.5cm,0.05cm);
%     \draw[dashed] (0cm,-0.025cm) -- (0.5cm,-0.025cm);
%     \draw[solid]  (0cm, 0.025cm) -- (0.5cm, 0.025cm);
%    },		
    %legend entries={
      %%\fontsize{6}{0}\selectfont\sffamily \, $C'=2$,
      %\fontsize{6}{0}\selectfont\sffamily \, $C'=5$,
      %\fontsize{6}{0}\selectfont\sffamily \, $C'=20$,
      %\fontsize{6}{0}\selectfont\sffamily \, $C'=200$
    %},
		%legend style={
%%			draw = none,
			%at={(0.95,0.95)},
%%			anchor=north,
			%legend columns=1,
			%row sep=-2pt,
			%%column sep=0.5cm,
      %%inner sep=1.5pt,
		%},
		%legend cell align=left,
    %%reverse legend,
  	]
	%%\addlegendimage{color=red!70!black}
  %\addlegendimage{color=blue!70!black}  	
  %\addlegendimage{color=green!50!black}  	
  %\addlegendimage{color=black}
%%  \addlegendimage{only marks, mark=*, color=black}
  %
  
  %--- Distances C'=2 ------------------------------------------
  % Phi Error
  % Lower bound (invisible plot)
  \addplot [draw=none, stack plots=y, forget plot] table[
    x=n,
    y expr=\thisrow{phi}-\thisrow{phi_err}
  ] {./figs/KDD_GMM_estimated_2.txt};	
  
  % Stack twice the error, draw as area plot
  \addplot [draw=none, fill=red!70!black, stack plots=y, fill opacity=0.15] table [
      x=n,
      y expr=2*\thisrow{phi_err}
  ] {./figs/KDD_GMM_estimated_2.txt}\closedcycle;
  
  % Reset stack using invisible plot
  \addplot [forget plot, stack plots=y,draw=none] table [x=n, y expr=-(\thisrow{phi}+\thisrow{phi_err})] {./figs/KDD_GMM_estimated_2.txt};
     
	\addplot [solid, mark=None, color=red!70!black] table[x=n, y=phi] {./figs/KDD_GMM_estimated_2.txt};	
	\addplot [dotted, thick, mark=None, color=red!70!black] table[x=n, y=phi] {./figs/KDD_GMM_estimated_2_best.txt};	

%  %--- Distances C'=5 ------------------------------------------
  % Phi Error
  % Lower bound (invisible plot)
  \addplot [draw=none, stack plots=y, forget plot] table[
    x=n,
    y expr=\thisrow{phi}-\thisrow{phi_err}
  ] {./figs/KDD_GMM_estimated_5.txt};	
  
  % Stack twice the error, draw as area plot
  \addplot [draw=none, fill=blue!70!black, stack plots=y, fill opacity=0.15] table [
      x=n,
      y expr=2*\thisrow{phi_err}
  ] {./figs/KDD_GMM_estimated_5.txt}\closedcycle;
  
  % Reset stack using invisible plot
  \addplot [forget plot, stack plots=y,draw=none] table [x=n, y expr=-(\thisrow{phi}+\thisrow{phi_err})] {./figs/KDD_GMM_estimated_5.txt};
     
	\addplot [solid, mark=None, color=blue!70!black] table[x=n, y=phi] {./figs/KDD_GMM_estimated_5.txt};	
	\addplot [dotted, thick, mark=None, color=blue!70!black] table[x=n, y=phi] {./figs/KDD_GMM_estimated_5_best.txt};

  %--- Distances C'=20 ------------------------------------------
  % Phi Error
  % Lower bound (invisible plot)
  \addplot [draw=none, stack plots=y, forget plot] table[
    x=n,
    y expr=\thisrow{phi}-\thisrow{phi_err}
  ] {./figs/KDD_GMM_estimated_20.txt};	
  
  % Stack twice the error, draw as area plot
  \addplot [draw=none, fill=green!50!black, stack plots=y, fill opacity=0.15] table [
      x=n,
      y expr=2*\thisrow{phi_err}
  ] {./figs/KDD_GMM_estimated_20.txt}\closedcycle;
  
  % Reset stack using invisible plot
  \addplot [forget plot, stack plots=y,draw=none] table [x=n, y expr=-(\thisrow{phi}+\thisrow{phi_err})] {./figs/KDD_GMM_estimated_20.txt};
     
	\addplot [solid, mark=None, color=green!50!black] table[x=n, y=phi] {./figs/KDD_GMM_estimated_20.txt};	
	\addplot [dotted, thick, mark=None, color=green!50!black] table[x=n, y=phi] {./figs/KDD_GMM_estimated_20_best.txt};	

  %%--- Distances C'=200 ------------------------------------------
  % Phi Error
  % Lower bound (invisible plot)
  \addplot [draw=none, stack plots=y, forget plot] table[
    x=n,
    y expr=\thisrow{phi}-\thisrow{phi_err}
  ] {./figs/KDD_GMM.txt};	
  
  % Stack twice the error, draw as area plot
  \addplot [draw=none, fill=black, stack plots=y, fill opacity=0.15] table [
      x=n,
      y expr=2*\thisrow{phi_err}
  ] {./figs/KDD_GMM.txt}\closedcycle;
  
  % Reset stack using invisible plot
  \addplot [forget plot, stack plots=y,draw=none] table [x=n, y expr=-(\thisrow{phi}+\thisrow{phi_err})] {./figs/KDD_GMM.txt};
     
	\addplot [solid, mark=None, color=black] table[x=n, y=phi] {./figs/KDD_GMM.txt};	
	\addplot [dotted, thick, mark=None, color=black] table[x=n, y=phi] {./figs/KDD_GMM_best.txt};	

  \end{axis}
\end{tikzpicture}
    \end{adjustbox}
	\end{subfigure}
	\begin{picture}(0,-20)
		\put(11,-38){\rotatebox{90}{\rlap{\makebox[5.80cm]{\hrulefill}}}}
		\put(10,-38){\rotatebox{90}{\rlap{\makebox[5.80cm]{\hrulefill}}}}
		\put(10,155){\rotatebox{90}{\rlap{\makebox[5.40cm]{\hrulefill}}}}
		\put(11,155){\rotatebox{90}{\rlap{\makebox[5.40cm]{\hrulefill}}}}
		\put(31,-42){\fontsize{8}{0}\selectfont\sffamily (c) var-$k$-means-X}
	\end{picture}
  \begin{subfigure}[c]{0.24\textwidth}
    \begin{adjustbox}{trim=6pt 0pt 0pt 14pt}
%      \tikzset{external/remake next}
      %\tikzset{external/remake next}
\tikzsetnextfilename{KDD_k-means_exhaustive_QE}
\pgfplotsset{
	grid style={dotted,gray},
	minor grid style={dotted,lightgray},
  tick label style = {font=\tiny\sansmath\sffamily},
  legend style = {font=\sansmath\sffamily},
  xlabel style = {font=\sansmath\sffamily},
  ylabel style = {font=\sansmath\sffamily},
	% to match the colors of the markers to the plot cycle list 'exotic'
  legend image code/.code={
    \draw[mark repeat=2,mark phase=2]
    plot coordinates {
      (0cm,0cm)
      (0.25cm,0cm)        %% default is (0.3cm,0cm)
      (0.5cm,0cm)         %% default is (0.6cm,0cm)
    };%	
  }
}

\begin{tikzpicture}%[rotate=-90]
	\tikzset{mark size={1.0}}
	\begin{axis}[
		colormap access=direct,
		width = 4.9cm,
		height = 3.5cm,
    	xmin=0,
		xmax=50,
		xtick = {0,10,...,50},
%		minor xtick = {0,10,...,500},
		scaled x ticks = false,
		xlabel={\scriptsize\sffamily Iteration},
		xlabel near ticks, xticklabel pos=lower,
		ymin= 0.13e12,
		ymax = 0.2e12,
%		ytick = {-9.3,-9.1,...,-8.5},
%		minor ytick = {5,10,...,95},
		ylabel={\scriptsize\sffamily \phantom{Quantization Error}},
		ylabel near ticks,
		yticklabel pos=left,
    y tick label style={
        /pgf/number format/.cd,
            fixed,
            fixed zerofill,
            precision=1,
        /tikz/.cd
    },
		grid = both,
		%
%   legend image code/.code={%
%     \draw[dash pattern=on 0.175cm off 0.05cm on 0.05cm off 0.05cm on 0.175cm, draw=red!70!black] (0cm,0.05cm) -- (0.5cm,0.05cm);
%     \draw[dashed] (0cm,-0.025cm) -- (0.5cm,-0.025cm);
%     \draw[solid]  (0cm, 0.025cm) -- (0.5cm, 0.025cm);
%    },		
    %legend entries={
      %\fontsize{6}{0}\selectfont\sffamily \, $C'=2$,
      %\fontsize{6}{0}\selectfont\sffamily \, $C'=5$,
      %\fontsize{6}{0}\selectfont\sffamily \, $C'=20$,
      %\fontsize{6}{0}\selectfont\sffamily \, $C'=200$
    %},
		%legend style={
%%			draw = none,
			%%at={(0.95,0.95)},
%%			anchor=north,
			%legend columns=1,
			%row sep=-2pt,
			%%column sep=0.5cm,
      %%inner sep=1.5pt,
		%},
		%legend cell align=left,
    %%reverse legend,
  	]
	%\addlegendimage{color=red!70!black}
  %\addlegendimage{color=blue!70!black}  	
  %\addlegendimage{color=green!50!black}  	
  %\addlegendimage{color=black}
%%  \addlegendimage{only marks, mark=*, color=black}
  %
  
  %--- Distances C'=2 ------------------------------------------
  % Phi Error
  % Lower bound (invisible plot)
  \addplot [draw=none, stack plots=y, forget plot] table[
    x=n,
    y expr=\thisrow{phi}-\thisrow{phi_err}
  ] {./figs/KDD_k-means_exhaustive_2.txt};	
  
  % Stack twice the error, draw as area plot
  \addplot [draw=none, fill=red!70!black, stack plots=y, fill opacity=0.15] table [
      x=n,
      y expr=2*\thisrow{phi_err}
  ] {./figs/KDD_k-means_exhaustive_2.txt}\closedcycle;
  
  % Reset stack using invisible plot
  \addplot [forget plot, stack plots=y,draw=none] table [x=n, y expr=-(\thisrow{phi}+\thisrow{phi_err})] {./figs/KDD_k-means_exhaustive_2.txt};
     
	\addplot [solid, mark=None, color=red!70!black] table[x=n, y=phi] {./figs/KDD_k-means_exhaustive_2.txt};	
	\addplot [dotted, thick, mark=None, color=red!70!black] table[x=n, y=phi] {./figs/KDD_k-means_exhaustive_2_best.txt};	

%  %--- Distances C'=5 ------------------------------------------
  % Phi Error
  % Lower bound (invisible plot)
  \addplot [draw=none, stack plots=y, forget plot] table[
    x=n,
    y expr=\thisrow{phi}-\thisrow{phi_err}
  ] {./figs/KDD_k-means_exhaustive_5.txt};	
  
  % Stack twice the error, draw as area plot
  \addplot [draw=none, fill=blue!70!black, stack plots=y, fill opacity=0.15] table [
      x=n,
      y expr=2*\thisrow{phi_err}
  ] {./figs/KDD_k-means_exhaustive_5.txt}\closedcycle;
  
  % Reset stack using invisible plot
  \addplot [forget plot, stack plots=y,draw=none] table [x=n, y expr=-(\thisrow{phi}+\thisrow{phi_err})] {./figs/KDD_k-means_exhaustive_5.txt};
     
	\addplot [solid, mark=None, color=blue!70!black] table[x=n, y=phi] {./figs/KDD_k-means_exhaustive_5.txt};	
	\addplot [dotted, thick, mark=None, color=blue!70!black] table[x=n, y=phi] {./figs/KDD_k-means_exhaustive_5_best.txt};

  %--- Distances C'=20 ------------------------------------------
  % Phi Error
  % Lower bound (invisible plot)
  \addplot [draw=none, stack plots=y, forget plot] table[
    x=n,
    y expr=\thisrow{phi}-\thisrow{phi_err}
  ] {./figs/KDD_k-means_exhaustive_20.txt};	
  
  % Stack twice the error, draw as area plot
  \addplot [draw=none, fill=green!50!black, stack plots=y, fill opacity=0.15] table [
      x=n,
      y expr=2*\thisrow{phi_err}
  ] {./figs/KDD_k-means_exhaustive_20.txt}\closedcycle;
  
  % Reset stack using invisible plot
  \addplot [forget plot, stack plots=y,draw=none] table [x=n, y expr=-(\thisrow{phi}+\thisrow{phi_err})] {./figs/KDD_k-means_exhaustive_20.txt};
     
	\addplot [solid, mark=None, color=green!50!black] table[x=n, y=phi] {./figs/KDD_k-means_exhaustive_20.txt};	
	\addplot [dotted, thick, mark=None, color=green!50!black] table[x=n, y=phi] {./figs/KDD_k-means_exhaustive_20_best.txt};	

  %--- Distances C'=200 ------------------------------------------
  % Phi Error
  % Lower bound (invisible plot)
  \addplot [draw=none, stack plots=y, forget plot] table[
    x=n,
    y expr=\thisrow{phi}-\thisrow{phi_err}
  ] {./figs/KDD_k-means.txt};	
  
  % Stack twice the error, draw as area plot
  \addplot [draw=none, fill=black, stack plots=y, fill opacity=0.15] table [
      x=n,
      y expr=2*\thisrow{phi_err}
  ] {./figs/KDD_k-means.txt}\closedcycle;
  
  % Reset stack using invisible plot
  \addplot [forget plot, stack plots=y,draw=none] table [x=n, y expr=-(\thisrow{phi}+\thisrow{phi_err})] {./figs/KDD_k-means.txt};
     
	\addplot [solid, mark=None, color=black] table[x=n, y=phi] {./figs/KDD_k-means.txt};	
	\addplot [dotted, thick, mark=None, color=black] table[x=n, y=phi] {./figs/KDD_k-means_best.txt};	

  \end{axis}
\end{tikzpicture}
    \end{adjustbox}
	\end{subfigure}
	\begin{picture}(0,0)
		\put(9,-38){\rotatebox{90}{\rlap{\makebox[5.80cm]{\hrulefill}}}}
		\put(9,155){\rotatebox{90}{\rlap{\makebox[5.40cm]{\hrulefill}}}}
		\put(29,-42){\fontsize{8}{0}\selectfont\sffamily (d) var-$k$-means-S}
	\end{picture}
  \begin{subfigure}[c]{0.24\textwidth}
    \begin{adjustbox}{trim=8pt 0pt 0pt 14pt}
%      \tikzset{external/remake next}
      %\tikzset{external/remake next}
\tikzsetnextfilename{KDD_k-means_estimated_QE}
\pgfplotsset{
	grid style={dotted,gray},
	minor grid style={dotted,lightgray},
  tick label style = {font=\tiny\sansmath\sffamily},
  legend style = {font=\sansmath\sffamily},
  xlabel style = {font=\sansmath\sffamily},
  ylabel style = {font=\sansmath\sffamily},
	% to match the colors of the markers to the plot cycle list 'exotic'
  legend image code/.code={
    \draw[mark repeat=2,mark phase=2]
    plot coordinates {
      (0cm,0cm)
      (0.25cm,0cm)        %% default is (0.3cm,0cm)
      (0.5cm,0cm)         %% default is (0.6cm,0cm)
    };%	
  }
}

\begin{tikzpicture}%[rotate=-90]
	\tikzset{mark size={1.0}}
	\begin{axis}[
		colormap access=direct,
		width = 4.9cm,
		height = 3.5cm,
    	xmin=0,
		xmax=50,
		xtick = {0,10,...,50},
%		minor xtick = {0,10,...,500},
		scaled x ticks = false,
		xlabel={\scriptsize\sffamily Iteration},
		xlabel near ticks, xticklabel pos=lower,
		ymin= 0.13e12,
		ymax = 0.2e12,
%		ytick = {-9.3,-9.1,...,-8.5},
%		minor ytick = {5,10,...,95},
		ylabel={\scriptsize\sffamily \phantom{Quantization Error}},
		ylabel near ticks,
		yticklabel pos=left,
    y tick label style={
        /pgf/number format/.cd,
            fixed,
            fixed zerofill,
            precision=1,
        /tikz/.cd
    },
		grid = both,
		%
%   legend image code/.code={%
%     \draw[dash pattern=on 0.175cm off 0.05cm on 0.05cm off 0.05cm on 0.175cm, draw=red!70!black] (0cm,0.05cm) -- (0.5cm,0.05cm);
%     \draw[dashed] (0cm,-0.025cm) -- (0.5cm,-0.025cm);
%     \draw[solid]  (0cm, 0.025cm) -- (0.5cm, 0.025cm);
%    },		
    %legend entries={
%%      \fontsize{6}{0}\selectfont\sffamily \, $C'=2$,
      %\fontsize{6}{0}\selectfont\sffamily \, $C'=5$,
      %\fontsize{6}{0}\selectfont\sffamily \, $C'=20$,
      %\fontsize{6}{0}\selectfont\sffamily \, $C'=200$
    %},
		%legend style={
%%			draw = none,
			%at={(0.95,0.95)},
%%			anchor=north,
			%legend columns=1,
			%row sep=-2pt,
			%%column sep=0.5cm,
      %%inner sep=1.5pt,
		%},
		%legend cell align=left,
    %%reverse legend,
  	]
%%	\addlegendimage{color=red!70!black}
  %\addlegendimage{color=blue!70!black}  	
  %\addlegendimage{color=green!50!black}  	
  %\addlegendimage{color=black}
%%  \addlegendimage{only marks, mark=*, color=black}
  %
  
%  %--- Distances C'=2 ------------------------------------------
%  % Phi Error
%  % Lower bound (invisible plot)
%  \addplot [draw=none, stack plots=y, forget plot] table[
%    x=n,
%    y expr=\thisrow{phi}-\thisrow{phi_err}
%  ] {./figs/KDD_k-means_estimated_2.txt};	
%  
%  % Stack twice the error, draw as area plot
%  \addplot [draw=none, fill=red!70!black, stack plots=y, fill opacity=0.15] table [
%      x=n,
%      y expr=2*\thisrow{phi_err}
%  ] {./figs/KDD_k-means_estimated_2.txt}\closedcycle;
%  
%  % Reset stack using invisible plot
%  \addplot [forget plot, stack plots=y,draw=none] table [x=n, y expr=-(\thisrow{phi}+\thisrow{phi_err})] {./figs/KDD_k-means_estimated_2.txt};
%     
%	\addplot [solid, mark=None, color=red!70!black] table[x=n, y=phi] {./figs/KDD_k-means_estimated_2.txt};	
%	\addplot [dotted, thick, mark=None, color=red!70!black] table[x=n, y=phi] {./figs/KDD_k-means_estimated_2_best.txt};	

%  %--- Distances C'=5 ------------------------------------------
  % Phi Error
  % Lower bound (invisible plot)
  \addplot [draw=none, stack plots=y, forget plot] table[
    x=n,
    y expr=\thisrow{phi}-\thisrow{phi_err}
  ] {./figs/KDD_k-means_estimated_5.txt};	
  
  % Stack twice the error, draw as area plot
  \addplot [draw=none, fill=blue!70!black, stack plots=y, fill opacity=0.15] table [
      x=n,
      y expr=2*\thisrow{phi_err}
  ] {./figs/KDD_k-means_estimated_5.txt}\closedcycle;
  
  % Reset stack using invisible plot
  \addplot [forget plot, stack plots=y,draw=none] table [x=n, y expr=-(\thisrow{phi}+\thisrow{phi_err})] {./figs/KDD_k-means_estimated_5.txt};
     
	\addplot [solid, mark=None, color=blue!70!black] table[x=n, y=phi] {./figs/KDD_k-means_estimated_5.txt};	
	\addplot [dotted, thick, mark=None, color=blue!70!black] table[x=n, y=phi] {./figs/KDD_k-means_estimated_5_best.txt};

  %--- Distances C'=20 ------------------------------------------
  % Phi Error
  % Lower bound (invisible plot)
  \addplot [draw=none, stack plots=y, forget plot] table[
    x=n,
    y expr=\thisrow{phi}-\thisrow{phi_err}
  ] {./figs/KDD_k-means_estimated_20.txt};	
  
  % Stack twice the error, draw as area plot
  \addplot [draw=none, fill=green!50!black, stack plots=y, fill opacity=0.15] table [
      x=n,
      y expr=2*\thisrow{phi_err}
  ] {./figs/KDD_k-means_estimated_20.txt}\closedcycle;
  
  % Reset stack using invisible plot
  \addplot [forget plot, stack plots=y,draw=none] table [x=n, y expr=-(\thisrow{phi}+\thisrow{phi_err})] {./figs/KDD_k-means_estimated_20.txt};
     
	\addplot [solid, mark=None, color=green!50!black] table[x=n, y=phi] {./figs/KDD_k-means_estimated_20.txt};	
	\addplot [dotted, thick, mark=None, color=green!50!black] table[x=n, y=phi] {./figs/KDD_k-means_estimated_20_best.txt};	

  %--- Distances C'=200 ------------------------------------------
  % Phi Error
  % Lower bound (invisible plot)
  \addplot [draw=none, stack plots=y, forget plot] table[
    x=n,
    y expr=\thisrow{phi}-\thisrow{phi_err}
  ] {./figs/KDD_k-means.txt};	
  
  % Stack twice the error, draw as area plot
  \addplot [draw=none, fill=black, stack plots=y, fill opacity=0.15] table [
      x=n,
      y expr=2*\thisrow{phi_err}
  ] {./figs/KDD_k-means.txt}\closedcycle;
  
  % Reset stack using invisible plot
  \addplot [forget plot, stack plots=y,draw=none] table [x=n, y expr=-(\thisrow{phi}+\thisrow{phi_err})] {./figs/KDD_k-means.txt};
     
	\addplot [solid, mark=None, color=black] table[x=n, y=phi] {./figs/KDD_k-means.txt};	
	\addplot [dotted, thick, mark=None, color=black] table[x=n, y=phi] {./figs/KDD_k-means_best.txt};	

  \end{axis}
\end{tikzpicture}
    \end{adjustbox}
	\end{subfigure}
	\vspace{12pt}
  \caption{
Results of \cref{alg:var-GMM-X,alg:var-GMM-S} on the BIRCH~$5\times 5$ and KDD2004 data sets.
Each column (a)~-~(d) shows results for one specific algorithm for different $G$-values, depending on the data set.
For each data set, the first row shows the mean log-likelihood (solid) and mean free energy (dashed), shaded with their respective SEM.
The second row for each data set shows the mean quantization error (solid), shaded with its SEM, as well as the single run with lowest final quantization error (dotted).
For var-GMM on KDD the green ($G=20$) and black ($G=200$) plots are nearly indistinguishable from another.
}
  \label{fig:Detailed_Results}
\end{figure*}
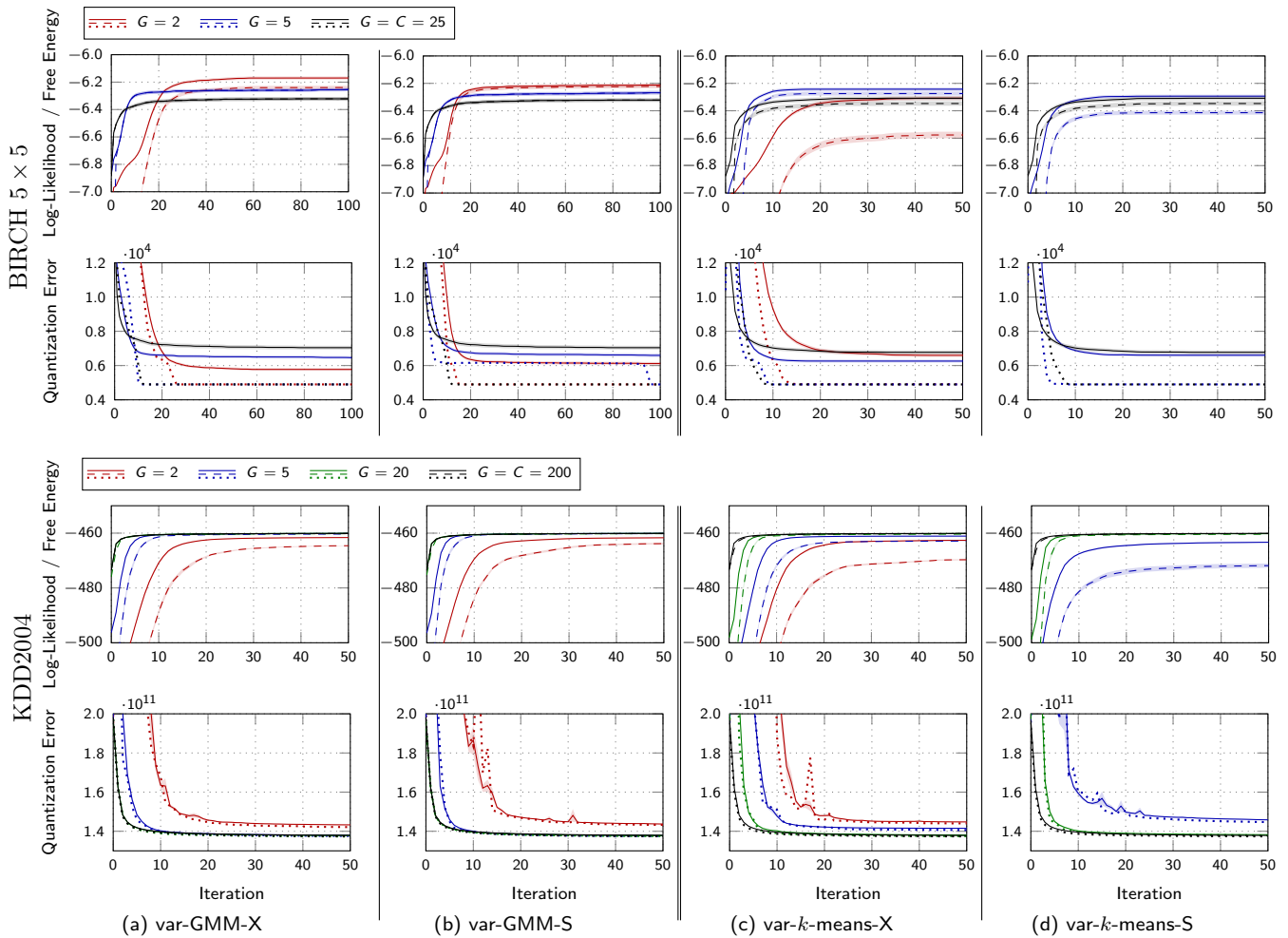

For \cref{alg:var-GMM-X,alg:var-GMM-S}, we monitor the free energy, log-likelihood and standard quantization error during training \citep[see][for the $k$-means free energy]{LuckeForster2017} on a $5 \times 5$ BIRCH data set and on KDD.
Means and standard errors of the means (SEM), as well as best runs in terms of lowest final quantization error are reported in \cref{fig:Detailed_Results} over 100~independent training runs for BIRCH and over 10~runs for KDD (except for standard GMM, where we only show results over a single training run because of its high computational demand).

As already noted, initially (in the first steps) var-GMM and var-$k$-means require more EM iterations than standard GMM or $k$-means to obtain comparable quantization errors due to the random $\KKn$/$\GGn$ initialization, which we see here also reflected in the likelihoods.
The number of additional EM iterations is however relatively small, and only more significant for very low values of $G$.

Most importantly for our study, low values of $G$, which result in strongly decreased run-time complexity, still optimize the clustering objective to values approximately equal to standard GMM and $k$-means within about the same number of EM iterations (in \cref{fig:Detailed_Results} the standard algorithms are $G=25$ for BIRCH and $G=200$ for KDD, black lines).
For BIRCH, the variational versions were even more effective in avoiding local optima.
The best final values of the objective function for var-GMM and var-$k$-means almost perfectly match those of the best standard $k$-means and standard GMM runs (black lines), which provides evidence for tight variational likelihood bounds.
Our primary goal is comparison to standard GMM and $k$-means but see, e.g., \citet{LucicEtAl2017} for results on KDD2004 of other recent approaches.
As $k$-means is itself a variational approximation of GMMs \citep{LuckeForster2017}, our partial E-step procedure makes the truncated free energy provably tight.
For var-GMM we have finite KL-divergence, but already for relatively small $G$ the free energy becomes almost tight (as the experiments confirm). 
Also for the large-scale and less regular KDD2004 clustering benchmark, var-GMM and var-$k$-means match the performance of standard GMM and $k$-means already for low $G$ ($G=5$ for var-GMM-S and $G=20$ for var-$k$-means-S).
Lower values of $G$ are still possible, but improved run-times trade off with a decrease in final objective function values.

Var-GMM-S on BIRCH with $G=2$ needs around ten more iterations to reach similar values than standard GMM with $C=25$, but each standard GMM iteration requires at least more than six times as many distance computations, making the few additional iterations negligible.
For the much larger KDD data set and $C=200$ clusters, the run-time difference further increase:
var-GMM-S requires with $G=5$ roughly an additional three iterations (i.e., plus three) than standard GMM, but each GMM iteration
requires at least eight times the number of data-to-cluster distance evaluations.
Still larger are the differences for var-$k$-means.
On BIRCH~$5\times 5$, var-$k$-means-S with $G=5$ converges basically as fast as $k$-means and standard GMM (maybe it needs one or two additional iterations), and it optimizes the objective to better final values on average.
The speedup per iteration is here at least a factor $(C/G=5)$.
On KDD-Cup 2004, var-$k$-means-S with $G=20$ optimizes the objective to the same values as $k$-means and GMM (requiring maybe two to four additional EM iterations).
But each var-$k$-means-S iteration is at least ten times more efficient than standard GMM or $k$-means.
With lower values of $G$ we can obtain even more significant speedups but then the final average objectives are more likely to be worse than for standard GMM and $k$-means.
In general, the more clusters we seek to find in the data, the more significant is the speedup of the variational algorithms (compare \cref{tab:speedup}).

\newpage

\section{\hspace{-0.6mm}Line-by-line Complexity, Algs.\,1 \&~2}
%\cref{alg:var-GMM-X,alg:var-GMM-S}}
\label{app:ComplexityAnalysis}

The clustering algorithms for arbitrary size $C'$ of $\KKn$ and arbitrary size $G$ of $\GG_c$ are given by \cref{alg:var-GMM-X,alg:var-GMM-S}.
In \cref{alg:var-GMM-X_Details,alg:var-GMM-S_Details} we rewrite the same algorithms such that the analysis of the complexity becomes particularly straight-forward.
For instance, we duplicate some loops that compute averages to show that such averaging does not increase the complexity.
The memory demand for storing all model parameters $\muVec_c$ and $\sigma$, and the new variational parameters $\KKn$ and nearest neighbors $\GGc$is of $\OO(CD+NC'+CG)$.
In \cref{alg:var-GMM-S_Details}, also the computed distances $\dcn$ have to be stored for all $N$ within each EM iteration (but do not have to be memorized across EM steps), leading to a memory demand of $\OO(CD+NC'G+CG)$.
For \cref{alg:var-GMM-X_Details}, memorizing all the distances is not necessary, as the loop over $N$ of the two last blocks can be combined and the updates can be computed without having to store all distances simultaneously.
Similar combinations may be possible for \cref{alg:var-GMM-S_Details} at least approximately but will require further investigations.

\newlength{\algrhswidth}
\setlength{\algrhswidth}{7cm}
\newcommand{\algrhsR}[1]{\Vhrulefill \ \parbox[t]{\algrhswidth}{#1}}
\newcommand{\algrhs}[1]{\hfill \ \parbox[t]{\algrhswidth}{#1}}
\newcommand{\algrhsup}[1]{\ \\[-5mm]\hfill \parbox[t]{\algrhswidth}{#1}}
\newcommand{\algro}[2]{\ \\[-4.7mm] \hfill \parbox[t]{1.5cm+#1}{#2}}
\newcommand{\dd}{\hspace{12mm}}
\newcommand{\ddd}{\dd\dd}
\newcommand{\dddd}{\dd\dd\dd}
\newcommand{\ddddd}{\dd\dd\dd\dd}

\vspace{2pt}
\begin{algorithm}[h]\vspace{0.5mm}
%\dontprintsemicolon
%
init $\muVec_{1:C}$, $\sigma$ and $\KKn$ for all $n$;\algBreak
%init $\Theta=(\muVec_{1:C},\sigma^2)$;\algBreak
%init $\KKn$ for all $n$;\algBreak
%
%init $\Theta=(\muVec_{1:C},\sigma^2)$;\algBreak
%
\Repeat{$\muVec_{1:C}$ and $\sigma^2$ have converged$\,^*$\vspace{1.mm}}
{ 
\For{$c=1:C$}{			        \algro{0cm}{\boldmath$\OO(C^2D)$}\algBreak								
  \For{$\ct=1:C$}{						\algro{0.2cm}{$\OO(CD)$}\algBreak			
    $\dcct = \|\muVec_{\ct}\,-\,\muVec_c\|$;  \algro{0.4cm}{$\OO(D)$}\algBreak  
  }
%  $\GGc=\{c'\,|\,d_{cc'}$ is among the $G$ shortest distaces$\}$;
%  $\GGc=\{c'\,|\,$for all $\ct\not\in\GGc: d_{cc'}<d_{c\ct}$$\}$;
  $\GGc=\{\ct\,|\,d_{c\ct}$ is among the $G$ \algro{0.2cm}{$\OO(C)$} \\
  \hspace{0mm}\hphantom{$\GGc=\{\ct\,|$} smallest distances $d_{c:}\}$;  \\[-1.5mm] \hrulefill 
}
\For{$n=1:N$}{    \algro{0.6cm}{\boldmath$\OO(NC'GD)$}\algBreak
  $\GGn =\bigcup_{c\in\KKn}\GGc$;    \algro{0.8cm}{$\OO(C'G)$}\algBreak
  \For{$c\in\GGn$}{                  \algro{0.8cm}{$\OO(C'GD)$}\algBreak
      $d^{(n)}_{c} = \|\yVecN\,-\,\muVec_{c}\|$; \algro{1cm}{$\OO(D)$}
  }
  $\KKn=\{c\,|\,d^{(n)}_{c}$ is among the \algro{0.8cm}{$\OO(C'G)$} \\ 
  \hspace{0mm}\hphantom{$\KKn=\{c\,|$} $C'$ smallest distances$\}$; \\[-1.5mm] \hrulefill 
}
\For{$n=1:N$}{    \algro{0cm}{\boldmath$\OO(NC')$}\algBreak	
%  $\GGn = \cup_{c \in \KKn}\GG_c$;\\
  $\overline{s}=0$;\algro{0.2cm}{$\OO(1)$}\algBreak 	
  \For{$c\in\KKn$}{    \algro{0.2cm}{$\OO(C')$}\algBreak
    $s_c^{(n)} = \exp\big(-\frac{1}{2}(\dcn / \sigma)^2\big)$; \algro{0.4cm}{$\OO(1)$}\algBreak
    $\overline{s} = \overline{s} + s_c^{(n)}$; \algro{0.4cm}{$\OO(1)$}\algBreak
  }
  \For{$c\in\KKn$}{ \algro{0.2cm}{$\OO(C')$}\algBreak
    $s_c^{(n)} = s_c^{(n)}\,/\,\overline{s}$;  \algro{0.4cm}{$\OO(1)$}\algBreak 
  } \hrulefill
} 
update $\muVec_{1:C}$ and $\sigma^2$ using \algro{0.3cm}{\boldmath$\OO(NC'D)$} \\[1mm] \cref{EqnGMMMStep} with \cref{EqnQMain};   \algBreak                  
}
\caption{Explicit reformulation of \cref{alg:var-GMM-X}.\label{alg:var-GMM-X_Details}}
\end{algorithm}

\begin{figure}[!b]
\vspace{-200pt}
\mbox{\footnotesize $^*$Except for $\KKn\!$ and $\GGc$, all sets and variables are reset after each EM iteration. First iteration of \cref{alg:var-GMM-S_Details} uses initial $\GGn\!$.}
\end{figure}

\newpage

In principle we could drop the last term $\OO(CG)$ in the memory demand of var-GMM-S as $NC'G > CG$.
For our purposes, we maintained the last term in order to make the linear $C$ dependence of the memory explicit. 
\vspace{-15.5pt}

\begin{algorithm}[h!]\vspace{0.5mm}
%\dontprintsemicolon
%
init $\muVec_{1:C}$ and $\sigma^2$;
%using seeding;\algBreak
%init $\Theta=(\muVec_{1:C},\sigma^2)$;\algBreak
init $\GGn$ for all $n$;\algBreak
\Repeat{$\muVec_{1:C}$ and $\sigma^2$ have converged$\,^*$\vspace{1.mm}}
{
\For{$n=1:N$}{    \algro{0.6cm}{\boldmath$\OO(NC'GD)$}\algBreak
  $\GGn =\bigcup_{c\in\KKn}\GGc$;    \algro{0.8cm}{$\OO(C'G)$}\algBreak
  \For{$c\in\GGn$}{                  \algro{0.8cm}{$\OO(C'GD)$}\algBreak
      $d^{(n)}_{c} = \|\yVecN\,-\,\muVec_{c}\|$; \algro{1cm}{$\OO(D)$}
  }
  $\KKn=\{c\,|\,d^{(n)}_{c}$ is among the \algro{0.8cm}{$\OO(C'G)$} \\ 
  \hspace{0mm}\hphantom{$\KKn=\{c\,|$}$C'$ smallest distances$\}$; \\[-1.5mm] \hrulefill 
}
\For{$n=1:N$}{ \algro{0.3cm}{\boldmath$\OO(NC'G)$}\algBreak		
  $\con = \myargmin{c\in\GGn}\,\big\{d^{(n)}_{c}\big\}$; \algro{0.6cm}{$\OO(C'G)$}\\
  $\II_{\con} = \II_{\con} \cup \{n\}$;               \algro{0.6cm}{$\OO(1)$} \\ \hrulefill 
} 
\For{$c=1:C$}{          \algro{0.3cm}{\boldmath$\OO(NC'G)$}\algBreak		
  \For{$n\in\II_c$}{    \algro{0.8cm}{$\OO((N/C)C'G)$}\algBreak	
    \For{$\ct\in\GGn$}{ \algro{1cm}{$\OO(C'G)$}\algBreak	
        $d_{c\ct} = d_{c\ct} + d^{(n)}_{\ct};$ \algro{1.2cm}{$\OO(1)$}\algBreak
        $b_{c\ct} = b_{c\ct} + 1;$             \algro{1.2cm}{$\OO(1)$}\algBreak
    }
  } \hrulefill 
}
\For{$c=1:C$}{							\algro{0.3cm}{\boldmath$\OO(NC'G)$}\algBreak	
  \For{$n\in\II_c$}{				\algro{0.8cm}{$\OO((N/C)C'G)$}\algBreak	
    \For{$\ct\in\GGn$}{     \algro{1cm}{$\OO(C'G)$}\algBreak
        \If{$\mathrm{normalized}_{c\ct}\neq{}1$}{	
          $d_{c\ct} = d_{c\ct} / b_{c\ct}$;\algro{1.2cm}{$\OO(1)$}\algBreak				
          $\mathrm{normalized}_{c\ct}=1$;\algro{1.2cm}{$\OO(1)$}
        }
    }
  }
  $d_{cc}=0;$           \algro{0.8cm}{$\OO(1)$}\algBreak
  $\GGc= \{\ct\,|\,d_{c\ct}$ is among the $G$ \algro{0.8cm}{$\OO((N/C)C'G)$} \\
  \hspace{0mm}\hphantom{$\GGc= \{\ct\,|$} smallest distances $d_{c:}\}$;  \\[-1.5mm] \hrulefill
}
\For{$n=1:N$}{								\algro{0cm}{\boldmath$\OO(NC')$}\algBreak	
%  $\GGn = \cup_{c \in \KKn}\GG_c$;\\
  $\overline{s}=0$;\algro{0.2cm}{$\OO(1)$}\algBreak 	
  \For{$c\in\KKn$}{           \algro{0.2cm}{$\OO(C')$}\algBreak
    $s_c^{(n)} = \exp\big(-\frac{1}{2}(\dcn / \sigma)^2\big)$;  \algro{0.4cm}{$\OO(1)$}\algBreak
    $\overline{s} = \overline{s} + s_c^{(n)}$   \algro{0.4cm}{$\OO(1)$}\algBreak
  }
  \For{$c\in\KKn$}{        \algro{0.2cm}{$\OO(C')$}\algBreak
    $s_c^{(n)} = s_c^{(n)}\,/\,\overline{s}$;  \algro{0.4cm}{$\OO(1)$}\algBreak
  } \hrulefill
}
update $\muVec_{1:C}$ and $\sigma^2$ using \algro{0.3cm}{\boldmath$\OO(NC'D)$} \\[1mm] \cref{EqnGMMMStep} with \cref{EqnQMain};   \algBreak
}
\caption{Explicit reformulation of \cref{alg:var-GMM-S}.\label{alg:var-GMM-S_Details}}
\end{algorithm}

\end{document}